\documentclass[pdflatex,sn-mathphys-num]{sn-jnl}

\usepackage{graphicx}
\usepackage{multirow}
\usepackage{amsmath,amssymb,amsfonts}
\usepackage{amsthm}
\usepackage{mathrsfs}
\usepackage[title]{appendix}
\usepackage{xcolor}
\usepackage{textcomp}
\usepackage{manyfoot}
\usepackage{booktabs}
\usepackage{pdfpages}
\usepackage{algorithm}
\usepackage{algorithmicx}
\usepackage{algpseudocode}
\usepackage{listings}
\usepackage{epigraph}
\usepackage{cleveref}
\usepackage{xr}
\usepackage{lmodern}
\usepackage{anyfontsize}
\usepackage{dcolumn,caption,booktabs}
\usepackage{subcaption}

\newcounter{extfig}
\newcounter{exttab}

\crefname{extfig}{Extended Data Figure}{Extended Data Figures}
\Crefname{extfig}{Extended Data Figure}{Extended Data Figures}
\crefname{exttab}{Extended Data Table}{Extended Data Tables}
\Crefname{exttab}{Extended Data Table}{Extended Data Tables}

\newcommand{\extfigcaption}[1]{%
  \refstepcounter{extfig}%
  \caption*{\textbf{Extended Data Figure \theextfig.} #1}%
}

\newcommand{\exttabcaption}[1]{%
  \refstepcounter{exttab}%
  \caption*{\textbf{Extended Data Table \theexttab.} #1}%
}

\usepackage{pifont}

\newcommand{\xmark}{\ding{55}}

\begin{document}

\title{The Shrinking Landscape of Linguistic Diversity in the Age of Large Language Models}

\author*[1,2]{\fnm{Zhivar} \sur{Sourati}}\email{souratih@usc.edu}

\author[2,3]{\fnm{Farzan} \sur{Karimi-Malekabadi}}\email{karimima@usc.edu}
\equalcont{These authors contributed equally to this work.}

\author[3]{\fnm{Meltem} \sur{Ozcan}}\email{ozcan@usc.edu}
\equalcont{These authors contributed equally to this work.}

\author[3]{\fnm{Colin} \sur{McDaniel}}\email{cbmcdani@usc.edu}

\author[1,2]{\fnm{Alireza} \sur{Ziabari}}\email{salkhord@usc.edu}

\author[2,3]{\fnm{Jackson} \sur{Trager}}\email{jptrager@usc.edu}

\author[1]{\fnm{Ala} \sur{N. Tak}}\email{nekouvag@usc.edu}

\author[3]{\fnm{Meng} \sur{Chen}}\email{mchen421@usc.edu}

\author[1,4]{\fnm{Fred} \sur{Morstatter}}\email{fred@isi.edu}

\author[1,2,3]{\fnm{Morteza} \sur{Dehghani}}\email{mdehghan@usc.edu}

\affil*[1]{\orgdiv{Department of Computer Science}, \orgname{University of Southern California}}

\affil[2]{\orgdiv{Center for Computational Language Sciences}, \orgname{University of Southern California}}

\affil[3]{\orgdiv{Department of Psychology}, \orgname{University of Southern California}}

\affil[4]{\orgdiv{Information Sciences Institute}, \orgname{University of Southern California}}

\abstract{
Language is far more than a communication tool; it encodes a wealth of information about a person's identity, psychological state, and social context, providing valuable insights for diverse fields including psychology, marketing, and healthcare. Across three studies spanning seven datasets in different domains and over 880,000 texts, we show that the widespread adoption of large language models (LLMs) as writing assistants is linked to declines in linguistic diversity, interfering with the societal and psychological insights language provides. While core content is retained when LLMs polish and rewrite texts, LLMs also homogenize writing styles, reducing writing-complexity variance by a statistically significant 21–50\% across datasets and models ($p \leq .05$), and amplify patterns associated with dominant characteristics while suppressing others, emphasizing conformity over individuality. These trends hold across different LLMs, prompts, and contexts, with potential implications for diagnostic processes, personalization efforts, hiring assessments, and cultural preservation.\footnote{This is the author's version of a work published in Nature Human Behaviour. The definitive version is available at https://doi.org/10.1038/s41562-026-02550-0}
}


\maketitle
\clearpage
\section*{Introduction}

``We're getting the language into its final shape-- the shape it's going to have when nobody speaks anything else.'' These words come from George Orwell's \textit{1984}\citep[][p.69]{georgeorwell}, a novel in which a government gradually constrains the language of its people, narrowing vocabulary and stripping away the nuances and variations through which different individuals express themselves until what remains is a single uniform language shared by all, a deliberate effort to narrow the range of thought itself. The setting is fictional, but the underlying intuition that the variety in our language carries something of real consequence is not. The words we speak tell a story, of our individual identities \citep{park2015automatic,oberlander2006language,moreno2021can,mairesse2007using,schwartz2013personality}, the diverse communities we inhabit \citep{kramsch2014language,gumperz1968speech,nguyen2011language,bamman2014gender}, and the vibrant societies that shape us \citep{pennebaker2011secret,robinson2017mind,huang2016understanding}.
They offer a unique lens through which we can explore the intricate tapestry of human thought, culture, and social dynamics. 
From the regional dialects that span a nation \citep{huang2016understanding,eisenstein2010latent} to the subtle markers of gender and power in our daily interactions \citep{bamman2014gender,peterson2011email}, language serves as a fingerprint for both individuals and groups \citep{stamatatos2009survey,grieve2007quantitative}. This rich variation allows researchers to trace the formation of cultural norms \citep{cassell2005language,danet2007multilingual}, reveal moral values and social biases \citep{kennedy2021moral,jackson2019loosening,hofmann2024ai}, and even diagnose and track cognitive and mental health conditions \citep{richard2024linguistic,eyigoz2020linguistic,roark2011spoken,trifu2024linguistic,weerasinghe2019because}. As these linguistic systems structure the ``kaleidoscopic flux of impressions'' \citep{sapir1956language} that constitutes our perception of the world, they are fundamental to understanding the human experience.

Crucially, the richness of these insights fundamentally relies on the fact that language is not monolithic; people use language differently \citep{pennebaker2011secret,eckert2012three,hofstede2001culture,kramsch2014language}, even when conveying similar ideas. This diversity transforms language from a mere medium of information exchange into a unique signature of individuals and societies, enabling analyses of cultural products and societal trends. 

However, the rise of large language models (LLMs), such as ChatGPT \citep{chatgpt} and Gemini \citep{gemini}, presents a substantial challenge to the preservation of linguistic diversity. As these models become increasingly integrated into our daily lives, serving over 800 million users worldwide \citep{bailyn2026chatgpt} for tasks ranging from composing emails to software development and drafting articles or other technical writing \citep{intelligent2024,mcclain2024,handa2025economic}, their very design poses a potential threat. LLMs are trained to generate the most statistically likely continuation of a given text, a process that inherently favors dominant language patterns. This focus on statistical likelihood risks undermining the rich diversity of linguistic expression, potentially eroding the variety of languages and dialects that contribute to the global cultural landscape. While prompts can introduce some customization \citep{mizrahi2024state,safdari2023personality}, the output remains constrained by the same underlying system \citep{ghosh2024closer,santurkar2023whose}. 

The widespread adoption of LLMs carries a substantial risk of homogenizing language use, potentially diminishing the linguistic diversity that allows us to analyze societal and individual trends \citep{ireland2014natural}. The loss of this informative linguistic variability could have far-reaching consequences, such as obscuring crucial markers used to identify specific subpopulations or individuals. For instance, detecting individuals expressing depression or suicidal thoughts could become more difficult with increasingly homogenized writing styles, hindering early diagnosis and treatment \citep{trifu2024linguistic,weerasinghe2019because,corona2023natural,rude2004language,coppersmith2018natural}. Furthermore, industries that rely on nuanced text-based analysis for mass personalization \citep{matz2017using, winter2021effects}, such as marketing, could be affected; by blurring linguistic variabilities, LLMs may reduce the effectiveness of targeted advertising and product development \citep{sundar2010personalization}. Finally, the homogenization of language use could also erase valuable cultural markers, threatening the preservation of linguistic heritage \citep{ryan2024unintended}. In essence, the pervasive use of LLMs could inadvertently lead to a flattening of linguistic expression, with potentially detrimental effects on our ability to understand and respond to individual and societal needs.

The threat of homogenization and loss of linguistic diversity becomes even more alarming when this homogenization aligns disproportionately with the linguistic patterns of specific subpopulations. Previous research has demonstrated that LLMs often overrepresent certain populations with specific affiliations or demographic attributes while underrepresenting others \citep[e.g.,][]{navigli2023biases,atari2023humans,wang2024large,rozado2024political,pan2023llms,abdurahman2023perils}. Such imbalances, combined with language homogenization, risk amplifying the voices of dominant groups while further marginalizing underrepresented communities.

Indeed, a growing body of research has begun to examine the linguistic trace of LLMs across online platforms. Much of this work has focused on documenting the prevalence of LLM-generated text through shifts in specific vocabulary items or stylistic markers \citep{kobak2025delving, liang2025quantifying, liang2025widespread, bao2025examining}, or on characterizing the properties of fully AI-generated outputs, such as reduced lexical diversity \citep{guo2024curious, munoz2024contrasting, xu2025echoes}. In AI-assisted settings, controlled experiments have shown that co-writing with LLMs can increase stylistic similarity among authors \citep{padmakumar2023does, doshi2024generative, anderson2024homogenization, agarwal2025ai}, and work on societal consequences has shown that fully AI-generated content can obscure socioeconomic and demographic diversity \citep{moon2024homogenizing, alvero2024large} and that LLMs fail to capture the psychological variability of human populations when responding to personality and values instruments \citep{abdurahman2023perils, wang2025large, zhang2025generative,lee2025poor}. However, these studies have not directly measured whether the overall diversity of writing styles within a community declines as LLM usage grows, nor have they examined the effect of AI on the unique sociolinguistic markers of individuals in scenarios where humans use LLMs to edit writings that already contain such rich signals. In this paper, we address these gaps. Using large-scale observational data spanning diverse communication domains, we directly measure temporal declines in linguistic diversity across real-world platforms and demonstrate a significant predictive link between AI adoption rates and these declines, and reveal how these dynamics vary across domains. We further shift the focus from fully generated content to the more common scenario where humans use LLMs to polish their own writing, allowing us to examine what happens to the identity signals already embedded in human-authored texts, trace homogenization effects in realistic editing settings and, crucially, to quantify the downstream consequences for the fields that depend on linguistic variation as a diagnostic tool, such as psychology, sociolinguistics, and clinical science.

To demonstrate this, our research proceeds in three stages. In Study 1 (a \& b), we first establish the phenomenon of language homogenization: Through observational analysis of real-world data as well as a controlled experiment, we demonstrate that linguistic diversity has indeed declined following the widespread adoption of LLMs. In Study 2, we then investigate a key societal implication of this trend: the erosion of identity signals. We examine whether linguistic homogenization compromises our ability to infer authors' personal traits, such as personality, political leanings, or empathy, from text. Finally, in Study 3, we probe the mechanisms underlying this erosion and explore its consequences by examining whether well-established associations between lexical cues and personal traits remain reliable with LLMs' involvement in writing.

By analyzing trends in linguistic diversity across a variety of communication forms and studying how LLMs affect human-authored texts (in realistic human–LLM editing scenarios), our studies offer a comprehensive understanding of how these models are reshaping human communication. We demonstrate that LLMs diminish linguistic diversity and alter the societal and psychological insights we can derive from language. To ensure the robustness and generalizability of our findings, we systematically vary LLMs, prompts, and classifiers and confirm that the observed trends are not artifacts of specific design choices but reflect consistent patterns that persist across different configurations. We make our code and data publicly available.

\section*{Results}

\subsection*{Study 1}
Study 1 examines whether the growing use of LLMs is associated with a homogenization of writing styles. We first investigate large-scale, real-world data to assess whether linguistic variance has declined alongside the rise of LLM adoption (Study 1a). We then complement this observational evidence with a controlled experiment that simulates a common real-world use case, where human-authored texts are polished by LLMs, to directly examine the influence of LLM involvement on stylistic diversity (Study 1b).

\subsubsection*{Study 1a}

To establish the real-world phenomenon of linguistic homogenization, Study 1a investigates whether the rise of LLMs correlates with a decline in linguistic diversity by tracking temporal shifts in stylistic variance in writings across three diverse, longitudinal datasets: creative writing, news articles, and academic preprints. Specifically, we investigate potential homogenization of writing styles after the introduction of LLMs by examining the relationship between LLM usage and the variance of features associated with linguistic complexity. This involves gathering diverse online user-generated content, assessing the extent to which this content is likely to be written by an LLM, measuring the variance in linguistic complexity within that content, and examining how the introduction of LLMs and the subsequent prevalence of AI-written content relate to changes in the observed variance of linguistic complexity.

Our analyses utilize three large-scale datasets of individual writing: creative stories from Reddit, community news articles from Patch, and computer science research abstracts from arXiv. Collectively, these sources provide a comprehensive basis for studying linguistic diversity as they span personal, professional, and scholarly writing styles (see \Cref{method:gathering-diverse-online-user-generated-content} for details). We classify texts as human- or LLM-written using \emph{Binoculars} \citep{hans2024spotting}, as a high-precision detection tool \citep{tufts2024practical,dugan2024raid,intellectualead2025ai_detectors} (see \Cref{method:detection-ai-written-text} for details). To examine how the prevalence of LLM-written texts relates to changes in the variance of linguistic complexity, we employ two complementary analyses (see \Cref{method:relationship-ai-usage-variance-writing-complexity} for details): a discontinuous growth model \citep{singer2003applied} to test whether ChatGPT’s launch produced a measurable shock in variance, and Granger causality tests \citep{linden2015conducting} to assess whether fluctuations in LLM usage predict subsequent changes in variance of linguistic complexity. To additionally capture potential nonlinear temporal dependencies, we complement the Granger test with a nonlinear extension based on generalized additive models \citep{bell1996non}. Together, these analyses capture both immediate and sustained effects of LLM adoption on linguistic homogenization.


\begin{figure}[!htbp]
    \centering
    \includegraphics[width=1\textwidth]{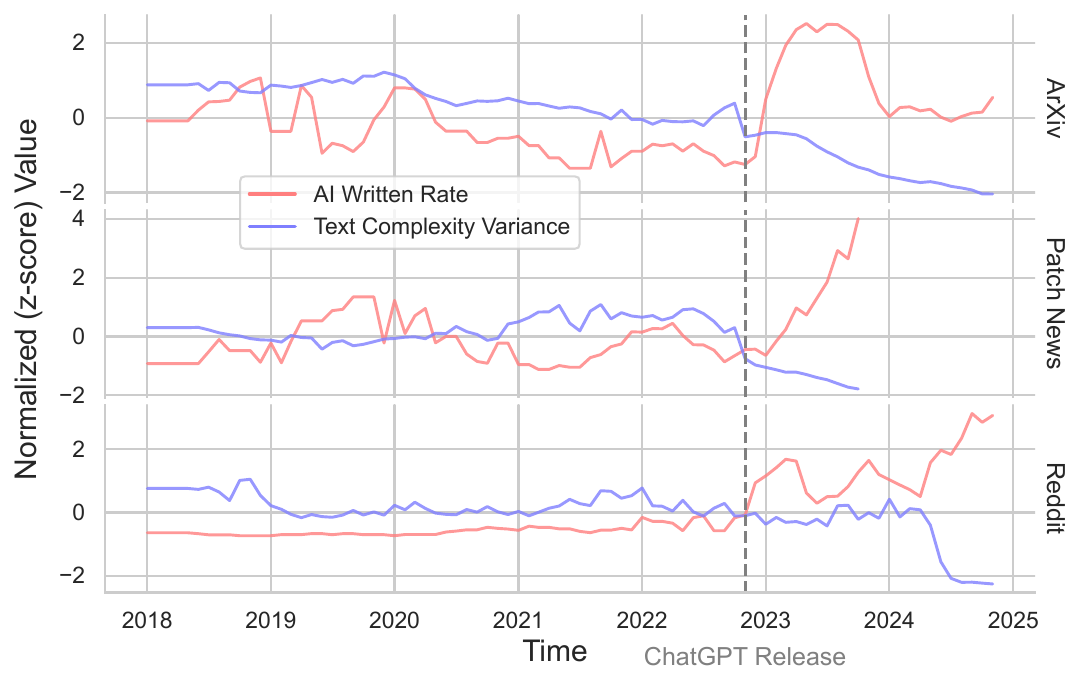}
    \caption{Trends in the variance of writing complexity and the attribution rate of texts as AI-generated, January 2018 to November 2024 (Reddit and arXiv) and November 2023 (Patch News) (monthly bins). Each panel corresponds to a dataset; within each panel, the red line shows the monthly proportion of texts attributed as AI-generated by Binoculars \citep{hans2024spotting} and the blue line shows the monthly variance of the aggregate writing-complexity score. Both series are normalized to $z$-scores within each dataset for visual comparability. The dashed vertical line marks ChatGPT's public release (November 2022). Number of monthly observations entering the time-series models: arXiv $= 83$, Reddit $= 83$, Patch News $= 71$ months. Sample sizes for the underlying texts are: arXiv $N=80{,}238$ papers, Patch News $N=379{,}583$ articles, Reddit $N=318{,}490$ stories.}
    \label{fig:complexity-trends}
\end{figure}

Our analyses revealed declines in the variance of writing complexity after the introduction of LLMs across all datasets, providing evidence for homogenization in writing styles (see \Cref{fig:complexity-trends}). Shock analyses demonstrated significant sustained post-launch reductions in writing complexity variance across the three datasets following the introduction of ChatGPT (see \Cref{tab:shock-analysis}). Similarly, Granger causality tests, and their nonlinear extension (GAM), suggested a predictive relationship between AI usage and changes in linguistic complexity variance for the Reddit and arXiv datasets (see \Cref{tab:granger_results}).

\begin{table}[h]
\caption{Shock analysis results assessing the effect of ChatGPT release on writing complexity variance. For each dataset, we fit a Discontinuous Growth Model (DGM) by Generalized Least Squares with an AR(1) error structure to the monthly variance series, and report the regression-coefficient estimate ($\beta$), standard error (SE), $t$-statistic, two-sided $p$-value (from the asymptotic Wald test of the regression coefficient), and 95\% confidence interval. Time captures the linear temporal trend, ONSET the immediate level shift at ChatGPT's launch, and POST the post-launch slope change.}
\begin{tabular}{llccccc}
\toprule
\textbf{Dataset} & \textbf{Term} & \textbf{Estimate ($\beta$)} & \textbf{SE} & \textbf{\emph{t}} & \textbf{\emph{p}} & \textbf{\emph{95\% CI}}\\
\midrule
\multirow{3}{*}{arXiv} 
 & Time & -0.0008 & 0.0000 & -14.6657 & $<$.001 & [-0.0010,  -0.0008]\\
 & ONSET & -0.0427 & 0.1100 & -0.3881 & .699 & [-0.2583, 0.1729] \\
 & POST & -0.0014 & 0.0002 & -6.6252 & $<$.001 & [-0.0019, -0.0010]\\
 \midrule
 \multirow{3}{*}{Patch News} 
 & Time & 0.0003 & 0.0001 & 3.3178 & .001 & [0.0001, 0.0005] \\
 & ONSET & -1.4054 & 0.2429 & -5.7859 & $<$.001 & [-1.8815, -0.9293] \\
 & POST & -0.0022 & 0.0011 & -2.0040 & .049 & [-0.0044, -0.0000] \\
\midrule
\multirow{3}{*}{Reddit} 
 & Time & -0.0002 & 0.0001 & -1.6375 & .106 & [-0.0005,  0.0000] \\
 & ONSET & 0.4161 & 0.2511 & 1.6568 & .102 & [-0.0761,  0.9083]\\
 & POST & -0.0021 & 0.0005 & -4.1796 & $<$.001 & [-0.0031, -0.0011] \\
\bottomrule
\end{tabular}
\label{tab:shock-analysis}
\begin{minipage}{\textwidth}
\footnotesize
\end{minipage}
\end{table}

\begin{table}[!htbp]
\caption{Temporal predictive relationship between AI usage and writing complexity variance. For each dataset, we report a linear Granger causality test ($F$-test on bivariate VAR coefficients, two-sided $p$-value) and a nonlinear extension based on generalized additive models (approximate $F$-test on smooth terms, with effective degrees of freedom edf, two-sided $p$-value). The optimal lag order was selected using the Akaike Information Criterion (AIC; \citealp{lutkepohl2005new,ivanov2005practitioner}).}
\centering
\begin{tabular}{lccc}
\toprule
\textbf{Dataset} & \textbf{Lag} & \textbf{Granger} & \textbf{GAM} \\
\midrule
arXiv & 5 & $F(5, 66) = 2.55,\; p = .036$ & $F = 2.45,\; \text{edf} = 7.4,\; p = .026$ \\
Reddit & 20 & $F(20, 21) = 3.75,\; p = .002$ & $F = 5.11,\; \text{edf} = 5.1,\; p < .001$ \\
Patch News & 20 & $F(20, 8) = 2.30,\; p = .114$ & $F = 1.82,\; \text{edf} = 3.5,\; p = .151$ \\
\bottomrule
\end{tabular}
\label{tab:granger_results}
\end{table}

In the arXiv dataset, shock analysis revealed a significant persistent decline in linguistic complexity variance following the introduction of ChatGPT (\(POST: \beta = -0.0014, p < .001\)). Notably, this decline further contributed to an existing downward trend in this measure (\(Time: \beta = -0.0008, p < .001\)). Furthermore, both the linear Granger test and the nonlinear GAM confirmed that AI usage significantly predicted reductions in the variance of writing complexity at the AIC-selected lag of five months (\(F(5, 66) = 2.55, p = .036\); GAM: \(F = 2.45, \text{edf} = 7.4, p = .026\)). These findings suggest that AI tools adoption is associated with convergence toward stylistic norms and reduced variability in scientific communication.

A similar pattern was observed in the Reddit dataset. Shock analysis revealed a significant and sustained post-launch decline in variance (\(POST: \beta = -0.0021, p < .001\)), indicating reduced linguistic complexity diversity after the introduction of ChatGPT. Both the linear Granger test and the nonlinear GAM confirmed that AI usage significantly predicted reductions in writing complexity variance at the AIC-selected lag of 20 months (\(F(20, 21) = 3.75, p = .002\); GAM: \(F = 5.11, \text{edf} = 5.1, p < .001\)). The longer delay observed in Reddit compared to those observed in the arXiv dataset suggests that users on this platform, reflecting a broad and diverse population, may adopt AI tools more gradually than those in domains with greater exposure to new technologies, like the authors of arXiv papers.

In the Patch News dataset, shock analysis identified significant changes in linguistic complexity variance, but neither the linear Granger test nor the nonlinear GAM were significant. Over time, there was a slight upward trend in complexity variance (\(Time: \beta = 0.0003, p = .001\)), but the onset of ChatGPT was associated with a marked decline (\(ONSET: \beta = -1.405, p < .001\)), and sustained post-launch effects were similarly associated with reduced variance (\(POST: \beta = -0.0022, p = .049\)). These findings suggest that while professional editorial workflows and institutional standards for consistency may buffer against the homogenizing effects of AI, subtle but measurable impacts on linguistic complexity variance remain. Taken together, these findings point to a temporal association between LLM adoption and declining linguistic diversity across diverse platforms and writing contexts.

\subsubsection*{Study 1b}
In this study, we simulate realistic writing-assisted scenarios in which humans draft original texts and then rely on LLMs to polish them using neutral prompts, allowing us to examine the role of LLMs in shaping writing style and their influence on linguistic diversity. To maximize the sensitivity of our test, we focus on the two domains where Study 1a revealed the strongest associations between AI-adoption rate and linguistic complexity variance: Reddit posts and arXiv papers. From each source, we randomly select 1,000 human-authored texts published before the release of ChatGPT and use three prominent LLMs, GPT-3.5 \citep{chatgpt}, Llama 3 70B \citep{dubey2024llama}, and Gemini Pro \citep{gemini}, to polish and rewrite them (see \Cref{method:controlled-llm-rewrites} for details). We then examine whether LLM rewriting is associated with measurable reductions in the variance of linguistic complexity by comparing the variance of linguistic complexity between original and LLM-rewritten texts using Levene's test (see \Cref{method:comparing-writing-complexity-variance-original-llm-written-texts} for details). At the same time, we evaluate the degree to which original meaning is preserved by computing the cosine similarity between embeddings of the original and rewritten texts (see \Cref{method:evaluation-semantic-preservation} for details).


\begin{figure}[!htbp]
    \centering
    \includegraphics[width=\textwidth]{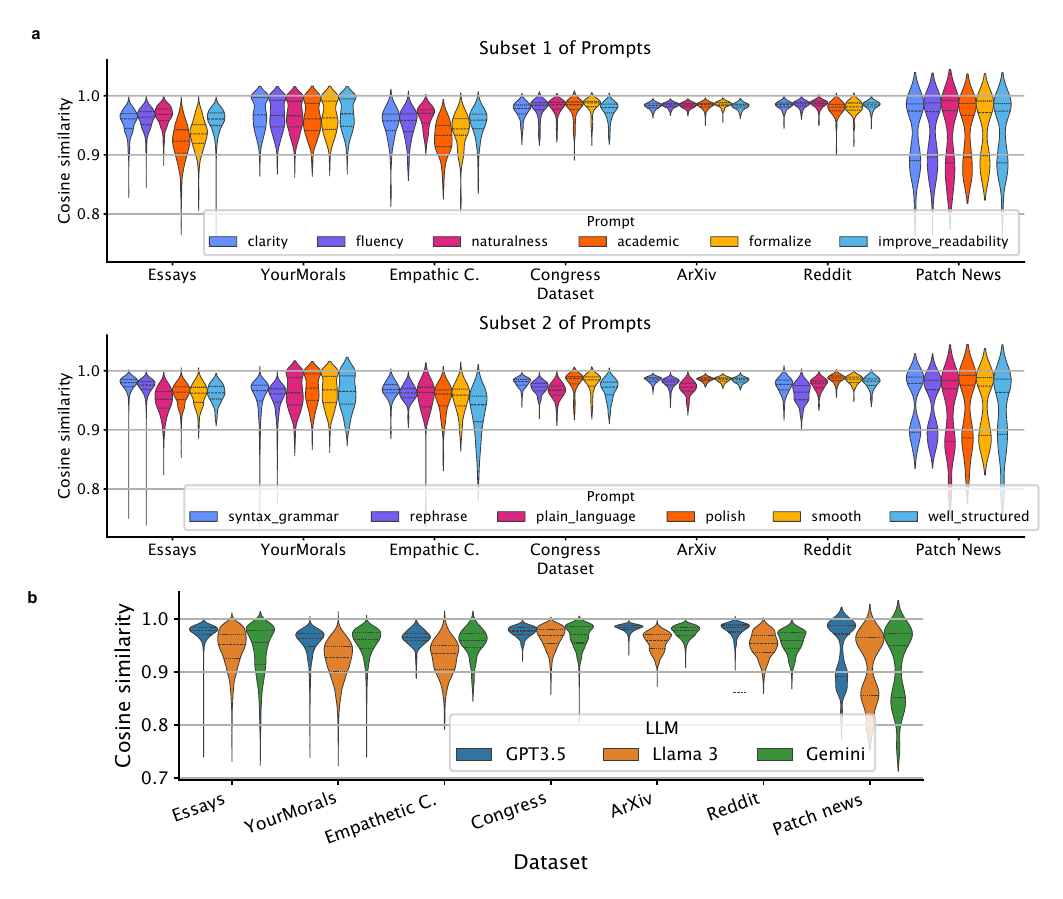}
   \extfigcaption{Semantic similarity between original and LLM-rewritten texts in different datasets, across (a) different prompts and (b) different LLMs. Each violin shows the kernel density of cosine similarities between OpenAI text-embedding-ada-002 embeddings of an original text and its LLM-rewritten counterpart, computed per text. The unit of analysis is one text; original and LLM-rewritten versions are paired by text (no separate control group is appropriate because each original text serves as its own paired baseline for its rewrites). Sample sizes per dataset: Essays n = 2,348, YourMorals n = 3,641, Empathic Conversations n = 711, Congress n = 710, ArXiv n = 1,000, Reddit n = 1,000, Patch News n = 1,000. Three thin horizontal lines inside each violin mark the 25th percentile, the median, and the 75th percentile of the underlying distribution.}
    \label{extended-fig:semantic-similarity}
\end{figure}

\begin{figure}[!htbp]
    \centering
    \exttabcaption{Variations of a sample text after LLM rewrites. Each row corresponds to a different prompt given to GPT-3.5, instructing it to modify the text in a particular way (e.g., rephrase, improve clarity, increase conciseness, formalise). The 'Original' row contains a paraphrased version of the source text, written by the authors to preserve scene, voice, and length while avoiding verbatim reproduction of third-party content; LLM rewrites in the remaining rows were produced under the procedure described in Methods.}
    \label{extended-tab:examples-of-llm-generated-data}
    \includegraphics[width=1\linewidth]{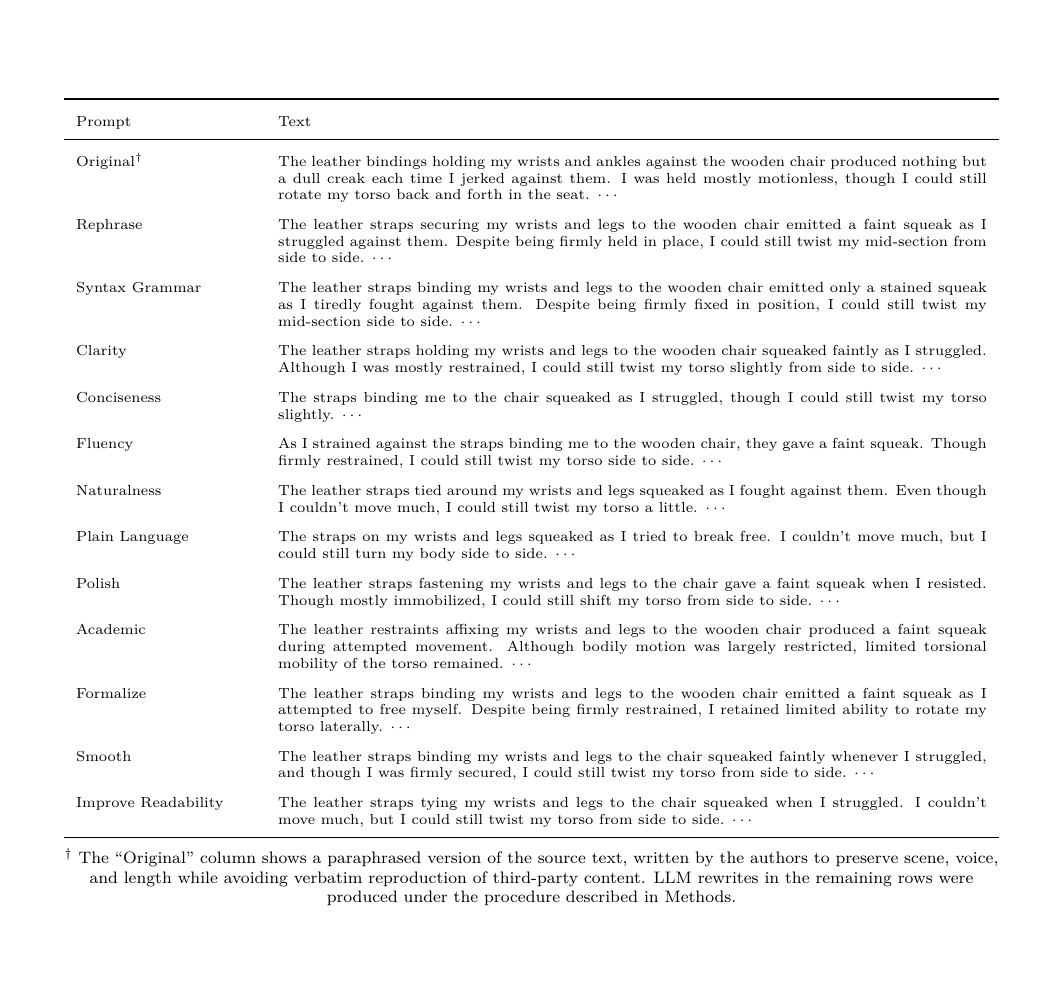}
\end{figure}

As expected, the semantic similarity between the original and LLM-rewritten texts remained high across all data sources, LLMs, and prompts. We found that 87\% of similarity scores were above 0.95 (see \Cref{extended-fig:semantic-similarity} and Appendix A in Supplementary Material, for details, and \Cref{extended-tab:examples-of-llm-generated-data} for examples).

\begin{table}[!htbp]
\caption{Changes in the variance of writing complexity between original and LLM-rewritten documents across datasets, LLMs, and prompts. Variance values before and after rewriting are displayed on either side of $\rightarrow$. R and SG refer to the Rephrase and Syntax\_Grammar prompts, respectively; full results across all prompts are provided in Appendix B in the Supplementary Material. Each cell reports, the variance change, Levene's test statistic $F$ with its two-sided exact $p$-value (median-centered robust form of Levene's test), and a non-parametric bootstrap 95\% confidence interval (10{,}000 resamples of the per-text complexity scores) on the ratio of variances $\sigma^{2}_{\text{orig}}/\sigma^{2}_{\text{LLM}}$. Values of the ratio $>1$ indicate reduced variance after rewriting; intervals excluding 1 indicate a reliable change. }
\label{tab:main-study-2-complexity-table}
\centering
\newcommand{\stk}[1]{\begin{tabular}[t]{@{}l@{}}#1\end{tabular}}
\resizebox{\textwidth}{!}{\begin{minipage}{\textwidth}
\begin{tabular}{llcc}
\toprule
 & \emph{Dataset} & \emph{arXiv} & \emph{Reddit}\\
LLM & Prompt & Variance; $F$; $p$; 95\% CI on $\sigma^{2}_{\text{orig}}/\sigma^{2}_{\text{LLM}}$ & Variance; $F$; $p$; 95\% CI on $\sigma^{2}_{\text{orig}}/\sigma^{2}_{\text{LLM}}$\\
\midrule
Gemini & R & \stk{0.0090 $\rightarrow$ 0.0071;\\ $F=5.062$; $p=.024$; 95\% CI [1.11, 1.42]} & \stk{0.0091 $\rightarrow$ 0.0076;\\ $F=2.126$; $p=.145$; 95\% CI [0.81, 1.49]} \\
 & SG & \stk{0.0090 $\rightarrow$ 0.0074;\\ $F=4.507$; $p=.034$; 95\% CI [1.21, 1.41]} & \stk{0.0091 $\rightarrow$ 0.0069;\\ $F=5.399$; $p=.020$; 95\% CI [1.14, 1.74]} \\
\midrule
GPT3.5 & R & \stk{0.0090 $\rightarrow$ 0.0070;\\ $F=5.031$; $p=.025$; 95\% CI [1.13, 1.63]} & \stk{0.0091 $\rightarrow$ 0.0063;\\ $F=5.978$; $p=.014$; 95\% CI [1.09, 1.97]} \\
 & SG & \stk{0.0090 $\rightarrow$ 0.0071;\\ $F=6.276$; $p=.012$; 95\% CI [1.07, 1.28]} & \stk{0.0091 $\rightarrow$ 0.0068;\\ $F=5.652$; $p=.017$; 95\% CI [1.21, 1.74]} \\
\midrule
Llama 3 & R & \stk{0.0090 $\rightarrow$ 0.0064;\\ $F=12.007$; $p<.001$; 95\% CI [1.13, 1.67]} & \stk{0.0091 $\rightarrow$ 0.0045;\\ $F=19.387$; $p<.001$; 95\% CI [1.29, 2.33]} \\
 & SG & \stk{0.0090 $\rightarrow$ 0.0111;\\ $F=3.377$; $p=.066$; 95\% CI [0.59, 0.85]} & \stk{0.0091 $\rightarrow$ 0.0069;\\ $F=3.952$; $p=.047$; 95\% CI [1.13, 1.65]} \\
\bottomrule
\end{tabular}
\end{minipage}}
\end{table}

Our analysis of LLMs' impact on writing complexity variance, as shown in \Cref{tab:main-study-2-complexity-table}, revealed a consistent trend: in both the Reddit and arXiv datasets, we observed significant reductions in variance when comparing original texts to their LLM-rewritten counterparts. These results, obtained under controlled conditions and across a range of rewriting prompts (see more prompts in Appendix B in the Supplementary Material), indicate that LLM rewriting is generally associated with reduced variance in linguistic complexity.

\subsubsection*{Discussion}

Across both observational and experimental analyses, Study 1 points to the same association: the use of LLMs in writing is linked to a homogenization of writing styles. In Study 1a, we showed that the launch of ChatGPT and its subsequent adoption across online platforms was associated with a marked reduction in the variance of linguistic complexity. Specifically, we found evidence of both an immediate shock, where the introduction of ChatGPT coincided with a sharp drop in variability, and a sustained effect, in which higher levels of LLM usage continued to predict reduced linguistic complexity variance over time. These temporal predictive relationships were confirmed by both linear and nonlinear methods. While our detection method prioritizes precision \citep{hans2024spotting}, showing greater sensitivity to texts that are largely LLM-written with less clear behavior on hybrid human–LLM texts, and thus offers a conservative estimate of LLM adoption, the consistency of the trends across datasets suggests that the true scale of homogenization may be even greater, consistent with findings in other domains \citep{liang2025quantifying,zhang2025generative}. Study 1b then replicated these results under controlled conditions by simulating realistic writing-assisted scenarios in which human-authored texts were polished or rephrased by different LLMs. The LLM-rewrites significantly reduced stylistic variance while preserving semantic content. Together, our findings demonstrate that LLM involvement is associated with compressed linguistic diversity while maintaining meaning,  indicating a homogenization effect of LLMs on people's writing styles.

\subsection*{Study 2}

In Study 1, we demonstrated that LLM adoption in writing is associated with a convergence towards more uniform language, illustrating the phenomenon of linguistic homogenization. This reduction in variation raises the critical question of how homogenization might impact the ability to discern and perceive individuals' personal traits from their writing, an area long investigated in sociolinguistics and psychology. As the ability to link lexical cues to personal traits (e.g., demographic attributes, personality dimensions, political orientation, or empathy) hinges on individual variability in language use, uniform linguistic patterns would obscure or weaken such associations. 

To assess how linguistic homogenization alters the identity cues embedded in language, we conduct a controlled experiment using texts from individuals whose personal traits are known from their responses to various psychological questionnaires. Each text is polished or rewritten by LLMs following the same procedure as in Study 1b. We then compare the predictability of authors' personal traits from original versus LLM-rewritten texts. For this purpose, we compile corpora that paired authors' writings with independently collected information about their traits, spanning \textbf{Demographics} (i.e., political affiliation, age, gender), \textbf{Personality}, \textbf{Empathy}, and \textbf{Moral values} (see \Cref{method:gathering-textual-data-alongside-authors-personal-traits} for dataset details). Classifiers are trained to predict these traits from linguistic features (e.g., TF–IDF or embeddings), with the traits serving as dependent variables ($DVs$) and the linguistic features as independent variables ($IVs$). We further analyze predicted class balances to capture whether LLM-rewritten texts are more often associated with a dominant class, and examine directional shifts to detect systematic biases (e.g., predictions skewing toward older, politically liberal, or more empathic profiles). Together, these analyses allow us to assess not only changes in predictive accuracy but also systematic ways in which LLM rewriting reshapes the linguistic signals tied to individual identity. Further details on features, models, and statistical tests are provided in \Cref{method:personal-traits-predictive-modeling-from-text}.

\begin{figure}[!htbp]
    \centering
    \includegraphics[width=\textwidth]{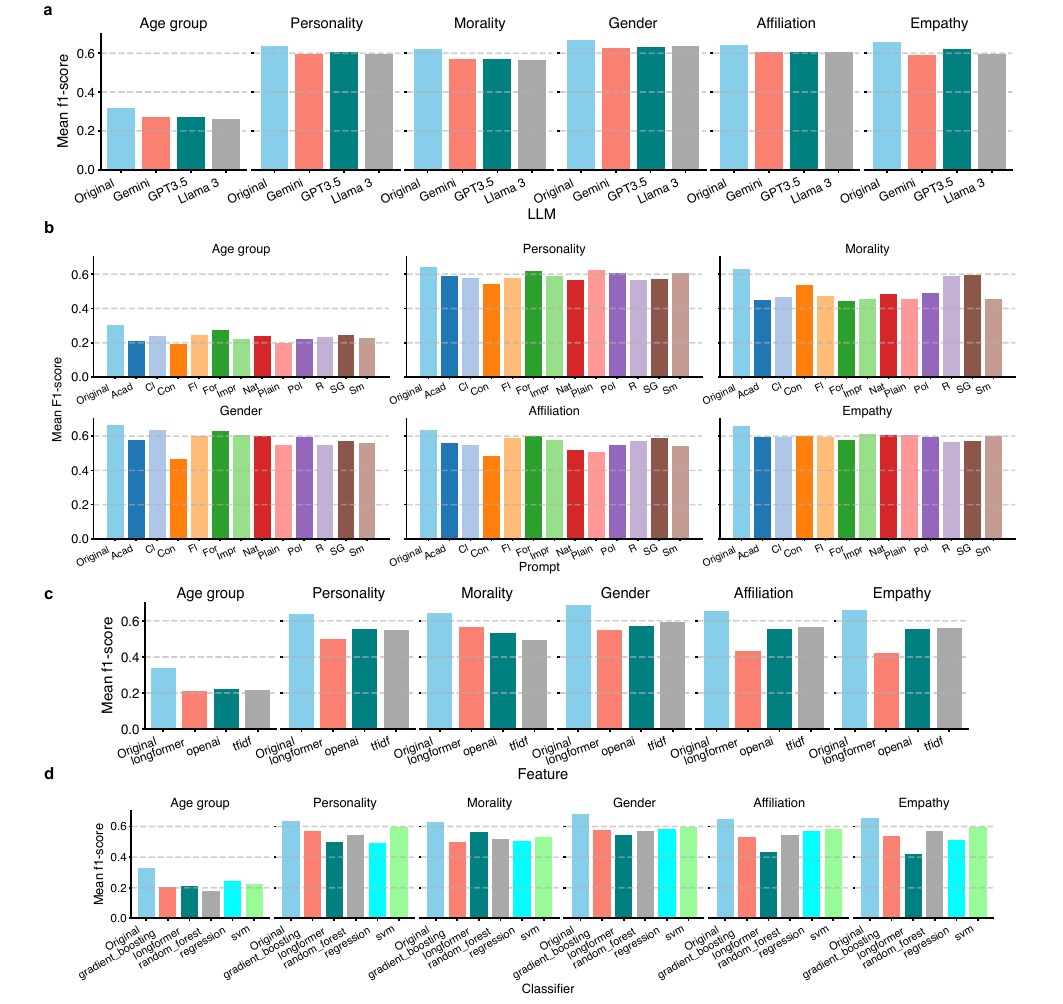}
   \extfigcaption{Predictive power (mean F1) of classifiers trained on the original texts and re-applied to the LLM-rewritten texts, across (a) different LLMs, (b) different rewriting prompts, (c) featurization techniques, and (d) classifier families. Each bar is the mean F1 across cross-validation folds and random seeds for the corresponding configuration. Abbreviations of prompt types: SG = Syntax\_Grammar, R = Rephrase, Cl = Clarity, Con = Conciseness, Fl = Fluency, Nat = Naturalness, Plain = Plain language, Pol = Polish, Acad = Academic, For = Formalize, Sm = Smooth, Impr = Improve readability. Sample sizes per personal-trait group: Age group n = 805 texts, Personality n = 2,348, Empathy n = 711, Gender n = 710, Morality n = 3,641, Affiliation n = 710.}
    \label{extfig:predictive-power-across-variables}
\end{figure}

\begin{table}[!htbp]
\caption{Performance comparison between trained personal-traits' classifiers on the original (OG) and the LLM-rewritten (LLM) texts. Two-sided paired Student's $t$-tests assess the statistical significance of differences in predictive power ($F_1$). All six tests were significant with $p<.001$ and remain significant after Bonferroni correction for the six trait comparisons. Cohen's $d$ effect sizes indicate the magnitude of these differences ($0.2<|d|<0.5$: small, $0.5<|d|<0.8$: medium; $|d|>0.8$: large, \citealp{cohen}); 95\% CIs are reported on the mean difference $\Delta_{\text{Mean}\,F_1}$. RB stands for the random baseline.}
\label{tab:ttest_res}
\centering
\resizebox{\textwidth}{!}{\begin{minipage}{\textwidth}
\begin{tabular}{lcccccrrrrc}
\toprule
\text{Trait}  & Mean $F_{1OG}$ & $\text{Mean $F_1$}_{\text{LLM}}$ & $\Delta_{\text{Mean } F_1 }$& 95\% CI & SE & \emph{t} & df & \emph{d} & RB\\
\midrule
Age group & .351 & .260 & .091 &[.079, .102] & 0.006 & 15.397 & 804 & 0.543 &  .244\\

Empathy & .657 & .603  & .054&[.050, .057] & 0.002 & 32.158 & 6556 &  0.397 & .541\\

Personality & .658 & .602 & .056&[.052, .060] & 0.002 & 30.046 & 7652 &  0.343 &  .514\\

Gender & .694 & .623 & .072&[.063, .080] & 0.004 & 15.965 & 1371 &  0.431 & .495\\

Morality & .640 & .578 & .062 &[.057, .066] & 0.002 & 25.601 & 5195 &  0.355 &  .521\\

Affiliation & .664 & .591 & .073&[.064, .082] & 0.005 & 15.707 & 1314 &  0.433 & .490\\
\bottomrule
\end{tabular}
\end{minipage}}
\end{table}

\Cref{tab:ttest_res} displays the average differences between $F_1$ scores of classifiers predicting authors' personal traits on the original texts versus the LLM-rewritten texts. We observed an average 6\% decline in predictive accuracy (absolute $F_1$ score) of classifiers, but interestingly, performances on the LLM-rewritten texts did not fall below the random baseline. Dimension-specific results for multidimensional traits, e.g., ``personality'' that includes the five Big Five traits (openness, conscientiousness, extraversion, agreeableness, neuroticism), are available in Appendix D.2 in the Supplementary Material. We observed similar trends across all LLMs, rewriting prompts, or linguistic features utilized for classification (see \Cref{extfig:predictive-power-across-variables}).

Paired $t$-tests indicated that the performance of classifiers on the LLM-rewritten texts was significantly lower compared to performance on the original texts across different traits ($p<.001$; small-to-moderate Cohen's $d$ effect sizes). In particular, among different personal traits, the linguistic markers of age groups were affected the most by LLMs, where the average performance of classifiers on the LLM-rewritten texts dropped to $F_1=.260$, compared to $F_1=.351$ on the original texts (moderate effect size).  

\begin{figure}[!htbp]
    \centering
    \includegraphics[width=\textwidth]{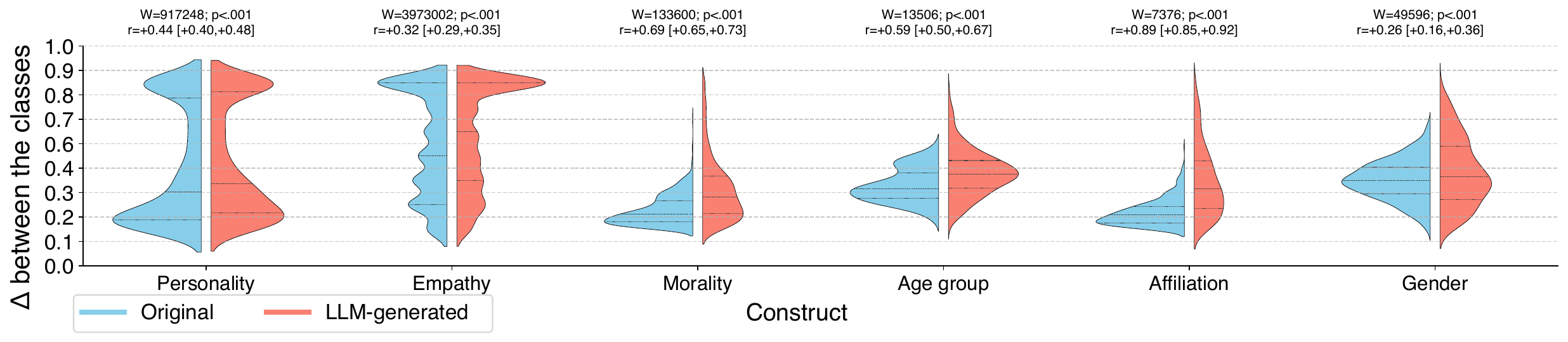}
   \caption{Distributions of $\Delta$ (a proxy for imbalances between predicted class frequencies on the original and LLM-rewritten texts) per construct, for GPT-3.5-rewritten texts. Violins show the kernel density of $\Delta$ for original (blue) vs.\ LLM-rewritten (salmon) texts; the unit of analysis is one classifier-training run, paired within-run between the two conditions. Number of paired runs: Personality $n=3{,}152$, Empathy $n=7{,}719$, Morality $n=1{,}325$, Age group $n=364$, Affiliation $n=518$, Gender $n=524$. Three thin horizontal lines inside each violin mark the 25th percentile, the median, and the 75th percentile of the underlying distribution. Annotations on top of each violin report the Wilcoxon signed-rank test statistic $W$ (two-sided, paired), the exact $p$-value, and the rank-biserial correlation $r$ with its 95\% bootstrap confidence interval (2{,}000 resamples). }
    \label{fig:difference-between-predictions-homogeneity}
\end{figure}

\begin{figure}[!htbp]
    \centering
    \includegraphics[width=\textwidth]{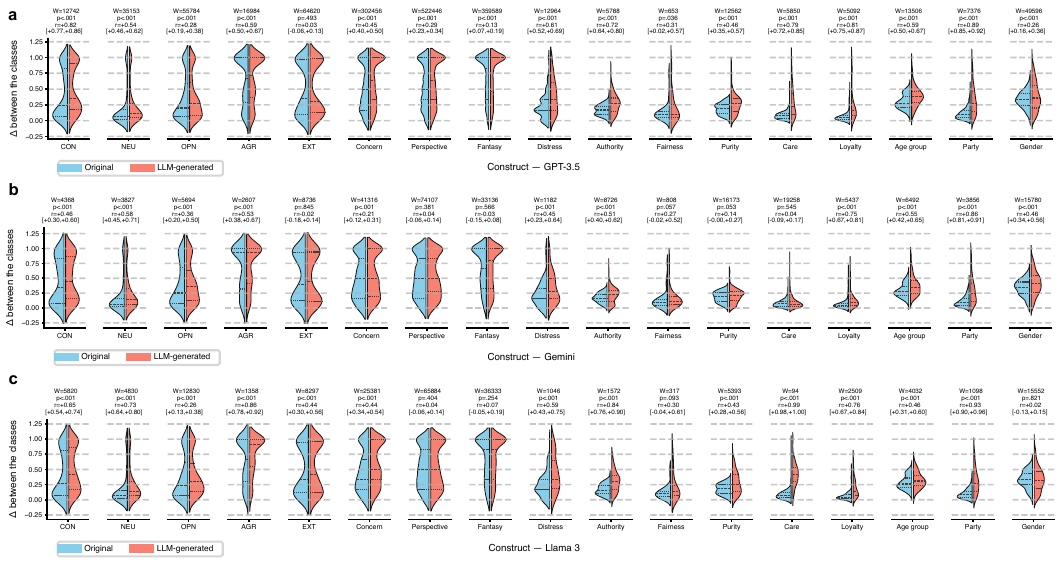}
   \extfigcaption{Distributions of Delta (a proxy for imbalances between predicted class frequencies on the original and LLM-rewritten texts) across different sub-dimensions of personal traits, using (a) GPT-3.5, (b) Gemini, and (c) Llama 3. Violins show the kernel density of Delta for original (blue) vs. LLM-rewritten (salmon) texts; the unit of analysis is one classifier-training run, paired within-run between the two conditions. Number of paired runs (n) per sub-dimension varies by LLM and is reported above each violin. Three thin horizontal lines inside each violin mark the 25th percentile, the median, and the 75th percentile of the underlying distribution. Annotations on top of each violin report the Wilcoxon signed-rank test statistic W (two-sided, paired), the exact p-value, and the rank-biserial correlation r with its 95\% bootstrap confidence interval (2,000 resamples). }
    \label{extfig:delta-across-llms}
\end{figure}

Next, we analyzed the distributions of $\Delta$ (i.e., a proxy for imbalances
between the predicted classes) for both the original and LLM-rewritten texts
to assess whether LLMs systematically skew predictions toward a particular
class (e.g., predicting ``older'' more often than ``younger,'' or ``low
empathy'' more often than ``high empathy''), or whether LLMs' impact on texts
lead to predictions that are more random and evenly distributed across
classes. Two-sided Wilcoxon signed-rank tests confirmed that $\Delta$ for
LLM-rewritten texts was significantly larger than for the original texts
across all six constructs (focusing on GPT-3.5, matching
\Cref{fig:difference-between-predictions-homogeneity}): Personality
($W=917{,}248$, $p<.001$, rank-biserial $r=.44$, 95\% CI $[.40, .48]$,
$n=3{,}152$ paired runs); Empathy ($W=3{,}973{,}002$, $p<.001$, $r=.32$,
95\% CI $[.29, .35]$, $n=7{,}719$); Morality ($W=133{,}600$, $p<.001$,
$r=.69$, 95\% CI $[.65, .73]$, $n=1{,}325$); Age group ($W=13{,}506$,
$p<.001$, $r=.59$, 95\% CI $[.50, .67]$, $n=364$); Affiliation ($W=7{,}376$,
$p<.001$, $r=.89$, 95\% CI $[.85, .92]$, $n=518$); and Gender
($W=49{,}596$, $p<.001$, $r=.26$, 95\% CI $[.16, .36]$, $n=524$). This
suggests that LLM-rewrites not only diminish the distinctiveness of
linguistic styles linked to different personal traits, but also
disproportionately shift texts toward resembling authors with particular
traits, and push classifiers to favor one dominant class (e.g., predicting
``older'' more often across age groups, or ``higher morality'' across moral
values). Notably, this trend persists across different LLMs and various
dimensions of the analyzed constructs (see \Cref{extfig:delta-across-llms}).

\begin{figure}[!htbp]
    \centering
    \includegraphics[width=\textwidth]{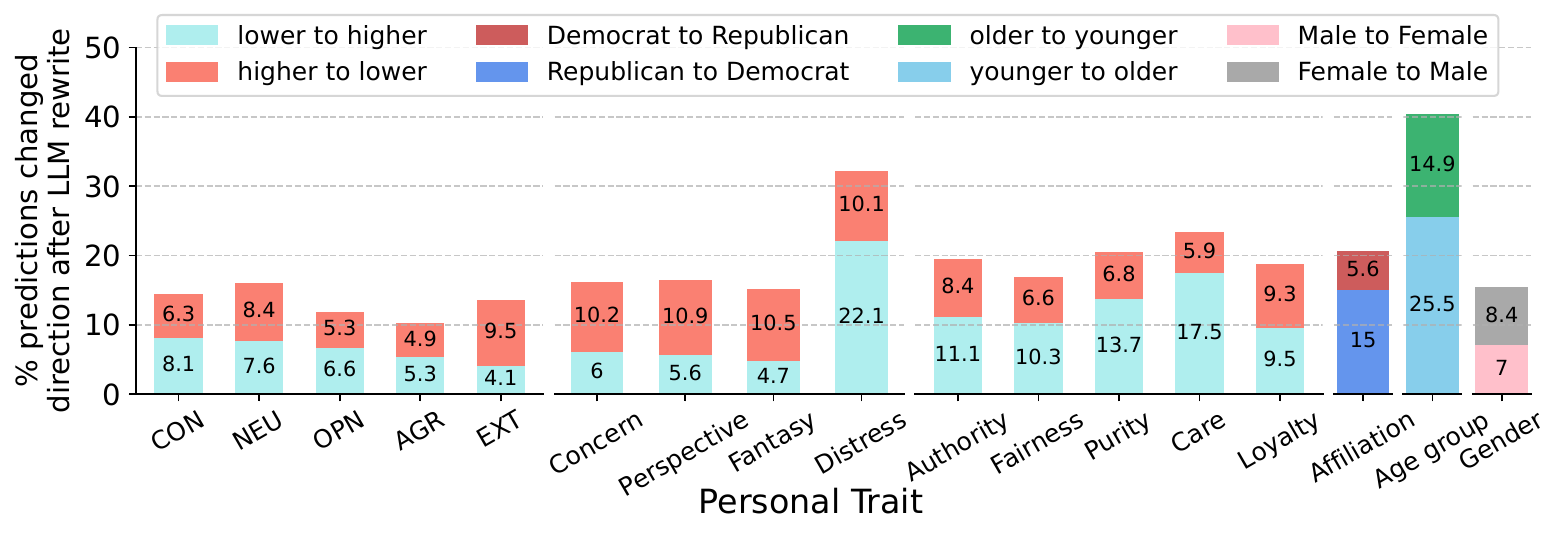}
    \caption{Percentage of original texts with correct author-attribute predictions that changed after LLM rewriting, grouped by the direction of change in predictions, focusing on the LLM-rewritten texts generated by GPT-3.5. For each personal-trait sub-dimension, only the subset of texts that were correctly classified on the original was retained as the denominator; the bars show, within that subset, the percentage of texts whose post-rewrite prediction shifted in each labeled direction (e.g.\ ``higher to lower'', ``older to younger''). The unit of aggregation is one paired classifier-training run; number of paired runs per sub-dimension ($n$): CON $632$, NEU $611$, OPN $652$, AGR $612$, EXT $645$, Empathic Concern $2{,}237$, Perspective Taking $2{,}601$, Fantasy $2{,}486$, Personal Distress $395$, Authority $288$, Care $334$, Fairness $62$, Loyalty $334$, Purity $307$, Affiliation $518$, Age group $364$, Gender $524$.}
    \label{fig:difference-in-predictions-agg}
\end{figure}

\begin{figure}[!htbp]
    \centering
    \includegraphics[width=\textwidth]{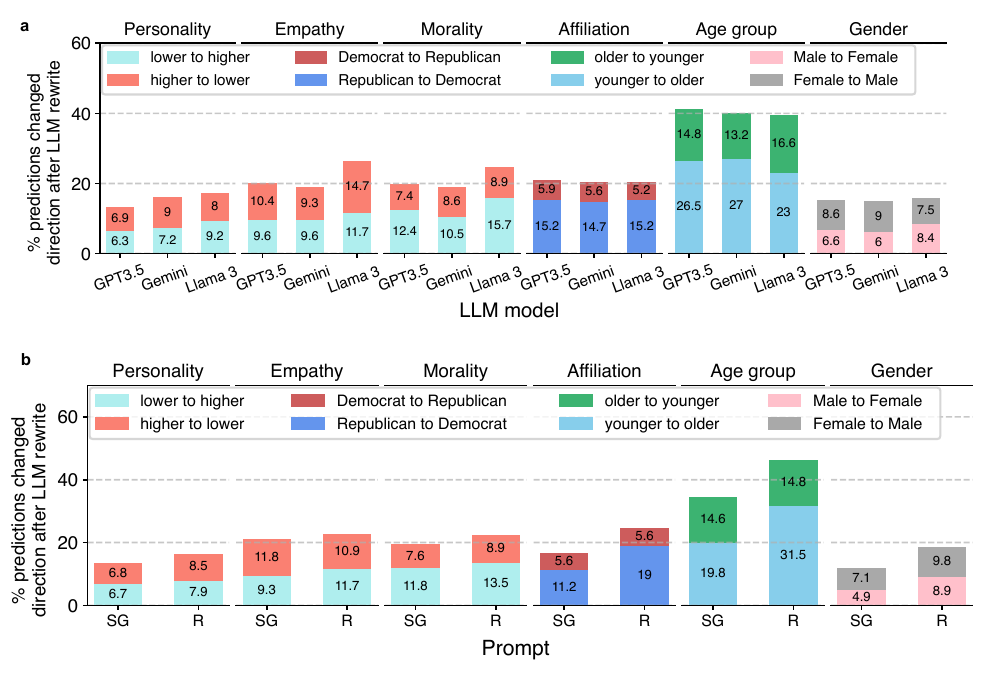}
   \extfigcaption{Percentage of original texts with correct author-attribute predictions that changed after LLM rewriting, grouped by the direction of change in predictions. (a) Results broken down by LLM (GPT-3.5, Gemini, Llama 3). (b) Results broken down by rewriting prompt: SG (Syntax\_Grammar) and R (Rephrase); similar patterns were observed under the other prompts as well. For each personal-trait sub-dimension and each (LLM or prompt), the denominator is the subset of original texts that were correctly classified, and the bar shows the percentage of those whose post-rewrite prediction shifted in each labelled direction. Within each panel the denominator is further partitioned across the three LLMs or the two prompt families.}
    \label{extended-fig:systematic-bias-in-predictions}
\end{figure}

\Cref{fig:difference-in-predictions-agg} illustrates changes in classifier predictions between original and LLM-rewritten texts, based on the proportion of predictions that shifted from correct to incorrect and the direction of those shifts across different personal traits. These analyses reveal distinct trends in certain personal traits becoming more prominently predicted than others when LLMs were involved in writing. For example, LLM-rewritten texts tended to reduce the likelihood that classifiers would identify authors as having higher empathy, leading to systematically lower predicted empathy scores (except for the personal distress dimension) compared to the ground-truth author labels ($t(6698) = 20.18, p <.001, \text{Cohen's}\;d = 0.35,\;95\%\;CI\;[0.31, 0.39]$ [small effect size]; paired $t$-test). At the same time, LLM-rewritten texts pushed classifiers to more frequently predict that authors had stronger moral concerns ($t(3051) = 21.25, p <.001, d = 0.61,\;95\%\;CI\;[0.55, 0.67]$ [medium effect size]), and belonged to older age groups ($t(597) = 16.84, p <.001, d = 1.09,\;95\%\;CI\;[0.94, 1.24]$ [large effect size]). We also observed that LLM-rewrites shifted classifier predictions in personality domains, making texts more likely to be classified as written by authors high in openness ($t(1074) = 6.50, p < .001, d = 0.21,\;95\%\;CI\;[0.15, 0.27]$ [small effect size]) or agreeableness ($t(1005) = 9.05, p < .001, d = 0.27,\;95\%\;CI\;[0.22, 0.32]$ [small effect size]), and low in extraversion ($t(1054) = 15.14, p < .001, d = 0.52,\;95\%\;CI\;[0.45, 0.59]$ [medium effect size]) compared to the actual author labels. There were also distinct shifts in political affiliation predictions: LLM-rewritten texts led classifiers to more often label authors as Democrats, even when the true author was Republican ($t(867) = 24.89, p <.001, d = 1.35,\;95\%\;CI\;[1.20, 1.50]$ [large effect size]). Finally, gender predictions were skewed, with classifiers biased toward predicting ``Male'' more often than ``Female'' when LLMs were involved in the writing ($t(613) = 5.68, p <.001, d = 0.37,\;95\%\;CI\;[0.24, 0.50]$ [small effect size]) (see Appendix D.3 in the Supplementary Material for more details). Taken together, these systematic shifts appeared consistently across different LLMs and prompts (see \Cref{extended-fig:systematic-bias-in-predictions}).

\subsubsection*{Discussion}

Diving deeper into the social implications of the homogenization effect of LLMs on people's writing styles, our analysis reveals that while LLM-rewritten texts preserve the core meaning, they reduce the ability to accurately infer authors' personal traits from their texts, because the distinctive linguistic cues that normally signal these traits are weakened or erased. These linguistic cues are double-edged: while their loss may be seen positively in reducing risks of misuse (e.g., surveillance or discrimination; \citealp{kosinski2013private,tufekci2014engineering,garg2018word}), it also diminishes their positive value for studying identity, personality, and mental health. More concerning, the changes are not random. Instead, LLM rewriting systematically reshapes writing in ways that push classifier predictions toward certain profiles of identity, thereby introducing new biases rather than simply erasing signals. Across datasets, we observe a consistent bias in LLM-rewritten texts, with writing styles often aligning with older, male, politically liberal individuals and exhibiting positive moral valence and lower empathy. Furthermore, while we observe consistent and measurable declines in the predictive power of linguistic cues, classifiers continue to perform well above chance, suggesting that LLMs do not completely erase the linguistic cues that convey personal traits. To further understand these shifts and to assess how LLMs alter the associations between linguistic features and personal traits, Study 3 takes a more granular approach, analyzing fine-grained lexical cues typically linked to these traits and examining how their presence and distributions change in LLM-rewritten texts.

\subsection*{Study 3}

In Study 2, we demonstrated that LLMs not only homogenize writing styles but also systematically remove and alter linguistic features that reflect authors' personal traits. We also found that homogenization was not accompanied by a complete loss of predictive power, suggesting that LLMs do not entirely eliminate the informative linguistic cues linked to personal traits. To explore the mechanisms behind these observations, and to understand why predictive power declines, we adopt a theory-driven approach \citep{kennedy2022handbook}, drawing on studies that highlight fine-grained linguistic cues associated with personal traits \citep[e.g.,][]{kennedy2021moral, hirsh2009personality, mehl2006personality}, and tracking well-documented links between specific linguistic markers (e.g., pronoun use, emotion words, moral language) and personal traits in both original and LLM-rewritten texts.

Using the same corpora from Study 2, which pair texts with independently collected author personal traits, we first replicate well-established associations between fine-grained lexical features (commonly used in psychology and social science, such as LIWC \citep{pennebaker2022linguistic} and the Moral Foundations Dictionary 2 (MFD2) \citep{frimer2019moral}) and personal traits in the original documents. We then examine how LLM rewriting alters these associations by comparing feature–trait correlations in original and rewritten texts. This approach provides two key insights: first, it pinpoints how homogenization operates at the level of specific linguistic markers (e.g., cues tied to morality, empathy, or personality); second, it highlights what is at stake for sociolinguistics and related disciplines, where such markers are routinely used to study how language reflects social and psychological variation. Further details on the lexical features, feature quantification, and statistical analyses are provided in \Cref{method:association-between-linguistic-features-personal-traits}.


\begin{table*}[htb]
    \caption{Highlighted correlations between various linguistic cues and authors' personal traits (see Appendix F in Supplementary Material for exact correlation coefficients and additional categories) on original and LLM-rewritten texts. \checkmark and \xmark~indicate significant and insignificant correlations, respectively.}
    \label{tab:checkmarks-personality-morality-empathy}
    \centering
    \begin{minipage}{.5\linewidth}
        \centering
        \resizebox{\textwidth}{!}{%

\centering
\begin{tabular}{llcc}
\toprule
\multicolumn{4}{c}{\textbf{Personality}} \\
Dimension & Linguistic Cue & Original & LLM-rewritten \\
\hline
OPN  &  i & \checkmark & \checkmark \\ 
&  swear & \checkmark & \checkmark \\ 
&  BigWords & \checkmark & \xmark\\ 
   \hline
CON & anger emo & \checkmark & \checkmark \\ 
&  swear & \checkmark & \checkmark \\ 
& neg emo & \checkmark & \xmark \\
  \hline
EXT & Social & \checkmark & \checkmark \\ 
&  pos emo & \checkmark & \checkmark \\
&  neg emo & \checkmark & \xmark\\
&  pronoun & \checkmark & \xmark \\
  \hline
AGR & pos emo & \checkmark & \checkmark \\
& swear & \checkmark & \checkmark\\
& neg emo & \checkmark & \xmark\\ 
& disgust & \checkmark & \xmark\\
\hline
NEU & pronoun & \checkmark & \checkmark \\ 
& emotion & \checkmark &  \checkmark \\[0pt]\\
\toprule\multicolumn{4}{c}{\textbf{Demographics}} \\
Dimension & Linguistic Cue & Original & LLM-rewritten \\
\hline
Gender & anx emo & \checkmark & \checkmark \\
& neg emo & \checkmark & \xmark \\
& Social & \checkmark & \checkmark \\
\hline
Age & we & \checkmark & \checkmark \\
& Focus Future & \checkmark & \xmark \\
\hline
Affiliation & adverb & \checkmark & \checkmark \\
& emo anx & \checkmark & \xmark \\
\bottomrule
\end{tabular}
        }
    \end{minipage}%
    \begin{minipage}{.5\linewidth}
        \centering
        \resizebox{\textwidth}{!}{%
            \centering
\begin{tabular}{llcc}
\toprule
\multicolumn{4}{c}{\textbf{Morality}} \\
Dimension & Linguistic Cue & Original & LLM-rewritten \\
\hline
Fairness  & religion & \checkmark & \xmark \\ 
& authority & \checkmark & \xmark \\
\hline
Loyalty & affiliation & \checkmark & \checkmark \\ 
&  family & \checkmark  & \checkmark \\ 
& friend & \checkmark & \xmark \\
& care & \checkmark & \xmark \\
  \hline
Authority &  friend & \checkmark & \xmark \\
& Social & \checkmark & \checkmark \\
   \hline
Purity &  family & \checkmark & \checkmark \\ 
&  religion & \checkmark & \checkmark  \\
& sanctity & \checkmark & \xmark \\
   \hline
Care & authority & \checkmark & \xmark \\ 
&  affiliation & \checkmark & \checkmark \\[11pt]\\
\toprule\multicolumn{4}{c}{\textbf{Empathy}} \\
Dimension & Linguistic Cue & Original & LLM-rewritten \\
\hline
PD  &  affect & \checkmark & \xmark \\ 
    &  differ & \checkmark & \checkmark \\ 
\hline
EC & we & \checkmark & \checkmark \\
& cogproc & \checkmark & \checkmark \\ 
\hline
FS & neg emo & \checkmark & \checkmark  \\
& empathy & \checkmark & \xmark \\
\hline
PT & we & \checkmark & \checkmark  \\
& pronoun & \checkmark & \checkmark \\ 
&  cogproc & \checkmark &\checkmark \\
\bottomrule
\end{tabular}
        }
    \end{minipage} 
\end{table*}

\Cref{tab:checkmarks-personality-morality-empathy} contains a simplified illustration of some important associations between lexical categories and various personal traits of authors in original and LLM-rewritten texts (see Appendix F in Supplementary Material for exact correlation coefficients and additional categories). 

Our analysis of the original texts successfully replicated many well-documented findings. For example, consistent with previous research, male authors used fewer social ($t(708) = -4.77, p < .001, d = -0.36$, 95\% CI on $d$ $[-0.51, -0.21]$) and anxiety-related words ($t(708) = -3.46, p < .001, d = -0.26$, 95\% CI on $d$ $[-0.41, -0.11]$) than female authors \citep{ishikawa2015gender}. Similarly, higher extroversion (EXT) was linked to greater use of positive emotions ($r(2346) = .096, p < .001$, 95\% CI $[.056, .136]$) and social words ($r(2346) = .075, p = .003$, 95\% CI $[.034, .115]$) \citep{chen2020meta}, and individuals scoring high on loyalty used more family-related words ($r(3639) = .114, p < .001$, 95\% CI $[.082, .146]$) \citep{LiTomasello}. We also observed links between openness (OPN) and the use of complex words (BigWords) ($r(2346) = .089, p < .001$, 95\% CI $[.049, .129]$), and between fairness and religion-related words ($r(3639) = -.063, p = .001$, 95\% CI $[-.096, -.031]$).

Comparing associations found in the original and LLM-rewritten documents, we found that many of these associations were washed away by LLM involvement. For example, the relationship between gender and negative emotion words, which was well-established in the original texts ($t(708) = -3.30, p = .001, d = -0.25$, 95\% CI on $d$ $[-0.40, -0.10]$), was largely erased in the LLM-rewritten texts ($t(708) = -1.20, p = .230, d = -0.09$, 95\% CI on $d$ $[-0.24, 0.06]$). Similarly, the relationship between extraversion (EXT) and the use of pronouns ($r(2346) = .064, p = .020$, 95\% CI $[.024, .104]$ in the original; $r(2316) = .055, p = .080$, 95\% CI $[-.005, .115]$ after LLM rewriting), loyalty and the use of friend-related words ($r(3639) = .075, p < .001$, 95\% CI $[.042, .107]$ in the original; $r(3640) = .060, p = .080$, 95\% CI $[-.005, .115]$ after LLM rewriting), and age with words focusing on the future ($r(708) = -0.09, p = .017$, 95\% CI $[-0.16, -0.02]$ in the original; $r(708) = -.037, p = .322$, 95\% CI $[-.110, .036]$ after LLM rewriting) were weakened in the LLM-rewritten texts. In contrast, some associations, such as the connection between neuroticism (NEU) and negative emotion words ($r(2346) = .190, p < .001$, 95\% CI $[.151, .229]$ in the original; $r(2316) = .199, p < .001$, 95\% CI $[.159, .238]$ after LLM rewriting), the relationship between purity and religion-related words ($r(3639) = .206, p < .001$, 95\% CI $[.175, .237]$ in the original; $r(3640) = .211, p < .001$, 95\% CI $[.180, .242]$ after LLM rewriting), and the use of social words in relation to gender ($t(708) = -3.66, p < .001, d = -0.28$, 95\% CI on $d$ $[-0.43, -0.13]$ after LLM rewriting), persisted. Taken together, these results suggest that the reduction in predictive power documented in Study 2 arises from the selective erosion of well-established linguistic markers rather than a uniform loss across all cues. For sociolinguistics, this implies that the impact of LLMs is highly complex: which associations disappear and which remain depends on both the personal trait in question and the specific lexical cues known to be associated with it.

\subsubsection*{Discussion}

To better understand the reduction in predictive power observed in Study 2, where classifiers lost some but not all of their ability to predict personal traits, we examined how LLMs alter well-established lexical cues linked to personal traits. Our findings reveal that this decline is reflected in the disruption of established associations between linguistic features and personal traits. While some informative linguistic markers remained intact, preserving a degree of predictive power, many key associations were erased. This selective disruption in well-established associations suggests that LLMs are not merely introducing random noise but are systematically altering the connection between language and identity. This carries important consequences for fields like psychology and sociolinguistics, as long-standing methods that rely on lexical cues to study how language encodes social and psychological variation, may become less reliable when texts have been altered by LLMs. If the links between cues and traits are systematically reshaped, researchers risk drawing distorted conclusions about identity, group differences, or cultural patterns. Thus, our results highlight that LLM-driven homogenization not only affects predictive modeling but also challenges the broader enterprise of using language as a window into human diversity.

\section*{General Discussion}

The words we speak paint a portrait of who we are, where we belong, and how we connect with the world around us, solidifying language's central place in the human story. This richness, stemming from the inherent diversity of human expression, allows us to glean insights into individual and societal characteristics. However, the increasing popularity of LLMs as writing assistants raises a key question: are these tools eroding the distinctiveness of individual expression? Given that LLMs generate text by predicting statistically probable continuations based on massive training corpora, their widespread adoption may risk homogenizing language, potentially obscuring the distinctive nature of language and limiting its ability to convey individuality.

To investigate this potential homogenization effect, we conducted three studies examining different aspects of LLMs' influence on language. First, we demonstrated a decline in the variance of writing complexity across multiple online platforms after the introduction of ChatGPT in November 2022, suggesting a trend toward more uniform writing styles. Controlled experiments then demonstrated that when LLMs are used to polish or rewrite text (e.g., rephrasing, improving clarity, or simplifying), the resulting outputs are associated with reduced variation in writing complexity, providing additional descriptive evidence of linguistic homogenization by LLMs. We established that LLMs systematically alter textual cues signaling authors' personal traits by homogenizing stylistic elements, reducing the overall predictive power of text for these traits, and shifting the perceptions about the authors toward specific demographic and personality profiles while preserving the core meaning of the original texts. Finally, we demonstrated that associations between specific, well-established lexical features and personal traits weaken in LLM-assisted texts, providing further insights into how LLMs strip cues about personal traits. Collectively, these findings highlight a transformation in the nature of individual expression, diminishing the identity signals embedded in language and leading to a loss of linguistic uniqueness when LLMs are integrated into the writing process.

The trend of homogenization of writing styles appeared across formal and informal settings, including Reddit, academic papers, and news articles, though its impact varied by domain. Specifically, this effect was more pronounced in Reddit, a platform traditionally rich in personal and creative expression, and in academic writing, where AI tools are associated with more standardized manuscripts but risk narrowing the range of narrative styles in scientific communication. In contrast, homogenization in news articles was less pronounced: while greater LLM usage coincided with a sustained reduction in the variance of writing complexity, no predictive relationship between AI adoption and homogenization was found, possibly reflecting existing editorial guidelines limiting direct AI adoption. However, subtle reductions in writing complexity variance suggest that AI is still associated with shifts in journalistic style, perhaps not through direct adoption but through its influence on broader writing norms. Even without direct use, increased exposure to and imitation of AI-generated content might gradually reshape stylistic standards, contributing to the erosion of linguistic diversity even in domains where LLMs are not the primary driver of homogenization. Together, these findings highlight how LLMs may reinforce and reshape stylistic norms through both direct use and indirect exposure, potentially suppressing individual expression in domain-specific ways.

Our investigation into LLM-driven homogenization of linguistic indicators of personal traits revealed that while the core meaning of texts is preserved, stylistic features are systematically altered or removed. This suggests that LLMs primarily modify stylistic elements, which, while arguably secondary to content, crucially convey information about the identity and background of authors. As a result, classifiers struggled to accurately predict authors' personal traits. Notably, LLM-rewritten texts tended to align with authors having traits commonly associated with older, male, politically liberal individuals and exhibited a more positive moral valence with lower empathy. These patterns align with existing research on LLMs' inherent biases \citep[e.g., ][]{navigli2023biases,atari2023humans,wang2024large,rozado2024political,pan2023llms,abdurahman2023perils}, and suggest that even when primed with text containing cues to an author's traits, LLMs tend to reinforce personality profiles consistent with their pre-existing biases. These shifts in linguistic expression can have far-reaching implications across domains such as social media, professional communication, clinical psychology, and mental healthcare, where diminished linguistic diversity may undermine efforts to recognize and leverage individual differences in expression \citep{nguyen2016computational,garg2018word,rude2004language,eyigoz2020linguistic}. Similarly, in hiring, applicants with AI-assisted writings whose texts conform to implicit stylistic expectations may be favored against those whose original writing is more distinctive but less conventionally polished, potentially exacerbating existing barriers to equity.
More broadly, declines in language's ability to authentically reflect and differentiate aspects of identity may result in a loss of diversity in human expression, subtly reshaping norms in ways that prioritize conformity over individuality.

Furthermore, we showed that the reductions observed in the predictive power of personal trait classifiers following LLM-rewrites do not occur randomly but rather arise from the systematic erosion of well-established linguistic markers associated with those traits. This matters because such linguistic cues are widely relied upon in both academic and industry settings \citep{tausczik2010psychological}, and their disruption could lead to misleading insights and decisions. If traits like neuroticism or values like loyalty lose their consistent linguistic signals, inferences about these characteristics may become unreliable. Similarly, in contexts like therapy sessions or customer service interactions, the masking or distortion of key emotional and empathetic cues by LLMs could result in misinterpretations with serious consequences.

The selective suppression of some associations and the preservation of others raises an important question: what factors determine which markers are retained and which are disrupted? One possibility is that LLMs implicitly recognize and retain certain linguistic distinctions while filtering out others due to biases in training data and alignment procedures. Since AI training prioritizes factual accuracy and may lack explicit mechanisms to preserve individual writing styles, such as mapping stylistic features to verified personality or demographic labels, LLMs may fail to reliably maintain fine-grained markers of identity. Further investigation into this process could offer valuable insights into how LLMs encode and alter linguistic features.

While the current research focused on writing complexity, and LLMs' direct effect on text and capturing thereof, both behavioral factors, such as individual adoption patterns, stylistic preferences, the degree to which users selectively accept or modify LLM suggestions, and dynamics between content creators and audiences, and non-behavioral factors, such as platform- or organization-level policies, can influence the form and extent of LLM-driven homogenization. Understanding how these elements interact is crucial for assessing the broader trend toward linguistic uniformity. This trend is complex and difficult to model, as LLMs and their training processes shape which features lose variability. However, the trend persists due to the nature of LLM training: during pre-training, LLMs optimize for the most statistically likely continuations, reinforcing dominant patterns present in the training data. In the alignment phase, outputs are further shaped by the preferences of a limited set of annotators, emphasizing prevailing linguistic norms while filtering out deviations. These processes inherently constrain the diversity of generated language, leading to a systematic narrowing of expression that reflects both data-driven patterns and the biases embedded in alignment. Importantly, trends will only intensify as new LLM-influenced text is added to online repositories, which, in turn, will be used for further training, perpetuating a feedback loop that progressively reduces linguistic variability.

Lastly, given the inextricable connection between language and thought \citep{evans2009myth, sapir1956language}; the homogenization of language through the widespread adoption of LLMs may have far-reaching long-term implications for cognitive diversity. By potentially restricting the range or quality of linguistic resources at our disposal, LLMs risk limiting variability in linguistic structures, reducing the cognitive flexibility and creativity that allows us to process and innovate upon complex ideas. This narrowing of linguistic diversity could, at an extreme, ``narrow the range of thought'' \citep[][p.70]{georgeorwell}, as the diminishing array of cognitive possibilities constrains our ability to think in diverse ways.

In conclusion, while LLMs offer powerful tools for enhancing clarity and accessibility in writing, our findings point to an alarming consequence: the erosion of linguistic distinctiveness. Across diverse sources and controlled experiments where LLMs were used to polish or rephrase text, the same trend toward homogenization appeared in each case. As AI-mediated communication becomes increasingly prevalent, it is imperative that we ensure these tools enhance, rather than extinguish, the rich tapestry of human linguistic diversity.

\backmatter


\section{Methods}
\label{sec:methods}

This work did not involve the collection of new data from human participants. All text, demographic, personality, empathy, and moral-foundation data analyzed here were obtained from previously published or publicly accessible corpora, each collected and shared by the original investigators under the institutional ethical approvals reported in the corresponding original publications and summarized in the Data availability statement. The brief in-house semantic-similarity sanity check described in \Cref{method:evaluation-semantic-preservation} was a quality-control exercise carried out by members of the research effort and does not constitute research with human participants. Consistent with this and with the precedent for secondary-analysis studies of public text corpora recently published in this journal (e.g., \citealp{liang2025quantifying}), no additional institutional ethical approval was required for the present analysis.

Unless otherwise specified, all statistical tests reported in this manuscript are two-tailed. The studies reported here were not preregistered.

\subsection{Gathering Diverse Online User-generated Content}
\label{method:gathering-diverse-online-user-generated-content}

To examine the impact of LLMs on linguistic diversity, we select data sources that are diverse, authored by individuals rather than organizations, and allow for a variety of writing styles without strict external constraints. These sources are characterized by their broad range of topics and independence from objectives that might inadvertently homogenize language, such as enforcing correctness or adherence to a uniform style. Our dataset comprises three distinct sources:  

1. \textbf{Reddit (r/WritingPrompts):} This subreddit features creative stories written by users as comments in response to open-ended prompts, encouraging originality with minimal external constraints. The lack of strict rules ensures that these stories reflect individual creativity and linguistic diversity. We collected $N = 318,490$ stories posted between January 2018 and November 2024, downloaded from Project Arctic Shift (\url{https://github.com/ArthurHeitmann/arctic_shift}). To focus on substantive user-generated narratives, we excluded short replies and feedback comments by filtering out texts with fewer than 200 words.

2. \textbf{Patch News:} \url{Patch.com} provides localized news from communities across the United States (488 counties in 50 states), written by a large number (N = 19,584) of contributors. 
Unlike larger outlets, Patch articles are less regulated, allowing reporters' voices to shine through without the imposition of unified organizational standards. We gathered $N = 379,583$ articles from January 2018 to November 2023, focusing on the main body of the text and removing advertisements.

3. \textbf{arXiv Preprints:} arXiv is an online repository where researchers routinely share their manuscripts, ensuring immediate access to their work. This allows us to capture papers closer to when the research was conducted, avoiding delays and edits associated with formal publication processes. Additionally, arXiv encompasses a wide range of subjects, promoting diversity in topics and writing styles. We specifically focused on papers from the \emph{Computer Science; Linguistics (CL)} and \emph{Computer Science; Vision (CV)} categories, as these fields have strong traditions of posting preprints on arXiv. We analyze abstracts from the publicly available metadata of arXiv papers \citep{arxiv_org_submitters_2024}, specifically those posted between January 2018 and November 2024. This subset consists of $N = 80,238$ papers with $N = 161,265$ contributing authors.

\subsection{Detection of AI-written Texts}
\label{method:detection-ai-written-text}

To identify whether a piece of text is AI-generated, we use \emph{Binoculars} \citep{hans2024spotting}, a tool that relies on normalized perplexity, a measure of how predictable a text is to a language model. AI-generated texts generally exhibit lower perplexity due to the inherent uniformity of language models, which optimize for the most probable word sequences. By normalizing perplexity, Binoculars effectively distinguishes AI-written content, even when prompts are designed to produce less typical outputs. Note that the method is premised on detecting LLM-written text, and its performance on texts partially written by LLMs and humans is yet to be systematically evaluated. The tool has demonstrated a high ($> 0.95$) true positive rate while maintaining a low ($< 0.01$) false positive rate, detecting AI-written content across benchmarks, particularly for text types similar to those in Study 1 (i.e., news articles, academic papers, and creative writing), compared to other baselines \citep{tufts2024practical,dugan2024raid,intellectualead2025ai_detectors}. Using the threshold recommended in its codebase, we classify texts as human- or AI-written based on their normalized perplexity scores. We chose Binoculars because of its high precision compared to other zero-shot baselines \citep{verma2023ghostbuster,gptzero2023more}, particularly in the domains that are the focus of our studies. Other approaches, such as the methods proposed by \citet{liang2025quantifying}, rely on supervised training with labeled corpora of human- and LLM-written texts \citep{liang2025quantifying,zhan2023g3detector,zellers2019defending}. While these methods are effective, they ultimately capture similar distributional differences between original and LLM-generated writing. In contrast, Binoculars does not require additional training data, making it more directly applicable across domains and more consistent with our goal of analyzing diverse real-world texts.

We also acknowledge limitations in how Binoculars recognizes LLM-written text. Because Binoculars relies on normalized perplexity, texts that mix human and LLM writing may increasingly appear fully AI-generated as the share of LLM content grows. While this means the method can capture partially co-written texts, our study does not attempt to disentangle such cases, and detection is not our main contribution. More broadly, it is unclear how people adopt LLM outputs in practice or what exact features detectors rely on, since these models are trained on corpora without explicit rule-based mechanisms. We, therefore, use Binoculars as a practical, high-precision tool for identifying AI-generated text, while our focus is on the surge of such texts and their implications of homogenization rather than on advancing detection methods.

\subsection{Calculation of Variance in Writing Complexity}
\label{method:calculation-variance-writing-complexity}

We define \(D = \{d_1, d_2, \dots, d_N\}\) as a collection of documents, where each document \(d_i\) is associated with a creation date, \(\text{date}(d_i)\). To quantify the complexity of these documents, we rely on the following set of features that capture lexical and syntactic dimensions of textual complexity:  
\begin{itemize}  
    \item \textbf{Vocabulary Simpson Index (\(\text{Simpson}(d_i)\); 
    \citealp{simpson1949measurement}}): Adapted from ecology, this measure quantifies vocabulary richness and complexity by calculating the probability that two randomly selected terms from \(d_i\) are identical. By capturing lexical uniformity, it offers insights into the diversity of word usage.
    \item \textbf{Vocabulary Shannon Entropy (\(\text{Shannon}(d_i)\); \citealp{shannon1948mathematical})}: Quantifies the unpredictability or richness of word usage in \(d_i\).  
    \item \textbf{Average Dependency Link Length (\(\text{DepLength}(d_i)\); \citealp{gibson1998linguistic})}: Evaluates syntactic complexity by calculating the average length of syntactic dependencies in the text.  
    \item \textbf{Type-Token Ratio (\(\text{TTR}(d_i)\); \citealp{johnson1944studies})}: Captures lexical diversity by dividing the number of unique words (types) by the total word count (tokens).  
    \item \textbf{Hapax Legomena (\(\text{Hapax}(d_i)\); \citealp{mardaga2012hapax})}: Reflects lexical novelty by counting words in \(d_i\) that occur only once.  
\end{itemize}  

To study temporal trends in the variance of writing complexity, we compute the variance of each complexity feature (e.g., \(\text{Simpson}\), \(\text{TTR}\)), aggregated monthly (\(m\)), denoted as \(\sigma^2_{\text{(feature, m)}}\). High \(\sigma^2_{\text{(feature, m)}}\) suggests heterogeneity in writing styles or complexity levels among documents, whereas low \(\sigma^2_{\text{(feature, m)}}\) indicates homogeneity, where documents tend to exhibit similar patterns in the respective feature. We use months as the temporal unit because more granular bins, such as days or weeks, lead to excessive variability and too few data points per interval to be meaningful, while broader bins, such as years, provide too few observations for robust temporal analyses. Months, therefore, offer a balance that reduces noise while preserving sufficient resolution for our shock analyses and Granger causality tests.

\begin{figure}[!htbp]
    \centering
    \includegraphics[width=0.92\textwidth]{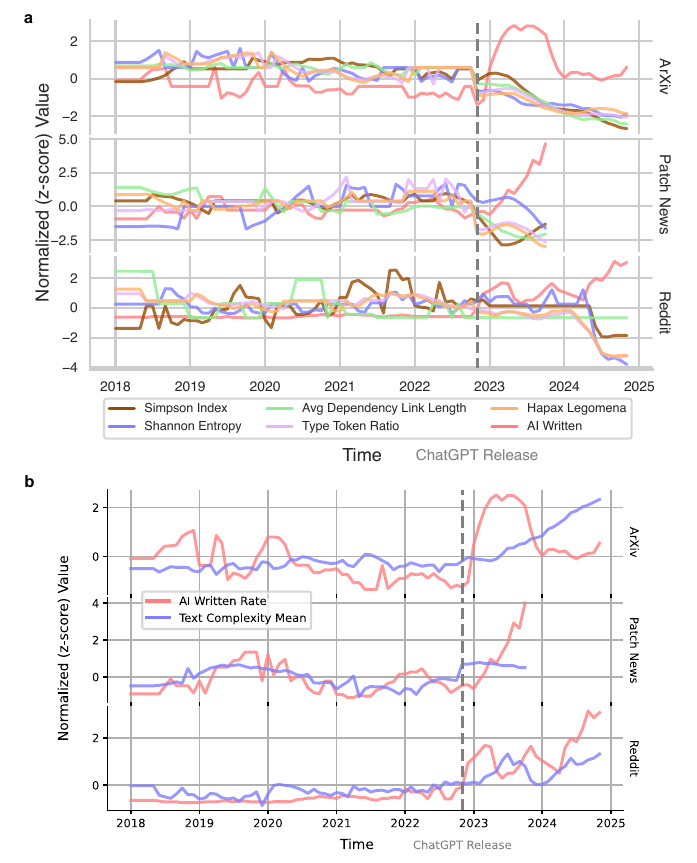}
   \extfigcaption{Homogenization of linguistic complexity following the introduction of ChatGPT. (a) Monthly variance of each complexity-related feature (Vocabulary Simpson Index, Vocabulary Shannon Entropy, Average Dependency Link Length, Type-Token Ratio, Hapax Legomena) shrinks after ChatGPT's release, alongside an increase in the AI-attribution rate. (b) Monthly mean of the same features increases over the same period. Each panel reports one dataset; series are normalized to z-scores within each dataset for visual comparability, and the dashed vertical line marks ChatGPT's public release (November 2022). Sample sizes for the underlying texts: ArXiv N = 80,238 papers, Patch News N = 379,583 articles, Reddit N = 318,490 stories.}
    \label{extended-fig:homogenization}
\end{figure}

As a general indicator of the overall trend and to facilitate easier interpretation of results, we further aggregate the variance of all complexity-related features into a single averaged measure, denoted as $\bar{\sigma^2}_m$. Cronbach's alpha \citep{cronbach1951coefficient} values indicate acceptable reliability and internal consistency for the composite measure in arXiv (\(\alpha = .965\); 95\% CI: [.951, .976]), Patch News (\(\alpha = .722\); 95\% CI: [.604, .813]) as well as Reddit (\(\alpha = .708\); 95\% CI: [.595, .796]), and support the use of a composite measure to represent the overall trend in writing complexity variance across time. We present the trends for all complexity-related features individually in \Cref{extended-fig:homogenization}.

\subsection{Relationship Between AI Usage and Variance in Writing Complexity}  
\label{method:relationship-ai-usage-variance-writing-complexity}

To investigate how the introduction and adoption of LLMs are associated with changes in the variance in writing complexity, we employ a two-tiered analytical approach. First, we conduct a shock analysis using a Discontinuous Growth Model \citep{singer2003applied} to assess whether trends in writing complexity variance shift after the introduction of LLMs. This approach provides insight into whether LLMs have impacted overall patterns in writing complexity variance. Next, we conduct a Granger Causality analysis \citep{linden2015conducting} to study the relationship between AI usage rates and writing complexity variance, complemented by a nonlinear extension based on generalized additive models \citep{bell1996non} to capture potential nonlinear temporal dependencies. This allows us to test the more specific hypothesis that AI usage can predict changes in the variance of writing complexity.

\paragraph{Shock analysis.}  

Taking ChatGPT's launch on November 30, 2022, which garnered widespread public attention and arguably signaled the beginning of widespread LLM adoption as a marker, we perform a Discontinuous Growth Model (DGM) analysis \citep{singer2003applied} to assess whether this event triggered a statistically significant shift in the variance of writing complexity. To ensure robust estimation of the ChatGPT's launch effects, we employ a Generalized Least Squares (GLS) method with an autoregressive correlation structure ($AR$(1)) to account for potential autocorrelations within our time series data. Specifically, we fit the following model:

\[
\bar{\sigma^2_m} \sim \text{Time}_m + \text{ONSET}_m + \text{POST}_m,
\]

\noindent where \(\text{Time}_m\) is a continuous variable representing time (in months), \(\text{ONSET}_m\) is a binary variable indicating whether the date falls before or after the launch (\(\text{ONSET}_m = 0\) for pre-launch and \(\text{ONSET}_m = 1\) for post-launch), and \(\text{POST}_m\) represents the elapsed time from launch for post-launch observations (\(\text{POST}_m = 0\) for pre-launch dates). In this model, \(\text{Time}_m\) captures the overall temporal trend in writing complexity variance, \(\text{ONSET}_m\) identifies any immediate changes following the introduction of LLMs, and \(\text{POST}_m\) evaluates potential longer-term effects that persist beyond the initial launch. The GLS framework with an AR(1) error structure does not require the raw outcome series to be normally distributed; inference on the regression coefficients is based on the asymptotic normality of the GLS estimator under standard regularity conditions for time-series regression. We additionally inspected the residuals of each fitted DGM and verified that residual skewness and excess kurtosis fell within commonly used thresholds (\(|\text{skewness}| \leq 2\), \(|\text{excess kurtosis}| \leq 7\); \citealp{west1995structural,curran1996robustness}).

\paragraph{Granger causality test.}  
We further investigate whether LLMs have a sustained impact on writing complexity variance beyond a simple initial shock or pulse effect by employing a Granger causality test \citep{linden2015conducting}, and a nonlinear extension based on generalized additive models (GAMs; \citealp{bell1996non}). These tests determine if past values of one time series (i.e., AI usage) can predict future values of another (i.e., linguistic complexity variance), indicating a temporal predictive relationship. While AI usage is estimated using text unpredictability (perplexity) and linguistic complexity includes measures influenced by unpredictability (e.g., Shannon entropy), our analysis specifically examines the relationship between the variance in linguistic complexity (dependent variable) and another construct based on the overall magnitude of unpredictability or complexity (AI usage rate; independent variable). This distinction mitigates concerns about direct dependence between the two. Following standard econometric practice \citep{lutkepohl2005new,ivanov2005practitioner}, we select the optimal lag order using the Akaike Information Criterion (AIC) applied to a bivariate vector autoregressive model with lag orders ranging from 1 to 24, which identifies the lag that best captures the temporal dynamics of the system while penalizing model complexity. This yields a single test per dataset, avoiding the multiple comparison concerns associated with testing across many lag orders \citep{bruns2019lag}.

Before applying the Granger causality tests, we verify the stationarity assumption, ensuring that the time series' statistical properties, such as mean and variance, remain constant over time, by performing the Augmented Dickey-Fuller (ADF) test \citep{dickey1979distribution} on all variables. Non-stationary variables indicated by the ADF test are differenced accordingly to achieve stationarity \citep{box2015time}. After first-order differencing, all time series met the stationarity assumption, allowing us to proceed with the analysis.

The null hypothesis of the Granger causality test posits that the lagged values of the predictor variable do not improve the prediction of the target variable beyond what can be achieved with the target variable's own historical values. The Granger causality test is conducted using an \(F\)-statistic to compare the fit of two models: one including only the lagged values of the target variable and the other incorporating both the lagged values of the target and predictor variables. A significant \(F\)-statistic indicates that the predictor variable provides additional explanatory power, rejecting the null hypothesis and supporting a temporal predictive relationship. We analyze complexity variance trends separately for each dataset to identify domain-specific patterns and lag effects.

To capture potential nonlinear temporal dependencies that the standard linear Granger test may miss, we additionally employ a nonlinear extension based on generalized additive models \citep[GAMs;][]{bell1996non,wood2017generalized}. This approach replaces the linear predictors in the Granger framework with smooth nonlinear functions estimated via penalized regression splines, while testing the same underlying hypothesis: whether past values of AI usage improve the prediction of future writing complexity variance beyond what its own history provides. The restricted model includes only smooth functions of the target variable's lagged values, while the full model additionally includes smooth functions of the predictor's lagged values. The two models are compared using an approximate $F$-test from an analysis of deviance \citep{wood2017generalized}, with effective degrees of freedom (edf) reflecting the complexity of the estimated smooth terms \citep{hastie1990generalized}. Following standard reporting conventions for GAM model comparisons \citep[e.g.,][]{wieling2018analyzing}, we report the $F$-statistic, edf, and $p$-value for each test. A significant test statistic indicates that the predictor variable's lagged values provide additional explanatory power beyond what the target variable's own history captures, consistent with the logic of the linear Granger test but without imposing a linear functional form.

\subsection{Controlled LLM Rewrites}
\label{method:controlled-llm-rewrites}

To examine how LLMs alter textual properties under realistic writing-assistant use, we simulate scenarios in which humans draft original texts and then use LLMs to polish them. We prompted three widely used models, GPT-3.5 \citep{chatgpt}, Llama 3 70B \citep{dubey2024llama}, and Gemini Pro \citep{gemini}, to rewrite human-authored texts without emphasizing any particular stylistic features. The goal was to capture neutral rewriting effects rather than targeted transformations.  

For this purpose, we employed a set of prompts designed to encourage revision in naturalistic ways, such as improving grammar, readability, or fluency, while preserving meaning. Examples include:  

\begin{itemize}
    \item \textbf{Syntax\_Grammar}: ``Rewrite the following text using the best syntax and grammar, and other revisions that are necessary: \{TEXT\}''
    \item \textbf{Rephrase}: ``Rephrase the following text: \{TEXT\}''
    \item \textbf{Clarity}: ``Rewrite the following text to improve clarity and readability: \{TEXT\}''
    \item \textbf{Conciseness}: ``Rewrite the following text to make it more concise while preserving meaning: \{TEXT\}''
    \item \textbf{Fluency}: ``Rewrite the following text to improve overall fluency and flow: \{TEXT\}''
    \item \textbf{Naturalness}: ``Rewrite the following text so it reads more naturally in English: \{TEXT\}''
    \item \textbf{Plain\_Language}: ``Rewrite the following text in clear and straightforward language: \{TEXT\}''
    \item \textbf{Polish}: ``Polish the following text to improve its overall quality without changing its meaning: \{TEXT\}''
    \item \textbf{Academic}: ``Rewrite the following text in a more polished academic style without altering meaning: \{TEXT\}''
    \item \textbf{Formalize}: ``Rewrite the following text in a more formal way without adding or removing content: \{TEXT\}''
    \item \textbf{Smooth}: ``Rewrite the following text to ensure smooth transitions and consistent flow: \{TEXT\}''
    \item \textbf{Improve\_Readability}: ``Rewrite the following text to enhance readability: \{TEXT\}''
\end{itemize}

\subsection{Evaluating Semantic Preservation}
\label{method:evaluation-semantic-preservation}

To assess how LLMs affect the semantic content of the original writings, we calculate the cosine similarity between embeddings representing semantic content of the original texts and their LLM-rewritten counterparts. We use OpenAI's \texttt{text-embedding-ada-002} model to generate high-dimensional embeddings for both the original and rewritten texts because of their excellent performance on sentence similarity tasks \citep{greene2024embedding,conneau2018senteval} and near-human annotation capabilities \citep{gilardi2023chatgpt,alizadeh2023open}. The cosine similarity between these embeddings provides a quantitative measure of how much of the meaning of the text is preserved. A higher cosine similarity score indicates that the LLM-rewrites retained the texts' original meaning, while a lower score suggests that significant changes were introduced.

To complement our quantitative analyses, we performed a brief in-house semantic-similarity sanity check on 20 randomly selected text pairs from the Reddit dataset. Four of the authors (three male and one female; adult graduate students in computer science or psychology, all fluent in English) independently rated each original-vs-rewritten pair using the following rubric: 1.~Different (information conveyed is mostly different); 2.~Somewhat similar (information conveyed has notable differences); 3.~Very similar (information conveyed only has minor differences). We then analyzed these ratings using descriptive statistics and computed Gwet's AC1 \citep{gwet2008computing} as a measure of inter-rater agreement. The check indicated a high degree of similarity, with a mean rating of $2.97 / 3\;(SD = 0.07)$ and a high inter-rater agreement of Gwet's AC1~=~0.947 (95\% CI: [0.868, 1.000]). This step was an internal quality-control exercise carried out by members of the author team and did not constitute the collection of new data from research participants; no compensation was provided.

\subsection{Comparing Writing Complexity Variance of Original vs. LLM-Rewritten Texts}
\label{method:comparing-writing-complexity-variance-original-llm-written-texts}

In Study 1b, we simulate realistic writing-assisted scenarios in which humans draft original texts and then rely on LLMs to polish them using neutral prompts, to examine LLMs' influence on linguistic diversity. To maximize the sensitivity of our test, we focus on the two domains where Study 1a revealed the strongest associations between AI-adoption rate and linguistic complexity variance: Reddit posts and arXiv papers. From each source, we randomly select 1,000 human-authored texts published before the release of ChatGPT and use three prominent LLMs, GPT-3.5 \citep{chatgpt}, Llama 3 70B \citep{dubey2024llama}, and Gemini Pro \citep{gemini}, to polish and rewrite them. A post-hoc power analysis confirmed that this sample size was sufficient: even the smallest observed variance difference (0.0015) was detected with 0.832 power. 

To analyze how LLMs affect the variability in writing complexity, we adopt the same complexity-related features and methodology as Study 1a. We compute an aggregate writing complexity score for each document by averaging five key features: Vocabulary Simpson Index, Vocabulary Shannon Entropy, Average Dependency Link Length, Type-Token Ratio, and Hapax Legomena. We then calculate the variance of this aggregate score for both the original and LLM-rewritten documents, allowing us to examine the overall difference in writing complexity variance between the two sets. To statistically compare these variances, we utilize Levene's test, which determines if the variances of two or more groups are significantly different. This approach helps us assess whether LLMs homogenize writing complexity and lead to a more uniform writing style. Levene's test is designed to be robust to non-normality of the underlying distributions \citep{levene1960robust}, making it appropriate for our setting.

\subsection{Gathering Textual Data Alongside Their Authors' Personal Traits}
\label{method:gathering-textual-data-alongside-authors-personal-traits}

We use diverse datasets comprising written texts (e.g., essays, social media posts, and personal reflections) from authors with known personal traits, including demographic information (e.g., gender, age, political affiliation), personality, dispositional empathy, and moral values. These datasets also serve as the source material for generating LLM-modified texts. Overall, the texts included in these datasets are mostly characterized by their relatively informal tone and the likely presence of personal expression.  

It is important to distinguish the datasets used in this study from those typically employed in tasks where predictions rely solely on textual content, such as sentiment analysis or emotion detection. In such tasks, the target attribute is often directly reflected in the words and phrases used within the text itself. In contrast, the personal traits we focus on, such as demographic information (e.g., age, gender, political affiliation) or personality, empathy, and moral values, were collected independently of the texts. This means that authors were not instructed to explicitly signal these attributes in their writing. Instead, they are inferred indirectly from the texts, since natural language use carries stylistic cues that can reveal aspects of identity. Below, we outline the specific personal traits and datasets that pair authors' texts with their corresponding personal traits for this study.  

\paragraph{Demographic attributes.} 
Demographic attributes are one type of personal trait we investigate. These include characteristics such as age, gender, and political affiliation, which are central to how individuals position themselves in society and are often reflected in distinct linguistic styles. In our experimental setting, the downstream task is to predict these demographic variables from authors' texts. Our data source is the United States Congressional Records \citep{gentzkow2018congressional}, which includes congressional floor speeches along with the speakers' demographic details \citep{fivethirtyeight_age_problem}. The dataset spans the 43$^{\text{rd}}$ to the 114$^{\text{th}}$ Congress, containing speeches from 8,520 speakers, with each speaker having delivered between 1 and 21,142 speeches of varying lengths. To ensure the quality and manageability of the data, we filter the dataset by selecting each speaker's longest utterances, aiming for a total of 4,000 words per speaker. This step removes short, uninformative utterances and accommodates computational limits on the number of tokens processed. We then sample the largest possible subset of speakers with a balanced representation of males/females, Republicans/Democrats, and four age groups: 27–40, 41–55, 56–70, and over 70, resulting in a final sample of 710 speakers. The final sample of 710 speakers is balanced by sex (males/females), political affiliation (Republicans/Democrats), and four age groups (27–40, 41–55, 56–70, and over 70 years), as detailed above.

\paragraph{Personality.} 
Personality traits capture enduring patterns of thought, feeling, and behavior, and they often manifest in people's writing styles. For example, more open individuals may use abstract or creative language \citep{abu2023reciprocal}, while extraverted individuals may employ more social and positive words \citep{chen2020meta}. To capture such traits, we rely on the Big Five personality framework \citep{goldberg2013alternative,goldberg1990standard}, which provides five well-established dimensions: openness (OPN), conscientiousness (CON), extraversion (EXT), agreeableness (AGR), and neuroticism (NEU). In our experimental setting, the downstream task is to predict whether an individual is high or low in each of the Big Five dimensions based on their writing. Our data source is the Essays dataset \citep{pennebaker1999linguistic}, which includes 2,348 essays, each written by a unique author, alongside the authors' scores on the 44-item Big Five Personality Inventory \citep{john1991big}. The essays were written using a stream-of-consciousness method, where authors were encouraged to freely express their thoughts. After completing the writing task, the authors completed the personality assessment. Following the approach of \citet{celli2013workshop}, we convert the authors' numerical self-assessments into nominal classes (low/high) using a median split based on $z$-scores. The Essays corpus aggregates stream-of-consciousness writing assignments from multiple introductory-psychology cohorts at the University of Texas at Austin and Southern Methodist University, collected by Pennebaker and King; participants are predominantly first-year undergraduates with a roughly balanced sex distribution typical of an introductory-psychology participant pool, as reported in the original publication.

\paragraph{Dispositional empathy.} 
Empathy refers to the ability to understand and share the feelings of others, and it represents another personal trait that can be inferred from text. For example, higher empathy may be reflected in perspective-taking or compassionate language. To capture this trait, we use the Interpersonal Reactivity Index (IRI; \citealp{davis1980multidimensional}), a 28-item self-report questionnaire that measures empathy across four dimensions: perspective-taking (PT), fantasy (FS), empathetic concern (EC), and personal distress (PD). In our experimental setting, the downstream task is to classify whether an author scores high or low on each empathy dimension based on their written texts. For our investigation, we utilize the Empathic Conversations dataset \citep{omitaomu2022empathic}, which was previously employed in the WASSA 2023 shared task \citep{barriere2023findings}. This dataset includes essays written in response to news articles along with the authors' IRI scores. Similar to the approach used for the Essays dataset, we convert the IRI scores into nominal classes indicating low/high levels for each dimension. After cleaning and preprocessing, the dataset comprises 711 essays written by 57 authors, each contributing between 1 and 72 reaction essays. We concatenate the essays from each author to create a more comprehensive representation of the authors' writing styles and better capture author-specific patterns while appropriately accounting for variation within each author's writings. The 57 contributing authors were adult U.S.-based Amazon Mechanical Turk workers; per-participant demographic details (sex/gender, age) are reported in the original dataset documentation.

\paragraph{Morality.} 
Moral values are the principles people use to distinguish right from wrong, and guide judgments about care, fairness, loyalty, authority, or purity. These values represent an important set of personal traits that can be predicted from language use. To operationalize morality, we rely on Moral Foundations Theory (MFT; \citealp{graham2013moral,haidt2004intuitive,atari2023morality}), which describes five core psychological systems: care/harm, fairness/cheating, loyalty/betrayal, authority/subversion, and purity/degradation. In our experimental setting, the downstream task is to classify whether an author scores high or low on each moral foundation dimension based on their writing. For our study, we utilize the YourMorals dataset \citep{kennedy2021moral}, which contains 107,798 Facebook posts from 2,691 users, along with their scores on the Moral Foundations Questionnaire (MFQ) \citep{graham2008moral}. Following the approach used in the Essays dataset, we convert the MFQ scores into nominal classes representing low/high levels for each moral foundation dimension. Consistent with our method in the Empathic Conversations dataset, we concatenate the posts from each author to create a comprehensive profile. The 2,691 contributing users self-reported a mean age of 32.8 years (SD=11.9, median 30), with 1,535 (57.0\%) identifying as male and 1,156 (43.0\%) identifying as female.

\subsection{Personal Traits Predictive Modeling from Text}
\label{method:personal-traits-predictive-modeling-from-text}

The primary goal of Study 2 is to quantify the reduction in predictive power for personal traits when classifiers, trained and tested on original texts, are applied to LLM-rewritten versions of those same texts.
To measure the predictive power of linguistic markers, we employ a bottom-up, data-driven approach \citep{kennedy2022handbook} that involves training classifiers on the authors' original texts to predict their personal traits. These classifiers are then applied to LLM-rewritten texts to compare their predictive performance.

The classifiers used in this study are:
\begin{itemize}
    \item Support Vector Machines (SVMs)
    \item Logistic Regression
    \item Random Forest
    \item Gradient Boosting
    \item Longformer \citep{beltagy2020longformer}, a Transformer model for long document processing.
\end{itemize}

These classifiers are used with two different featurization techniques:
\begin{itemize}
    \item Term Frequency-Inverse Document Frequency (TF-IDF)
    \item OpenAI text-embedding-ada-002 embeddings
\end{itemize}

This combination of traditional and transformer-based techniques allows us to cover a broad spectrum of feature extraction methods and classification models capturing different aspects of linguistic cues. Zero-shot classification by LLMs was considered, but given the complexity of this task \citep{abdurahman2023perils} as well as their inferior performance compared to trained classifiers presented in Appendix D.1 in Supplementary Material, we opted for training and fine-tuning classifiers in each experimental setup.

Each dataset is split into train, validation, and test sets using 5-fold cross-validation. We measure classifiers' performance using the $F_1$-macro score to account for imbalanced classes. The experiment is repeated 40 times with different random seeds to ensure robustness. Only models that achieve above-random performance on the original texts and have at least 20 successful runs are included in the analysis.

To compare the average predictive performance of classifiers on the LLM-rewritten texts and the original texts, we use paired $t$-tests. Paired $t$-tests rely on approximate normality of the paired differences. Following recent recommendations to avoid relying solely on null-hypothesis tests of normality, given that they reject trivially small deviations at large sample sizes \citep{shatz2024assumption}, we examined the skewness and excess kurtosis of the paired $F_1$ differences for each trait against commonly cited thresholds ($|\text{skewness}| \leq 2$, $|\text{excess kurtosis}| \leq 7$; \citealp{west1995structural,curran1996robustness}). Two traits (Empathy, Morality) fell within these thresholds; the remaining four exhibited heavier tails. Because the parametric $t$-statistic relies on the sampling distribution of the mean difference rather than the differences themselves, and because previous work has shown that Gaussian-model $p$-values are robust to non-normality at large sample sizes \citep{knief2021violating}, we additionally computed 95\% bootstrap confidence intervals on each mean difference using 10{,}000 resamples. Bootstrap and parametric confidence intervals agreed to within 1\% of their bounds for all six traits, confirming that the reported inferences are not driven by the normality assumption. Note that when comparing different experimental setups (e.g., two prompts), other variables are held constant (e.g., LLM and classifier) to isolate the effect of hyperparameters while accounting for the inherent characteristics of each model, prompt, or classifier. Given the large number of comparisons across classifiers, prompts, and LLMs, we apply Bonferroni corrections \citep{bonferroni1936teoria} to adjust significance thresholds. 

To more closely evaluate any patterns in the models' predictions and capture any imbalance between predictions, we define a new metric $\Delta_{r} = \frac{|P_{r, 1} - P_{r, 0}|}{P_{r, 1} + P_{r, 0}}$, where $P_{r, i}$ is the frequency of predictions of class $i$ ($i \in \{0, 1\}$) in run $r$. To quantify the prediction imbalance in age groups which have more than two classes, we focus on the imbalance between the most and the least frequently predicted classes. Since the random baseline across the classification tasks is close to $50\%$, indicating a balance between the two classes, a large $\Delta$ signals that the model's predictions are skewed toward one class, either 0 or 1. Thus, the further $\Delta$ is from zero, the more biased the predictions are, suggesting that the text being analyzed contains linguistic features that disproportionately influence the model toward associating the texts with one particular class. We perform a Wilcoxon signed-rank test, a non-parametric counterpart of paired $t$-test, due to non-normality in the data, to test the significance of the differences in $\Delta$ values between original and LLM-rewritten texts.

Finally, we examine changes in classifier predictions between the original and the LLM-rewritten texts to identify which personal profiles become more prominently predicted or reinforced in LLM-rewritten texts. In particular, we consider the direction of changes in classifier predictions as a proxy for what the LLMs promote in their generated output, essentially revealing which groups the text tends to reflect more. We perform paired $t$-tests between the number of prediction changes from 0 to 1 and from 1 to 0 after LLM-rewrites to quantify the significance of the shift in predicted personal traits.

\subsection{Associations between Linguistic Features and Personal Traits}
\label{method:association-between-linguistic-features-personal-traits}

\begin{figure}[!htbp]
    \centering
    \includegraphics[width=\textwidth]{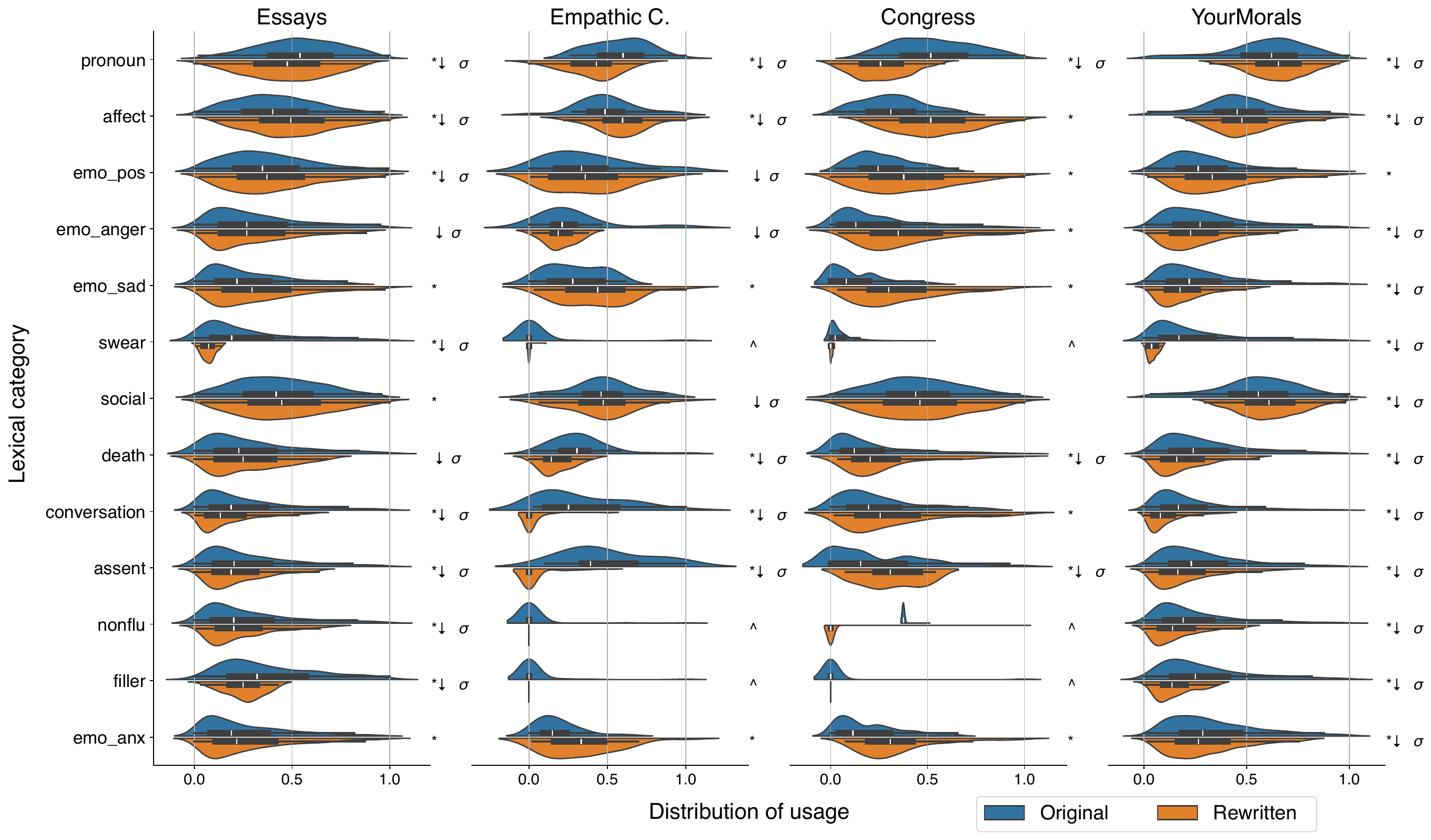}
   \extfigcaption{Lexical homogenization across domains. Distributions of per-text usage (relative frequencies, min–max scaled to [0, 1]) for representative lexical categories in original (blue) vs. LLM-rewritten (orange) texts, using GPT-3.5 with the Rephrase prompt, across four datasets. The unit of analysis is one text; original and LLM-rewritten versions are paired by text (no separate control group is appropriate because each original text serves as its own paired baseline). Sample sizes per dataset (all categories share the same n within a dataset): Essays n = 2,348, Empathic Conversations n = 711, Congress n = 710, YourMorals n = 3,641. Box-plot elements inside each violin: center line, median; box bounds, 25th–75th percentiles (interquartile range, IQR); whiskers, $1.5\times$IQR. Categories with a statistically significant shift in central tendency between original and LLM-rewritten distributions are marked with * (two-sided independent-samples Student's t-test, $p < .05$); categories whose variance is smaller after rewriting are marked with $\downarrow\sigma$; categories for which $\leq10\%$ of texts in either condition contain the feature are marked with $\wedge$. Exact t-statistics, exact p-values, and Levene's-test results for every (dataset $\times$ category) cell are reported in Supplementary Table C7 (SM Section C, Lexical Shifts).}
    \label{extended-fig:lexical-variability}
\end{figure}

To study the associations between linguistic features and personal traits, we adopt a top-down, theory-driven approach \citep{kennedy2022handbook}. Specifically, we extract linguistic features from texts using validated lexicons commonly employed in psychology and social science research. By comparing the associations between these linguistic cues and personal traits in original and LLM-rewritten texts, we aim to identify how LLMs influence these relationships to gain insight into how LLMs alter writing styles, reducing but not eliminating the predictive power of linguistic features in text. We further analyze the distributional shifts and the homogenization of these important linguistic features (see \Cref{extended-fig:lexical-variability} and Appendix C in Supplementary Material).

To extract associations between linguistic features and personal traits, we focus on various well-established dictionaries. For each dictionary category, we calculated the ratio of words belonging to that category relative to the total word count, then standardized these frequencies as scores. We computed Pearson correlations ($r$) between these standardized word frequencies and the $z$-scores of continuous traits (e.g., age, personality, empathy, morality) and performed $t$-tests to compare categorical traits (e.g., gender, political affiliation). Pearson correlations assume bivariate normality and linearity. The test statistic has been shown to be robust to departures from normality \citep{havlicek1976robustness}, although Pearson's $r$ can remain sensitive to heavier-tailed distributions and other features of the joint distribution even at large sample sizes \citep{wilcox2001modern}, motivating the cross-checks described below. We examined the skewness and excess kurtosis of each variable entered into the correlations against thresholds of $|\text{skewness}| \leq 2$ and $|\text{excess kurtosis}| \leq 7$ \citep{west1995structural,curran1996robustness}, following recent recommendations to favour effect-size moment statistics over significance tests of normality at large sample sizes \citep{shatz2024assumption}. All trait variables and the majority of lexical features fell within these bounds. A small subset of zero-inflated, low-frequency lexical features (e.g., NRC negative-word and swear-word frequencies) showed heavier tails; for these we additionally computed Pearson correlations on log-transformed values and Spearman rank correlations, and confirmed that direction and statistical significance were concordant with the original Pearson results. For the independent-samples $t$-tests comparing categorical-trait groups (e.g., male vs.\ female; Democrat vs.\ Republican), we additionally examined per-group skewness and excess kurtosis and the equality of group variances via Levene's test. While most features met both normality and homogeneity-of-variance criteria, a subset showed deviations from either or both. To verify that this did not affect inference, we re-ran each comparison using Welch's $t$-test, which relaxes the equal-variances assumption and has been recommended as the default for two-sample comparisons \citep{delacre2017psychologists}, and additionally using the Mann-Whitney $U$ test as a non-parametric alternative. Across all features tested, Student's, Welch's, and Mann-Whitney tests agreed in direction; Welch's and Student's reached the same Bonferroni-adjusted significance decision, and Mann-Whitney $U$ agreed with Student's in all cases, confirming that the reported group differences are not driven by violations of the parametric assumptions. To control for multiple comparisons, we applied Bonferroni corrections. 

We use the following general-purpose and content-specific dictionaries:

\begin{itemize}
    \item \textbf{Linguistic Inquiry and Word Count (LIWC)}: We use the merged version of LIWC 22 and LIWC 07 \citep{pennebaker2022linguistic}, a widely used lexicon that captures psychological, social, cognitive, and affective processes.
    \item \textbf{NRC Emotion Lexicon}: A 10-category lexicon \citep{Mohammad13, mohammad-turney-2010-emotions} that captures emotional expression, used to explore associations with personality traits.
    \item \textbf{Empathy Lexicon}: A lexicon categorized into high-empathy, low-empathy, and distress-related words (four categories), used to examine links between language use and dispositional empathy \citep{sedoc2019learning}.
    \item \textbf{Moral Foundations Dictionary 2 (MFD2)}: A lexicon capturing moral language across five foundations, with separate categories for virtue and vice \citep{frimer2019moral}.
\end{itemize}

\section*{Data availability}
Of the seven datasets analyzed in this article, six are publicly available and one is access-restricted.

\textbf{Publicly available datasets.} Reddit r/WritingPrompts posts were obtained from Project Arctic Shift (\url{https://github.com/ArthurHeitmann/arctic_shift}), which mirrors publicly accessible Reddit content (excluding private subreddits and content for which the original poster has filed a removal request) and is itself a successor archive to the Pushshift Reddit dataset \citep{baumgartner2020pushshift}. arXiv abstracts were obtained from the publicly released metadata collection \citep{arxiv_org_submitters_2024}, distributed under a Creative Commons CC0 1.0 Universal Public Domain Dedication. Patch News articles were accessed via the public Patch.com archive. United States Congressional Records floor speeches were obtained from the publicly released parsed corpus \citep{gentzkow2018congressional}, with the underlying speeches in the public domain as works of the U.S.\ federal government. YourMorals Facebook posts and Moral Foundations Questionnaire responses are publicly released by the original investigators \citep{kennedy2021moral}; the data were collected from participants who voluntarily completed self-report measures on yourmorals.org and consented at registration to having their Facebook posts accessed for research purposes, under University of Southern California Institutional Review Board protocol UP-07-00393-AM019 (M.\ Dehghani, PI). Empathic Conversations are publicly released by the original investigators \citep{omitaomu2022empathic}; the dataset was collected under University of Pennsylvania IRB protocol \#826448 from Amazon Mechanical Turk workers who were informed that their conversations, demographic information, personality surveys, and essays would be used for academic research and distributed externally for research purposes, and were compensated per HIT.

\textbf{Access-restricted dataset.} The Essays corpus \citep{pennebaker1999linguistic} is not publicly redistributable and is governed by access conditions imposed by the dataset owner. The corpus can be requested directly from Prof.\ James W.\ Pennebaker (University of Texas at Austin; \href{mailto:pennebaker@utexas.edu}{pennebaker@utexas.edu} or \href{mailto:jwpennebaker@gmail.com}{jwpennebaker@gmail.com}); the original data were collected at UT-Austin from undergraduate psychology students who participated for course credit, with informed consent and institutional research approval reported in the original paper. Access is granted at the discretion of the dataset owner and does not require additional ethics approval from the requester beyond standard institutional review. We obtained the corpus via direct request and were not permitted to redistribute it.

\textbf{Use of Reddit and Patch News content.} In line with the secondary-analysis posture taken by recent \emph{Nature Human Behaviour} work on public text corpora \citep{liang2025quantifying}, all use of Reddit and Patch.com content in this article is restricted to aggregate quantitative analysis of linguistic patterns over time. No individual Reddit post or Patch.com article is reproduced, republished, or redistributed in this paper or its accompanying data release. Usernames, post identifiers, and article identifiers are not retained in the analysis pipeline. Use is academic and non-commercial. Future deletion requests submitted through Arctic Shift's public removal mechanism will be honored in any continuing analyses.

\section*{Code availability}  

All custom code used in this article (implemented in Python 3.11.7 and R 4.4.2) is publicly available under the MIT license at
\url{https://github.com/zhpinkman/LLMs-effect-on-linguistic-diversity}
and permanently archived at \url{https://doi.org/10.5281/zenodo.21633457}. The repository includes scripts for data preprocessing, LLM rewriting, classifier training and evaluation, statistical analyses, and figure generation. No proprietary software is required to reproduce the analyses.

\section*{Funding Statement}
This research was supported, in part, by the Army Research Laboratory under contract W911NF-23-2-0183, by DARPA INCAS HR001121C0165, and by Air Force Office of Scientific Research A9550-23-1-0463. The funders had no role in study design, data collection and analysis, decision to publish, or preparation of the manuscript. The views and conclusions contained herein are those of the authors and should not be interpreted as necessarily representing the official policies, either expressed or implied, of DARPA, AFOSR, or the U.S. Government. The U.S. Government is authorized to reproduce and distribute reprints for governmental purposes notwithstanding any copyright annotation therein.

\section*{Acknowledgements}
We thank James W. Pennebaker for their support and invaluable insights throughout the research process.

\section*{Author Contributions Statement}

Z.S. conceived the study, designed the experimental setup, collected datasets, performed experiments and analyses, and wrote the manuscript.  
F.K.-M. contributed to the design and execution of the time-series analyses, interpretation of findings, and writing of the corresponding sections.  
M.O. contributed to the design and implementation of Study 3, interpretation of results, and writing, as well as supporting overall manuscript development.  
C.M. contributed to the implementation and analysis of Study 3 and to writing the corresponding sections.  
A.Z. contributed to data preparation, experimental support, and conceptualization of the study.  
J.T. contributed to conceptualization of the study.  
A.T. contributed to conceptualization and writing.  
M.C. contributed to the design and implementation of the time-series analyses.  
F.M. contributed to the overall conceptualization of the study.  
M.D. supervised the project and contributed to conceptualization, data collection, and writing.  

\section*{Competing Interests Statement}

The authors declare no competing interests.

\clearpage






\clearpage


\section*{Supplementary material}

# Appendix A  Homogenization Trends

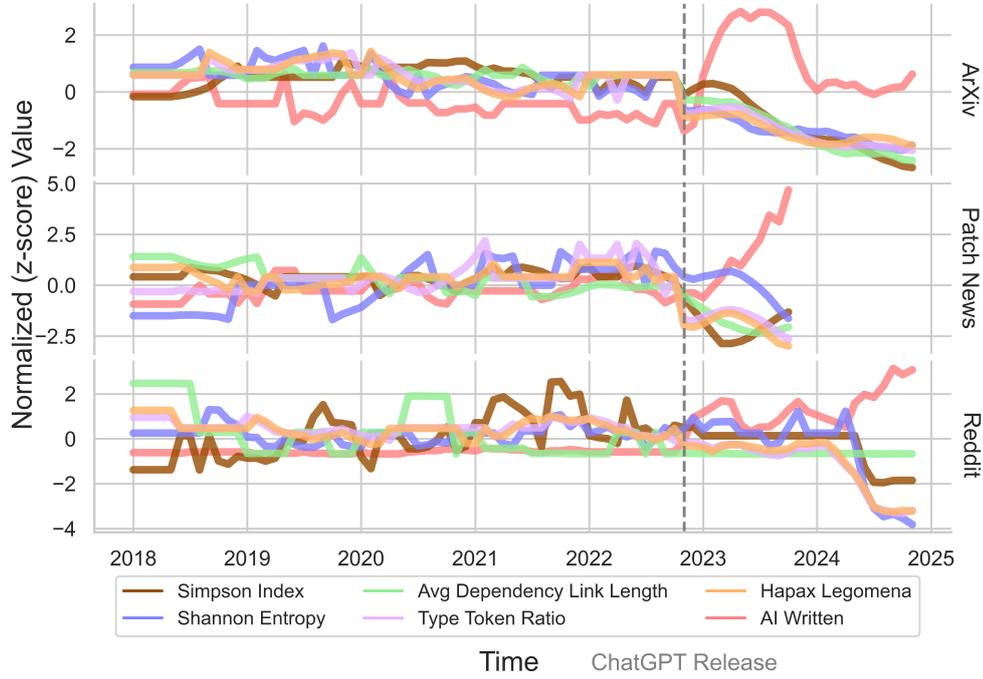

**Fig. A1**: Trends in variance of complexity-related features in texts and the attribution rate of texts as AI-generated.

Study 1 provided strong evidence for the homogenization effect of LLMs, showing a marked reduction in linguistic complexity variance across documents after the introduction of ChatGPT. This trend was shown by aggregating multiple features related to linguistic complexity. Figure A1 illustrates this trend across all complexity-related features analyzed in this paper, revealing a consistent decline in variability following the introduction of ChatGPT on November 30, 2022. Notably, this homogenization is not due to a simplification of language but rather an overall enhancement of complexity—Figure A2 shows that complexity-related features increased on average. This suggests that LLMs are promoting a more polished and elaborately structured style of writing, amplifying certain sophisticated linguistic patterns while narrowing the range of individual expression.

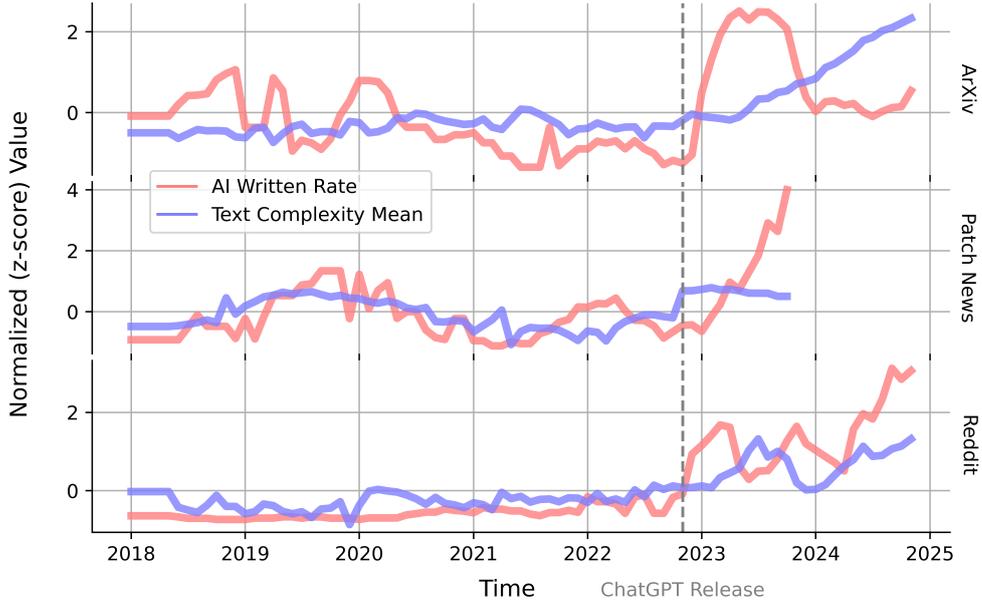

**Fig. A2**: Trends in average writing complexity in texts and the attribution rate of texts as AI-generated.

## Appendix B  LIWC and Construct-specific Lexicons

In this section, we present the details of the word categories that were used from LIWC or other construct-specific lexicons in Study 4.

### B.1  LIWC

The Linguistic Inquiry and Word Count (LIWC; 94) dictionary is a collection of words categorized into different psychological and linguistic dimensions. It is the backbone of the LIWC software that analyzes text by counting occurrences of words in specific categories, offering insights into the involved emotions, cognition, social interactions, and other linguistic styles. The LIWC categories that we used in this study are as follows:

- Affect: The Affect category is a measure of the emotional tone as well as emotion content of text such as 'happy', 'hate', 'love', and 'wrong'.
- emotion: Emotion words represent a broad category of emotional expressions referring to specific emotional states, such as 'good', 'love', 'happy', 'hope'.
- assent: Assent words are expressions of agreement, confirmation, or approval, typically used in conversational contexts. These include terms like 'yeah', 'yes', 'okay', and 'ok', which signal a positive or affirmative response in communication.
- emo_neg: Negative emotion words express negative affective states, including terms like 'bad', 'hate', 'hurt', 'tired'.

- emo_pos: Positive emotion words convey positive emotions, including terms like 'happy', 'excited', and 'grateful'.
- emo_anger: Anger words express anger or frustration, including terms like 'hate', 'mad', and 'angry'.
- emo_sad: Emotionally sad words convey feelings of sadness or melancholy, including terms like ':(', 'sad', and 'disappoint'.
- swear: Swear words consist of profane or vulgar language used to express strong emotions or taboo subjects, such as 'damn', 'f***', and 'sh*t'.
- affiliation: Affiliation words indicate a sense of belonging or connection to others, including terms like 'friend', 'community', and 'support'.
- socrefs: Social referents encompass words that relate to social interactions, like 'we', 'you', or 'he', as well as words related to family and friends (e.g., 'parent', 'mother', 'girlfriend').
- family: Family words relate to familial relationships or members, such as 'mother', 'father', 'sibling', and 'cousin'.
- friend: Friend words relate to friendship or friendly interactions, including terms like 'friend', 'buddy', 'dude', and 'girlfriend'.
- pronoun: Pronouns are words used to replace nouns indicating individuals or groups, such as 'I', 'you', 'he', 'she', 'we', and 'they'.
- we: This category refers to words indicating group membership or inclusion, such as 'we', 'us', and 'our'.
- you: Second-person pronouns, such as 'you' and 'your'.
- shehe: Third-person pronouns refer to individuals such as 'he', 'she', 'him', and 'her.'
- i: First-person singular pronouns refer to the speaker or author, such as 'I', 'me', 'myself', and 'mine'.
- differ: Differentiation words express contrast or distinction between entities or ideas, including terms like 'but', 'not', 'or', and 'if'.
- tentat: Tentative words express uncertainty or hesitation, including terms like 'maybe', 'if', and 'something'.
- cogproc: Cognitive processing words indicate intellectual or cognitive engagement, including terms like 'think', 'understand', 'analyze', and 'consider'.
- prosocial: Prosocial words denote behaviors or attitudes that benefit others or society, such as 'help', 'care', 'thank', and 'please'.
- BigWords: Big words are words with complex structures (7 letters or longer) that may indicate intellectual complexity or formality, such as 'procrastination', 'circumstantial', and 'phenomenon'.
- Drives: Drive words refer to motivational or goal-oriented language, encompassing affiliation (e.g., 'we', 'our', 'us', 'help'), achievement (e.g., 'work', 'better', 'best'), and power (e.g., 'own', 'order', 'allow').
- Social: The Social category stands for social processes, and encompasses words pertain to various types of social behavior ('love', 'care', 'please', 'good morning', 'attack', 'deserve', 'judge') and social referents as defined above.
- achieve: Achievement words denote actions or concepts related to accomplishment or success, such as 'work', 'bonus', 'beat', and 'overcome'.

- inhib: The inhibition category refers to words related to restraint, suppression, and inhibition, such as 'block' and 'constrain'.
- religion: Religion words pertain to religious concepts, practices, or institutions, such as 'God', 'hell', and 'church'.

## B.2 NRC Emotion Lexicon

We use the dictionary from Mohammad and Turney [85, 86] with the following word categories:

- nrc.positive: Words that have a positive sentiment, such as 'acceptable', 'boon', and 'civil'.
- nrc.negative: Words that have a negative sentiment, such as 'aberrant', 'abort', and 'begging'.
- nrc.anger: Words that relate to the emotion of anger, such as 'arguments', 'confront', and 'friction'.
- nrc.disgust: Words that relate to the emotion of disgust, such as 'barf', 'decompose', and 'gut'.
- nrc.sadness: Words that relate to the emotion of sadness, such as 'blue', 'cloudy', and 'emptiness'.
- nrc.anticipation: Words that relate to the emotion of anticipation, such as 'accelerate', 'announcement', and 'approaching'.
- nrc.trust: Words that relate to the emotion of trust, such as 'adhering', 'advice', and 'collaborator'.
- nrc.joy: Words that relate to the emotion of joy, such as 'whimsical', 'beach', and 'doll'.

## B.3 Distress and Empathy Lexicon

We use the dictionary from Sedoc et al. [87] with the following word categories:

- empathy.low: Low-empathy words from the empathy lexicon (based on a median split from lexicon weights), such as 'joke', 'bizarre', and 'stupidest'.
- empathy.high: High-empathy words from the empathy lexicon (based on a median split from lexicon weights), such as 'healing', 'grieve', and 'heartbreaking'. This category did not significantly correlate with any dimension of dispositional empathy.
- distress.low: Low-distress words from the distress lexicon (based on a median split from lexicon weights), such as 'dunno', 'guessing', and 'anyhow'.
- distress.high: High-distress words from the distress lexicon (based on a median split from lexicon weights), such as 'inhumane', 'dehumanizes', and 'mistreating'. This category did not significantly correlate with any dimension of dispositional empathy.

## B.4 Moral Foundations Dictionary 2 (MFD2)

We use the dictionary from Frimer et al. [88] with the following word categories:

- mfd.authority.virtue: Words related to the "virtue" dimension of the moral foundation of Authority, such as 'respect', 'obey', and 'honor'.

- mfd.authority.vice: Words related to the "vice" dimension of the moral foundation of Authority, such as 'disrespect', 'disobey', and 'chaos'.
- mfd.care.virtue: Words related to the "virtue" dimension of the moral foundation of Care, such as 'compassion', 'generosity', and 'pity'.
- mfd.care.vice: Words related to the "vice" dimension of the moral foundation of Care, such as 'harm', 'threatens', and 'injured'. This category did not significantly correlate with the Care foundation.
- mfd.purity.virtue: Words related to the "virtue" dimension of the moral foundation of Purity, such as 'sacred', 'wholesome', and 'divine'.
- mfd.purity.vice: Words related to the "vice" dimension of the moral foundation of Purity, such as 'sin', 'defiled', and 'contaminate'.
- mfd.loyalty.virtue: Words related to the "virtue" dimension of the moral foundation of Loyalty, such as 'loyalty', 'allegiance', and 'follower'. This category did not significantly correlate with the Loyalty foundation.
- mfd.loyalty.vice: Words related to the "vice" dimension of the moral foundation of Loyalty, such as 'disloyal', 'treason', and 'enemy'. This category did not significantly correlate with the Loyalty foundation.
- mfd.fairness.virtue: Words related to the "virtue" dimension of the moral foundation of Fairness, such as 'fairness', 'justice', and 'equality'. This category did not significantly correlate with the Fairness foundation.
- mfd.fairness.vice: Words related to the "vice" dimension of the moral foundation of Fairness, such as 'cheat', 'unjust', and 'unequal'. This category did not significantly correlate with the Fairness foundation.

## Appendix C  Semantic Similarities

Figures provided in Study 2 depict the semantic similarity between the original texts and the LLM-rewritten texts when the rephrase prompt (R) is used with GPT3.5. We saw similar trends using other prompts as well as other LLMs. The semantic similarity of the original and LLM-rewritten texts, categorized by the utilized prompt and LLM are presented in Figure C3 and Figure C4, respectively.

A Kruskal-Wallis (H) test, a non-parametric alternative to ANOVA, was performed to compare similarity scores across LLMs and datasets, while a Mann–Whitney (U) test was used to compare the similarities across two prompts. Our results indicated that computed similarities were significantly different across LLMs ($H = 3138, p < .001, \eta^2 = 0.08$ [moderate effect size]), two prompt conditions ($U = 247042398, p < .001, r = 0.32$ [medium effect size]), as well as datasets ($H = 2289, p < .001, \eta^2 = 0.06$ [moderate effect size]).

Furthermore, the quantitative analysis of the pairwise comparison between different prompts, LLMs, and different datasets, based on the semantic similarity between original and LLM-generated texts, are demonstrated in Table C1, Table C2, and Table C3, respectively.

Pairwise post-hoc Dunn's tests [95][5] with Benjamini-Hochberg corrections for multiple comparisons showed significant differences between all LLM pairs ($p < .001$; see

---

[5]Dunn's tests are conventionally used for pairwise comparisons after a Kruskal-Wallis test is rejected.

Table C2), with GPT3.5 being the most preservative LLM and Llama 3 being the least preservative LLM in terms of the level of semantic preservation when rewriting.

With a similar post-hoc analysis, comparing the two prompt conditions, the Syntax_Grammar prompt was more preservative than the Rephrase prompt with median similarity values of 0.97 and 0.96, respectively (see Table C1). Finally, underscoring the role of context when studying LLMs' impact on authors' texts, our pairwise post-hoc tests showed significant differences between all dataset pairs aside from YourMorals vs. Empathetic Conversations and Patch News vs. Empathetic Conversations (see Table C3), which showed the highest level of semantic preservation for the ArXiv dataset and the least level of preservation for the Patch news dataset.

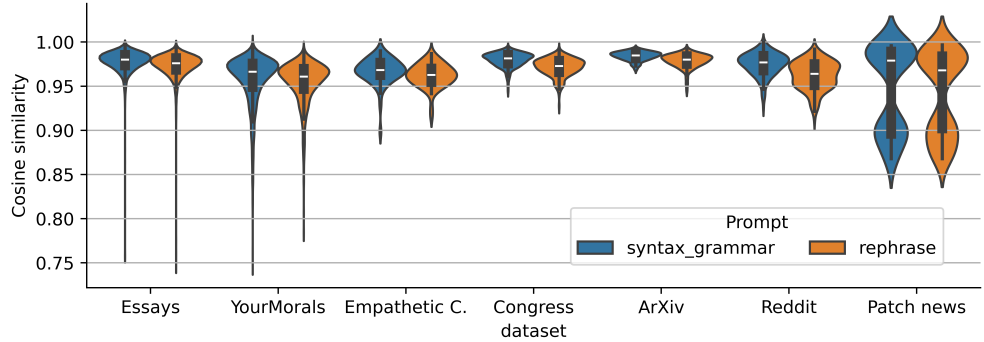

**Fig. C3**: Semantic similarity between original and LLM-generated texts across different data sources and prompts.

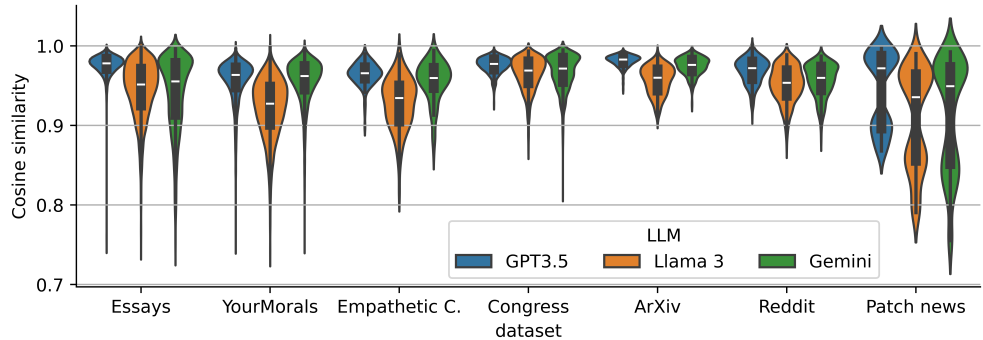

**Fig. C4**: Semantic similarity between original and LLM-generated texts across different data sources and LLMs.

**Table C1**: Pairwise comparisons between different prompts (**R**ephrase and **S**yntax_**G**rammar), based on the semantic similarity computed between the original and LLM-generated texts in each dataset, with Mann-Whitney (U) test.

| Dataset | prompt 1 | prompt 2 | U | Adjusted p-value |
|---|---|---|---|---|
| Essays | R | SG | 4290249 | <.001 |
| YourMorals | R | SG | 8673761 | <.001 |
| Congress | R | SG | 450712 | <.001 |
| Empathetic C. | R | SG | 2371 | <.001 |
| ArXiv | R | SG | 17495 | .984 |
| Reddit | R | SG | 36775 | <.001 |
| Patch news | R | SG | 22669 | .010 |

## Appendix D  Lexical Shifts

Study 4 revealed that many well-established associations between lexical categories and personal traits diminished in LLM-rewritten texts. Moreover, we observed significant shifts in both the frequency and diversity of lexical categories often linked to personal traits, such as pronouns [10]. Figure D5 illustrates these differences for GPT-3.5 outputs with the Rephrase prompt.

Notably, LLM-rewritten texts exhibited a sharp decline in the use of pronouns, swear words, death-related terms, and conversational markers (e.g., "yeah," "oh," "um"). At the same time, they showed an increase in words reflecting positive emotions, sadness, and socially oriented language. This pattern suggests that LLMs tend to reduce informal, conversational, and negatively charged language while amplifying structured, emotionally positive, and socially engaging expressions—potentially reflecting biases introduced during training.

Beyond frequency shifts, we observed a systematic decline in lexical diversity across multiple word categories, further supporting the homogenization effects identified in Studies 1 and 2. Categories such as pronouns, non-fluent words, and conversational speech markers exhibited reduced variability. As shown in Table D4, which presents variance in usage distributions normalized via min-max scaling across original and LLM-generated documents, for each lexical category and data source, the majority of cases displayed a decrease in diversity (indicated by the ↓ symbol). This reduced lexical variety may diminish the predictive power of linguistic markers for personal traits, as discussed in Studies 3 and 4.

Similar trends emerged across different models and prompting strategies. Figure D6 shows the lexical shifts induced by the syntax_grammar prompt on GPT-3.5, while Figure D7 and Figure D8 demonstrate similar effects for Gemini. Likewise, Figure D9 and Figure D10 show comparable shifts in Llama 3.

These trends may stem from model alignment during reinforcement learning from human feedback (RLHF; 96). While RLHF is designed to encourage helpful and less harmful language [96], our findings suggest that its effects extend beyond content moderation. The observed reduction in pronouns and conversational markers—features

**Table C2**: Pairwise comparisons between different LLMs, based on the semantic similarity computed between the original and LLM-generated texts in each dataset, with Dunn's test and Benjamini-Hochberg adjustments for multiple comparisons.

| Dataset | LLM 1 | LLM 2 | $z$ | Adjusted $p$-value |
| --- | --- | --- | --- | --- |
| Essays | Gemini | GPT3.5 | 40.610 | <.001 |
|  | Gemini | Llama 3 | -5.614 | <.001 |
|  | GPT3.5 | Llama 3 | -47.728 | <.001 |
| YourMorals | Gemini | GPT3.5 | 0.123 | .902 |
|  | Gemini | Llama 3 | -41.870 | <.001 |
|  | GPT3.5 | Llama 3 | -41.955 | <.001 |
| Congress | Gemini | GPT3.5 | 8.915 | <.001 |
|  | Gemini | Llama 3 | -4.506 | <.001 |
|  | GPT3.5 | Llama 3 | -13.421 | <.001 |
| Empathetic C. | Gemini | GPT3.5 | 2.637 | .008 |
|  | Gemini | Llama 3 | -6.883 | <.001 |
|  | GPT3.5 | Llama 3 | -9.597 | <.001 |
| ArXiv | Gemini | GPT3.5 | 8.462 | <.001 |
|  | Gemini | Llama 3 | -13.889 | <.001 |
|  | GPT3.5 | Llama 3 | -22.346 | <.001 |
| Reddit | Gemini | GPT3.5 | 8.095 | <.001 |
|  | Gemini | Llama 3 | -3.086 | .002 |
|  | GPT3.5 | Llama 3 | -11.177 | <.001 |
| Patch news | Gemini | GPT3.5 | 8.427 | <.001 |
|  | Gemini | Llama 3 | -2.632 | .008 |
|  | GPT3.5 | Llama 3 | -11.059 | <.001 |

that are neither harmful nor undesirable—raises the possibility that RLHF inadvertently suppresses stylistic diversity, potentially diminishing the linguistic signals that contribute to individual expression and identity.

# Appendix E  Analysis of Predictive Power of Personal Traits from Text

## E.1  Using LLMs as Zero-shot Classifiers

The mean performance of trained classifiers discussed in Study 3 in comparison with a Llama3.1 [58] as a zero-shot classifier [97] for each personal trait prediction task is demonstrated in Table E5. Across all prediction tasks and contexts, trained classifiers outperform Llama3.1, which was chosen as a representative powerful open-source large language model for task automation in scale. Since Study 3 required classifiers that were already proficient in predicting personal traits based on authors' original texts, we selected trained classifiers for this task. These classifiers outperformed large language models (LLMs) like Llama3.1, which reinforced our decision to rely on them.

**Table C3**: Pairwise comparisons between different datasets, based on the semantic similarity computed between the original and LLM-generated texts in each dataset, with Dunn's test and Benjamini-Hochberg adjustments for multiple comparisons.

| Dataset1 | Dataset2 | $z$ | Adjusted p-value |
| --- | --- | --- | --- |
| Essays | YourMorals | -25.314 | <.001 |
| Essays | Congress | 21.479 | <.001 |
| Essays | Empathetic C. | -7.138 | <.001 |
| Essays | Patch news | -16.706 | <.001 |
| Essays | ArXiv | 14.667 | <.001 |
| Essays | Reddit | -3.802 | <.001 |
| YourMorals | Congress | 39.110 | <.001 |
| YourMorals | Empathetic C. | -1.842 | .065 |
| YourMorals | Patch news | -6.983 | <.001 |
| YourMorals | ArXiv | 24.629 | <.001 |
| YourMorals | Reddit | 6.017 | <.001 |
| Congress | Empathetic C. | -13.589 | <.001 |
| Congress | Patch news | -26.837 | <.001 |
| Congress | ArXiv | 2.093 | .040 |
| Congress | Reddit | -14.936 | <.001 |
| Empathetic C. | Patch news | -1.744 | .081 |
| Empathetic C. | ArXiv | 13.502 | <.001 |
| Empathetic C. | Reddit | 4.525 | <.001 |
| Patch news | ArXiv | 23.156 | <.001 |
| Patch news | Reddit | 9.521 | <.001 |
| Arxiv | Reddit | -13.631 | <.001 |

Additionally, the datasets used in this study have been previously published, making them potential sources of training data for LLMs. Given the opacity surrounding what is included in LLM training data, it is difficult to confirm whether these models had prior exposure to the test splits reserved for performance evaluation.

### E.2 Systematicity of Change in Predictive Powers

In Study 3, we observed that changes in model predictions on LLM-rewritten texts were not random but systematically biased toward a specific class and this was the case across all data sources. This suggests that LLMs influence linguistic cues in a way that consistently alters how personal traits are inferred. Here, we provide an additional sensitivity check using Gemini and Llama, where we find the same systematic trends: $\Delta$ for LLM-rewritten texts remains consistently larger than for original texts across nearly all categories (see Figure E11 and Figure E12). This further supports our finding that LLMs homogenize text in a way that skews predictions toward particular traits, reinforcing the systematic nature of their effect on linguistic markers.

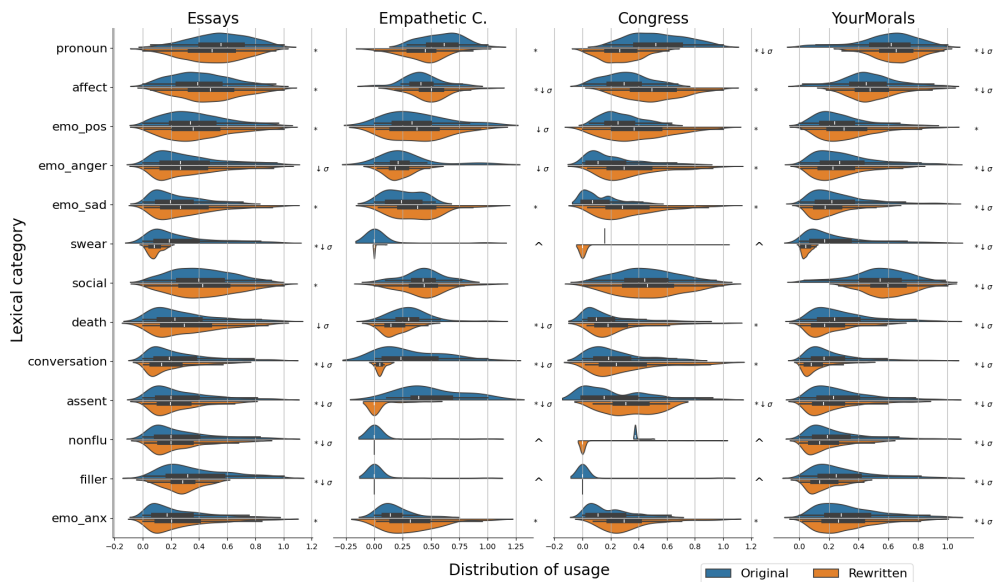

**Fig. D5**: Distribution of usage of different lexical word categories in the original and LLM-rewritten texts (with rephrase prompt on GPT3.5) across different data sources. Items with a significant difference (adjusted $p$-value $< 0.05$) between original and LLM-rewritten distributions according to a Mann-Whitney U test are indicated with a **\***, while the items in which no more than 10% of the LLM-rewritten texts remain to contain the specified lexical feature are indicated using a **^**. Furthermore, items where variance in features is significantly less in LLM-generated text compared to original documents using a Levene's test are denoted by $\downarrow \sigma$.

## E.3 Additional Insights about Loss in Predictive Power of Personal Traits from Text

The impact of LLMs on the predictive power of linguistic patterns over authors' personal traits across various psychological and demographic dimensions is shown in Table E6. In all datasets and across different dimensions, we observe a consistent reduction in predictive power when LLMs are involved as writing assistants. However, one notable exception is the Care dimension in moral values, where LLM involvement actually increases the predictive power.

This unique result may be related to how LLMs are trained with reinforcement learning from human feedback (RLHF), where they are optimized to be more helpful, cooperative, and less harmful. Given that the Care dimension is closely associated with helpfulness, being less harmful, and prosocial behaviors, the alignment with RLHF objectives might explain why LLM-generated texts preserve or even enhance linguistic patterns tied to this trait. LLMs may inherently promote language that reflects values aligned with Care as part of their effort to avoid harmful language and foster positive, helpful communication.

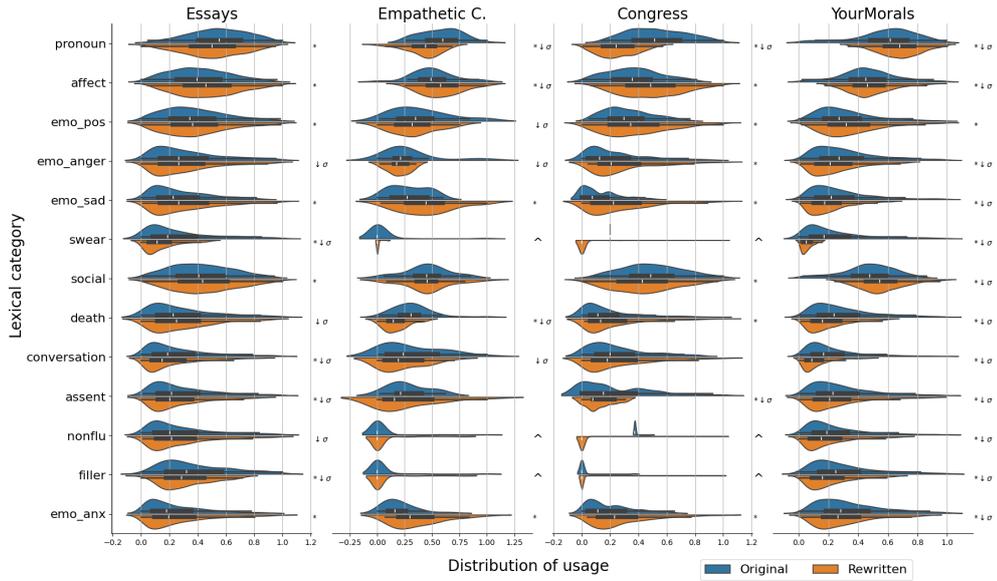

**Fig. D6**: Distribution of usage of different lexical word categories in the original and LLM-rewritten texts (with syntax_grammar prompt on GPT3.5) across different data sources. Items with a significant difference (adjusted $p$-value $< 0.05$) between original and LLM-rewritten distributions according to a Mann-Whitney U test are indicated with a *, while the items in which no more than 10% of the LLM-rewritten texts remain to contain the specified lexical feature are indicated using a ˆ. Furthermore, items where variance in features is significantly less in LLM-generated text compared to original documents using a Levene's test are denoted by $\downarrow \sigma$.

In the main body of the paper, we demonstrated how LLMs decrease the predictive power of text over authors' personal traits, aggregating over all different LLMs, prompts, classifiers, and featurization techniques. We saw similar trends using different LLMs, prompts, classifiers, and featurization techniques that are shown in Figure E13, Figure E14, Figure E15, and Figure E16, respectively.

Although using different LLMs, prompts, classifiers, or featurization techniques did not change the obtained results regarding the reduction of predictive power over personal traits when LLMs are involved as writing assistants, we observed significant differences between different choices of LLMs, prompts, classifiers, and featurization techniques.

Considering $\Delta = \frac{F_1(original) - F_1(LLM-rewritten)}{F_1(original)}$ as the statistic of interest, where $F_1(original)$ is the $F_1$ score obtained by the trained classifier on the unseen test set, and $F_1(LLM-rewritten)$ is the $F_1$ score obtained by the same classifier on the LLM-rewritten version of the texts, we performed Kruskal-Wallis (for comparisons with more than two categories) and Mann–Whitney (for comparisons with two categories) tests to compare the drops in the predictive powers across personal traits, LLMs, prompts, classifiers, and featurization techniques.

44

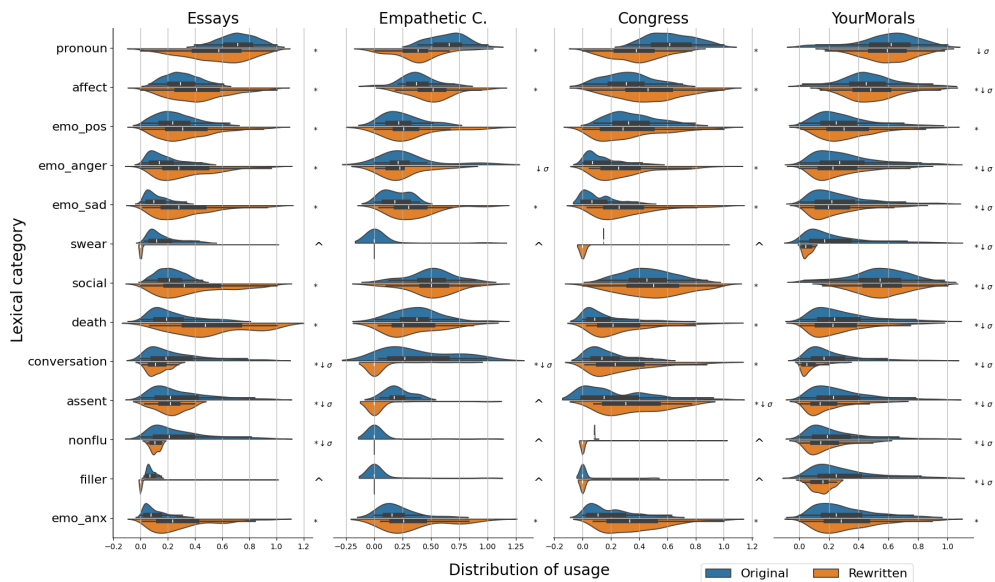

**Fig. D7**: Distribution of usage of different lexical word categories in the original and LLM-rewritten texts (with Rephrase prompt on Gemini) across different data sources. Items with a significant difference (adjusted $p$-value $< 0.05$) between original and LLM-rewritten distributions according to a Mann-Whitney U test are indicated with a **∗**, while the items in which no more than 10% of the LLM-rewritten texts remain to contain the specified lexical feature are indicated using a **^**. Furthermore, items where variance in features is significantly less in LLM-generated text compared to original documents using a Levene's test are denoted by $\downarrow \sigma$.

Our results indicated that computed similarities were significantly different across personal traits ($H = 41.91, p < .001, \eta^2 = 0.76$ [Large effect size]), LLMs ($H = 6.71, p < 0.03, \eta^2 = 0.051$ [small effect size]), two prompt conditions ($U = 3780, p = .005, r = 0.23$ [small effect size]), utilized classifiers ($H = 12.49, p = 0.014, \eta^2 = 0.13$ [Moderate effect size]), as well as the utilized featurization technique in classifiers ($H = 12.30, p = 0.002, \eta^2 = 0.08$ [Moderate effect size]).

Pairwise post-hoc Dunn's tests with Benjamini-Hochberg corrections for multiple comparisons showed that the drops in the predictive powers when predicting age groups were significantly more than all other personal traits ($p < .001$ for all other personal traits); Llama 3 caused more drops in the predictive powers compared to Gemini ($p = 0.076$) and GPT3.5 ($p = 0.072$); SVM and Random Forest classifiers experienced less drops in the predictive power compared to Longformer classifier (differences were not significant—$p$-value $< 0.05$—after the applied corrections), while the drops comparing other classifiers were not significantly different. Finally, in terms of the featurization technique utilized in the trained classifiers, classifiers based on OpenAI embeddings experienced less drops in the predictive power compared to both TF-IDF ($p = 0.015$) and Longformer embeddings ($p = 0.021$).

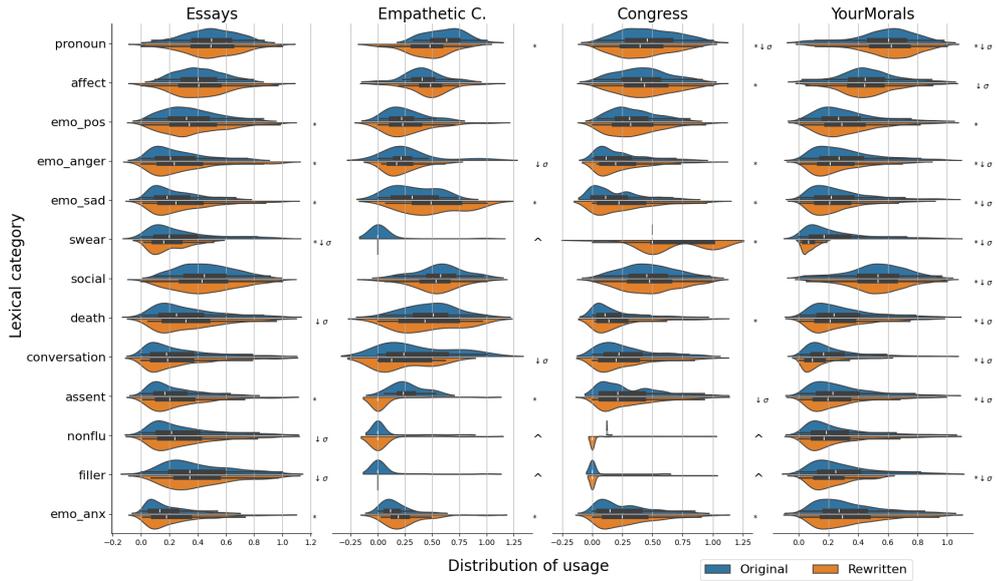

**Fig. D8**: Distribution of usage of different lexical word categories in the original and LLM-rewritten texts (with syntax_grammar prompt on Gemini) across different data sources. Items with a significant difference (adjusted $p$-value $< 0.05$) between original and LLM-rewritten distributions according to a Mann-Whitney U test are indicated with a *, while the items in which no more than 10% of the LLM-rewritten texts remain to contain the specified lexical feature are indicated using a ^. Furthermore, items where variance in features is significantly less in LLM-generated text compared to original documents using a Levene's test are denoted by $\downarrow \sigma$.

## E.4 What Author Attributes Do LLMs Promote?

In Study 3, utilizing the direction of prediction changes, we iterated on the characteristics that LLMs promote in their own version of authors' texts. We found our results to be consistent across different LLMs (see Figure E17) and prompts (see Figure E18).

In this section, we tried to provide a more fine-grained analysis for the investigated psychological constructs to contextualize our findings better with respect to the dimensions of each construct. In the case of specific characteristics related to personality, we observed that LLM-generated text is associated with people with higher levels of openness ($t(1074) = 6.50, p < .001, d = 0.21$ [small effect size]), agreeableness ($t(1005) = 9.05, p < .001, d = 0.27$ [small effect size]), and lower levels of extraversion ($t(1054) = 15.14, p < .001, d = 0.52$ [medium effect size]) compared to the actual authors.

In the case of specific characteristics related to dispositional empathy, we observed that LLM-generated text is associated with people with higher levels of personal distress ($t(319) = 3.90, p < .001, d = 0.36$ [small effect size]), and lower levels of

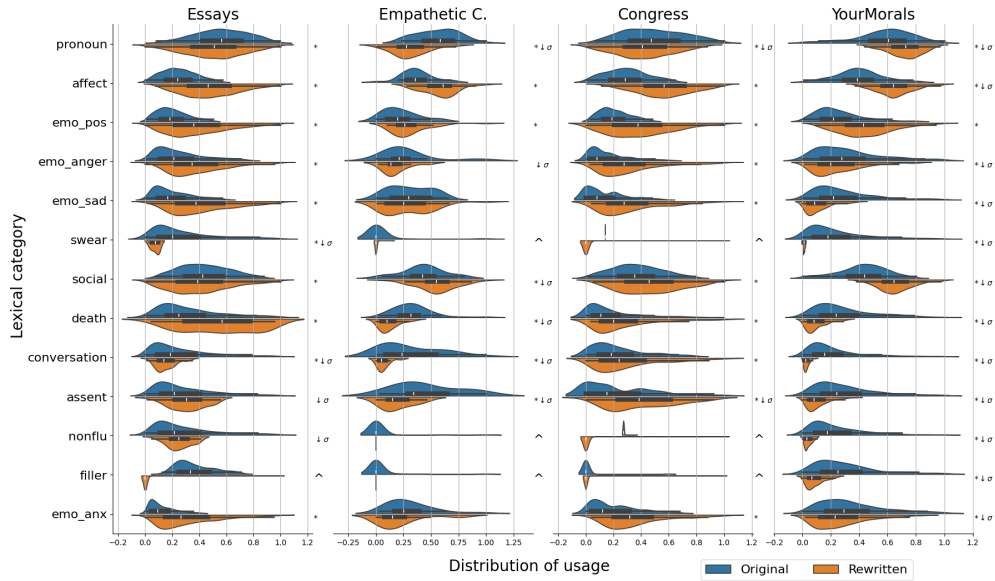

**Fig. D9**: Distribution of usage of different lexical word categories in the original and LLM-rewritten texts (with Rephrase prompt on Llama 3) across different data sources. Items with a significant difference (adjusted $p$-value $< 0.05$) between original and LLM-rewritten distributions according to a Mann-Whitney U test are indicated with a **\***, while the items in which no more than 10% of the LLM-rewritten texts remain to contain the specified lexical feature are indicated using a **^**. Furthermore, items where variance in features is significantly less in LLM-generated text compared to original documents using a Levene's test are denoted by $\downarrow \sigma$.

empathetic concern ($t(1945) = 10.6, p < .001, d = 0.35$ [small effect size]), perspective-taking ($t(2341) = 14.17, p < .001, d = 0.42$ [small effect size]), and fantasy ($t(2090) = 14.00, p < .001, d = 0.42$ [small effect size]) compared to the actual authors.

Finally, in the case of specific characteristics related to morality, we observed that LLM-generated text is associated with people with higher levels of fairness ($t(158) = 4.34, p < .001, d = 0.53$ [medium effect size]), care ($t(734) = 25.44, p < .001, d = 1.54$ [large effect size]), purity ($t(711) = 9.90, p < .001, d = 0.59$ [medium effect size]), and loyalty ($t(752) = 5.51, p < .001, d = 0.29$ [small effect size]) compared to the actual authors.

## Appendix F  Top-down Analyses

Table F7 demonstrates the associations between lexical cues in authors' texts and their personalities. Using the original texts, we replicated previously known associations in the literature [e.g., 82, 98, 99]. Namely, higher OPN was significantly associated with more complex language usage (BigWords) and swear words, higher CON was significantly associated with fewer swear and negative emotion words, higher EXT was

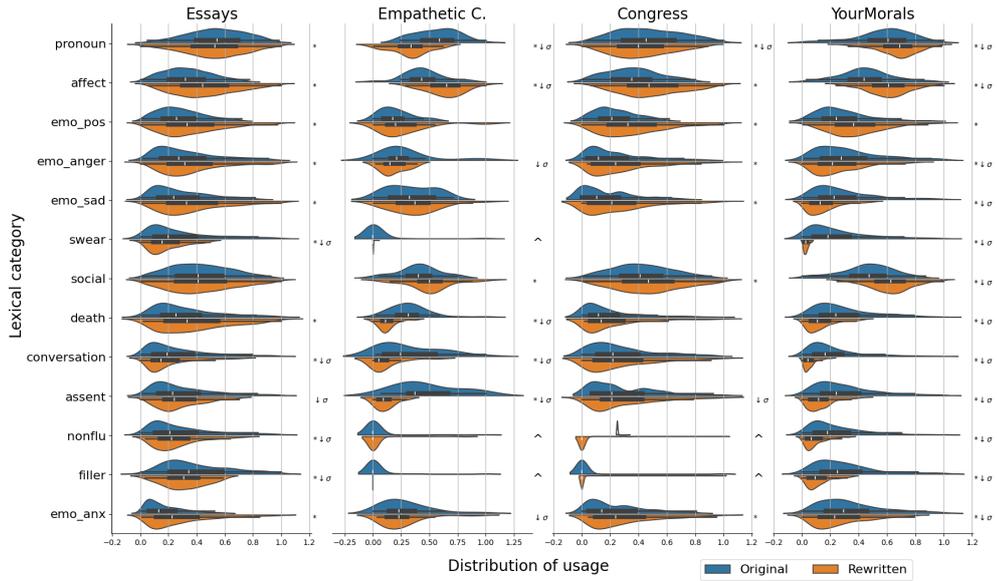

**Fig. D10**: Distribution of usage of different lexical word categories in the original and LLM-rewritten texts (with syntax_grammar prompt on Llama 3) across different data sources. Items with a significant difference (adjusted $p$-value $< .05$) between original and LLM-rewritten distributions according to a Mann-Whitney U test are indicated with a *, while the items in which no more than 10% of the LLM-rewritten texts remain to contain the specified lexical feature are indicated using a ^. Furthermore, items where variance in features is significantly less in LLM-generated text compared to original documents using a Levene's test are denoted by $\downarrow \sigma$.

significantly associated with greater use of positive emotion and social words, higher AGR was significantly associated with fewer swear and negative emotion words, and higher NEU was significantly associated with greater use of negative emotion words and pronouns. Certain associations present in the original texts were retained regardless of the LLM involved or prompt used (e.g., EXT and affiliation-related words; AGR and anger emotion words and NEU and negative emotion words). Other associations disappeared, regardless of the specific LLM or prompt (OPN and BigWords, and NEU and anger emotion words). Overall, these results suggest that the fine-grained lexical cues that allow for personality detection using theoretically grounded approaches might not be reliable when LLMs are involved in the writing process, and might be somewhat dependent on which LLM users choose to utilize and the personality dimension of interest.

Table F8 demonstrates the associations between lexical cues in authors' texts and their moral values. In line with evolutionary accounts of morality linking the development of moral values to the necessity of cooperation and interdependence [91], we found expected significant associations between MFT dimensions and social and affiliation-related words (e.g., between Loyalty and family-related words), and between

**Table D4**: Changes in the variance of various lexical word categories between original and LLM-rewritten documents across datasets, using GPT-3.5 and the rephrase prompt. Similar patterns were observed using other prompts and LLMs too. Variance values before and after rewriting are displayed on either side of $\rightarrow$. The statistics and $p$-values from Levene's test are provided in columns labeled $F$ and $p$. $p < .05$ is marked with $^*$ and $^{***}$ denotes $p < .001$.

| Lexical Category | Essays | $F$ | $p$ | Empathetic C. | $F$ | $p$ | Congress | $F$ | $p$ | YourMorals | $F$ | $p$ |
|---|---|---|---|---|---|---|---|---|---|---|---|---|
| pronoun | .044 → .042 | .075 | .783 | .032 → .027 | 1.06 | .304 | **.042 → .018** (↓) | 123.077 | < .001 *** | **.044 → .018** (↓) | 342.316 | < .001 *** |
| affect | .045 → .044 | .001 | .970 | .036 → .026 | .366 | .546 | **.027 → .040** (↑) | 25.132 | < .001 *** | **.035 → .022** (↓) | 96.369 | < .001 *** |
| emo_pos | .050 → .049 | .004 | .949 | .073 → .051 | .330 | .567 | **.022 → .057** (↑) | 139.282 | < .001 *** | **.034 → .040** (↑) | 29.640 | < .001 *** |
| emo_anger | **.057 → .043** (↓) | 11.009 | < .001 *** | .060 → .008 | 2.281 | .139 | .048 → .056 | 2.966 | .085 | **.045 → .023** (↓) | 11.990 | < .001 *** |
| emo_sad | **.037 → .054** (↑) | 28.547 | < .001 *** | .031 → .048 | 1.747 | .189 | **.020 → .050** (↓) | 63.854 | < .001 *** | **.040 → .015** (↓) | 191.865 | < .001 *** |
| swear | **.057 → .001** (↓) | 107.645 | < .001 *** | .033 → .000 | 3.074 | .082 | .0000 → .006 | 1.706 | .191 | **.049 → .000** (↓) | 283.577 | < .001 *** |
| social | .048 → .052 | 3.542 | .059 | .043 → .042 | .000 | .995 | .049 → .054 | 3.623 | .057 | **.042 → .021** (↓) | 243.742 | < .001 *** |
| death | **.056 → .033** (↓) | 6.872 | .008 * | **.033 → .010** (↓) | 4.158 | .044 * | .045 → .044 | .028 | .866 | **.043 → .017** (↓) | 157.110 | < .001 *** |
| conversation | **.050 → .020** (↓) | 117.703 | < .001 *** | **.071 → .005** (↓) | 51.776 | < .001 *** | **.042 → .064** (↑) | 16.420 | .000 *** | **.035 → .007** (↓) | 577.565 | < .001 *** |
| assent | **.0512 → .023** (↓) | 58.888 | < .001 *** | **.070 → .011** (↓) | 19.542 | < .001 *** | **.064 → .022** (↓) | 11.480 | .000 *** | **.042 → .022** (↓) | 107.830 | < .001 *** |
| nonflu | **.053 → .027** (↓) | 4.319 | < .001 *** | .023 → .000 | 2.09 | .150 | .000 → .003 | .084 | .771 | **.039 → .014** (↓) | 119.135 | < .001 *** |
| filler | **.058 → .008** (↓) | 87.057 | < .001 *** | .021 → .000 | 3.136 | .079 | **.024 → .000** (↓) | 18.442 | < .001 *** | **.050 → .008** (↓) | 212.726 | < .001 *** |
| emo_anx | .047 → .051 | 2.429 | .119 | **.020 → .048** (↑) | 7.247 | .008 * | **.026 → .041** (↑) | 3.910 | .048 * | **.048 → .032** (↓) | 33.227 | < .001 *** |

**Table E5**: Comparison between the performance of trained classifier and Llama 3.1 [58] zero-shot capabilities on each personal trait prediction task.

|  | Mean $F_{1\text{trained classifiers}}$ | Mean $F_{1\text{LLM as classifier}}$ | Random baseline |
|---|---|---|---|
| Age group | .351 | .121 | .244 |
| Empathy | .657 | .436 | .541 |
| Personality | .658 | .569 | .514 |
| Gender | .694 | .494 | .495 |
| Morality | .664 | .405 | .521 |
| Affiliation | .640 | .602 | .490 |

**Table E6**: Paired $t$-tests for testing the difference in predictive powers ($F_1$) of classifiers on the original and LLM-generated texts for different dimensions of personal traits, and Cohen's $d$ effect sizes for the magnitude of these differences ($|d| < 0.2$: negligible, $|d| < 0.5$: small, $|d| < 0.8$: moderate; 81).

| Dataset | Dim. | Mean $F_{1\text{original}}$ | Mean $F_{1\text{LLM}}$ | CI | SE | t | df | p | d | N | magnitude | Random Baseline |
|---|---|---|---|---|---|---|---|---|---|---|---|---|
| Personality | CON | 0.647 | 0.593 | [0.045, 0.061] | 0.159 | 13.085 | 1534 | 0 | 0.451 | 1535 | Small | 0.511 |
|  | NEU | 0.618 | 0.556 | [0.054, 0.071] | 0.162 | 14.727 | 1457 | 0 | 0.533 | 1458 | Moderate | 0.501 |
|  | OPN | 0.673 | 0.615 | [0.05, 0.065] | 0.164 | 14.438 | 1693 | 0 | 0.487 | 1694 | Small | 0.516 |
|  | EXT | 0.663 | 0.602 | [0.052, 0.07] | 0.161 | 14.797 | 1530 | 0 | 0.463 | 1531 | Small | 0.516 |
|  | AGR | 0.685 | 0.640 | [0.037, 0.055] | 0.170 | 10.224 | 1434 | 0 | 0.370 | 1435 | Small | 0.531 |
| Empathy | Perspective | 0.637 | 0.584 | [0.046, 0.06] | 0.138 | 18.613 | 2334 | 0 | 0.463 | 2335 | Small | 0.527 |
|  | Fantasy | 0.659 | 0.619 | [0.034, 0.047] | 0.117 | 15.727 | 2080 | 0 | 0.393 | 2081 | Small | 0.547 |
|  | Concern | 0.680 | 0.614 | [0.059, 0.073] | 0.145 | 19.966 | 1939 | 0 | 0.587 | 1940 | Moderate | 0.572 |
|  | Distress | 0.637 | 0.554 | [0.057, 0.108] | 0.164 | 7.115 | 200 | 0 | 0.640 | 201 | Moderate | 0.518 |
| Morality | Authority | 0.649 | 0.548 | [0.091, 0.112] | 0.180 | 19.581 | 1208 | 0 | 0.775 | 1209 | Moderate | 0.534 |
|  | Purity | 0.658 | 0.572 | [0.076, 0.096] | 0.177 | 17.249 | 1258 | 0 | 0.645 | 1259 | Moderate | 0.536 |
|  | Loyalty | 0.634 | 0.555 | [0.069, 0.088] | 0.181 | 16.206 | 1385 | 0 | 0.638 | 1386 | Moderate | 0.503 |
|  | Care | 0.619 | 0.648 | [-0.037, -0.021] | 0.102 | -9.900 | 1207 | 0 | 0.301 | 1208 | Small | 0.507 |
|  | Fairness | 0.635 | 0.521 | [0.079, 0.149] | 0.224 | 5.872 | 133 | 0 | 0.789 | 134 | Moderate | 0.528 |
| Affiliation |  | 0.664 | 0.591 | [0.064, 0.082] | 0.168 | 15.707 | 1314 | 0 | 0.610 | 1315 | Moderate | 0.500 |
| Gender |  | 0.694 | 0.623 | [0.063, 0.08] | 0.166 | 15.965 | 1371 | 0 | 0.603 | 1372 | Moderate | 0.500 |
| Age group |  | 0.351 | 0.260 | [0.079, 0.102] | 0.167 | 15.397 | 804 | 0 | 0.776 | 805 | Moderate | 0.250 |

foundations such as Purity and Authority and word categories such as religion-related words. Some associations present in the original texts were washed away after LLM

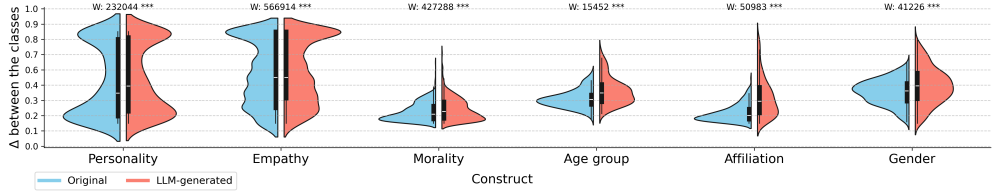

**Fig. E11**: The distribution of Δ (a proxy for imbalances between the predicted class frequencies on the original and LLM-rewritten texts) across different personal traits, focusing on the LLM-rewritten texts generated by Gemini. The $W$ statistics from the Wilcoxon test are displayed on top, with $p < .05$, $p < .01$, and $p < .001$, marked with *, **, ***, respectively.

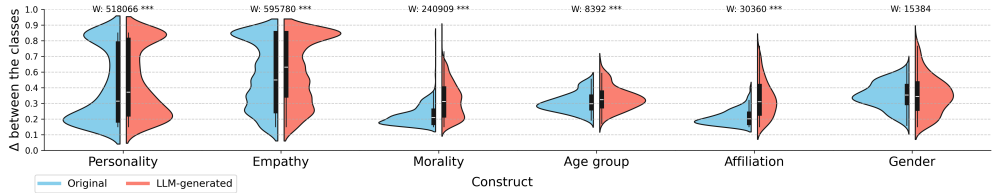

**Fig. E12**: The distribution of Δ (a proxy for imbalances between the predicted class frequencies on the original and LLM-rewritten texts) across different personal traits, focusing on the LLM-rewritten texts generated by Llama 3. The $W$ statistics from the Wilcoxon test are displayed on top, with $p < .05$, $p < .01$, and $p < .001$, marked with *, **, ***, respectively.

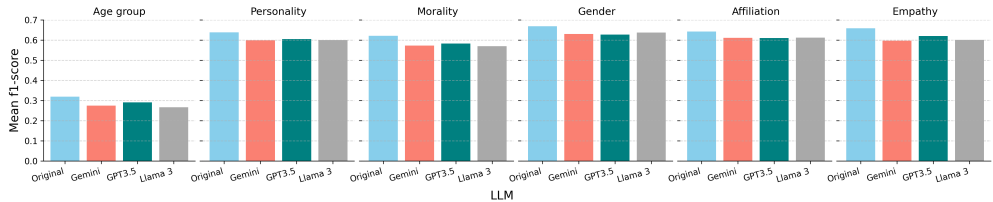

**Fig. E13**: Predictive power (mean $F_1$) of text over personal traits in original and LLM-rewritten texts across different LLMs and data sources.

involvement, primarily with Llama 3 (e.g., between Purity and family-related words, Authority and religion-related words). These results are aligned with our observations in Section E.3, underscoring Gemini and GPT3.5 as more preservative of lexical cues predictive of authors' moral values, than Llama 3.

Table F9 demonstrates the associations between lexical cues in authors' texts and authors' dispositional empathy. For different dimensions of dispositional empathy, we expected and found significant associations with the use of pronouns, emotion words, and words related to cognitive processes (i.e., cogproc), as well as with words from the

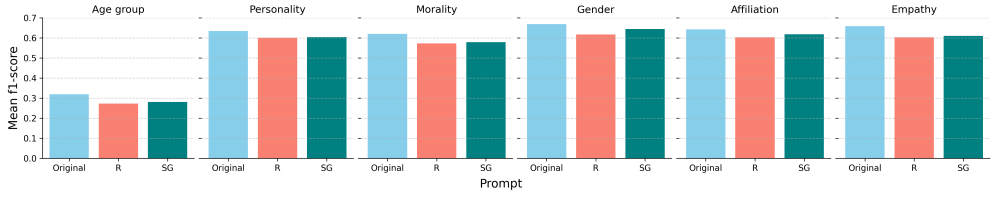

**Fig. E14**: Predictive power (mean $F_1$) of text over personal traits in original and LLM-rewritten texts across different prompts and data sources. SG represents the Syntax_grammar, and R represents the Rephrase prompts.

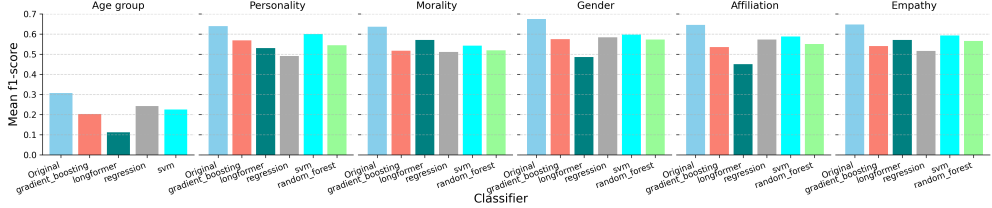

**Fig. E15**: Predictive power (mean $F_1$) of text over personal traits in original and LLM-rewritten texts across different classifiers and data sources.

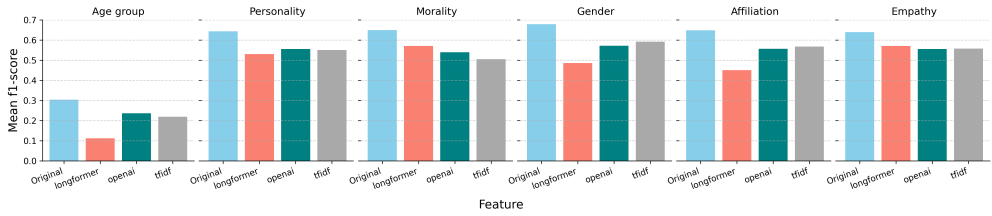

**Fig. E16**: Predictive power (mean $F_1$) of text over personal traits in original and LLM-rewritten texts across different featurization techniques and data sources.

empathy and distress lexicons. Hypothesized associations were present in the original text for all subdimensions of IRI, e.g., between PD and affect-related words, EC and pronouns, as well as tentative-related words (tentat), PT and pronouns, and FS and affect. The significant negative association of IRI dimensions PT and EC with first-person plural words (we) was the only association that retained its significance across almost all LLM rewrite conditions, with all other associations becoming unreliable.

The associations between lexical cues in authors' texts and their demographic variables are demonstrated in Table F10, Table F11, and Table F12. We found several expected significant differences in the usage of lexical cues across political affiliations (we did not find hypothesized differences across political affiliation in language use related to inhibition and reaction to threats, e.g., negative emotion words; 100) and genders. Namely, we found that Republicans used a significantly greater number of adverbs than Democrats, and males used a significantly greater number of articles

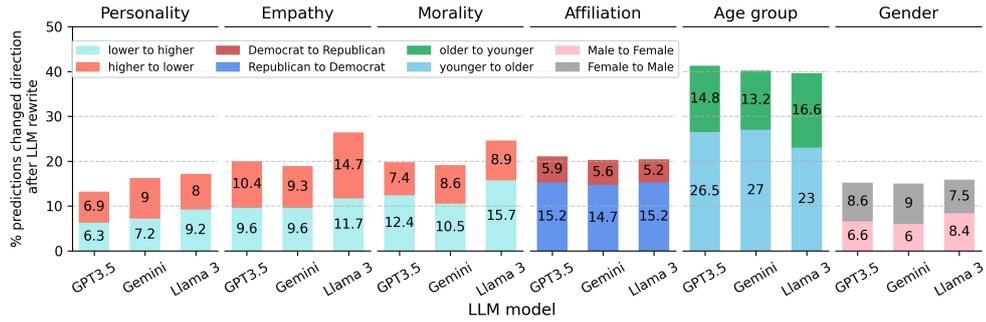

**Fig. E17**: Percentage of original texts with correct author attribute predictions that changed after LLM rewriting, grouped by the direction of change in predictions, using different LLMs.

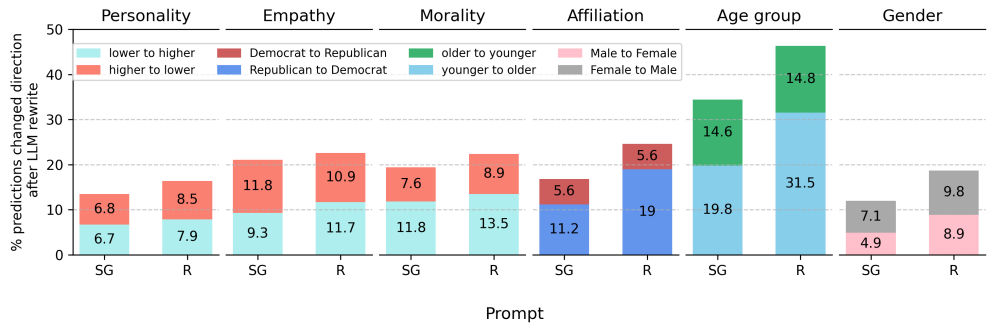

**Fig. E18**: Percentage of original texts with correct author attribute predictions that changed after LLM rewriting, grouped by the direction of change in predictions, using different prompts. **SG** and **R** stand for Syntax_Grammar and Rephrase, respectively.

and a significantly smaller number of social and anxiety-related words than females [89]. We also found associations between first-person plural words and words related to cognitive processes (e.g., but, not, if, or, know) with age as well as words related to focusing on the future with this construct.

After the LLM rewrites, some previously expected associations for political affiliation retained significance, while others were washed away. The linguistic cues for gender were generally preserved for article and social word usage, but only GPT-3.5 using a Syntax_Grammar prompt, preserved the linguistic cues related to negative emotions. After the LLMs' involvement in rewriting the texts, although the associations related to first-person plural words and words related to cognitive processes with age remained preserved, the associations with the words related to focusing on the future were washed away.

52

**Table F7**: Pearson correlations between LIWC and NRC dictionary word frequencies, and the Big Five personality dimensions, on the authors' original texts and LLM-generated texts, across different utilized LLMs and prompts. Significant correlations are **boldfaced**.

| Personality Dimension | Linguistic Cue | Original | Rephrase | | | Syntax-Grammar | | |
|---|---|---|---|---|---|---|---|---|
| | | | Gemini | GPT3.5 | Llama 3 | Gemini | GPT3.5 | Llama 3 |
| OPN | nrc.negative | **0.06** | 0.02 | -0.01 | -0.02 | 0.04 | 0.01 | 0.01 |
| | i | **-0.13** | **-0.07** | **-0.11** | **-0.07** | **-0.14** | **-0.12** | **-0.09** |
| | swear | **0.09** | 0.04 | **0.09** | **0.06** | **0.08** | **0.07** | **0.09** |
| | BigWords | **0.09** | 0.04 | 0.00 | 0.04 | **0.07** | -0.01 | 0.03 |
| | nrc.disgust | **0.09** | 0.05 | 0.03 | 0.03 | 0.06 | 0.03 | **0.07** |
| CON | nrc.negative | **-0.06** | -0.03 | **-0.06** | -0.04 | -0.02 | -0.06 | -0.02 |
| | emo_anger | **-0.12** | **-0.12** | **-0.11** | -0.04 | -0.06 | **-0.12** | **-0.08** |
| | swear | **-0.11** | -0.01 | **-0.07** | -0.03 | -0.05 | **-0.09** | **-0.08** |
| | nrc.disgust | **-0.08** | **-0.07** | **-0.08** | -0.03 | -0.00 | **-0.07** | -0.04 |
| | nrc.sadness | **-0.07** | -0.03 | **-0.07** | -0.04 | -0.02 | -0.05 | -0.03 |
| EXT | pronoun | **0.06** | 0.04 | 0.05 | -0.02 | 0.04 | **0.06** | 0.04 |
| | nrc.sadness | **-0.06** | -0.01 | -0.04 | **-0.06** | -0.02 | -0.02 | -0.02 |
| | nrc.negative | **-0.06** | 0.00 | -0.06 | -0.04 | -0.02 | -0.04 | -0.01 |
| | emo_pos | **0.10** | **0.09** | **0.12** | **0.10** | 0.05 | **0.11** | **0.09** |
| | nrc.joy | **0.08** | **0.07** | **0.09** | **0.11** | **0.07** | **0.09** | **0.08** |
| | drives | **0.09** | **0.07** | **0.07** | **0.08** | **0.08** | **0.09** | **0.10** |
| | Social | **0.07** | 0.01 | **0.08** | 0.06 | 0.04 | **0.06** | **0.07** |
| | affiliation | **0.11** | **0.09** | **0.11** | **0.09** | **0.10** | **0.11** | **0.13** |
| AGR | emo_pos | **0.06** | 0.04 | **0.06** | 0.05 | **0.07** | 0.05 | **0.07** |
| | nrc.negative | **-0.10** | -0.04 | -0.05 | -0.06 | **-0.08** | **-0.07** | **-0.07** |
| | swear | **-0.12** | -0.02 | **-0.06** | -0.03 | **-0.08** | **-0.08** | **-0.08** |
| | nrc.disgust | **-0.08** | -0.06 | -0.05 | -0.05 | -0.05 | -0.06 | -0.05 |
| | emo_anger | **-0.09** | **-0.08** | **-0.11** | **-0.08** | **-0.10** | **-0.08** | **-0.10** |
| | affiliation | **0.08** | **0.08** | **0.09** | **0.07** | **0.08** | **0.08** | **0.07** |
| NEU | nrc.positive | **-0.07** | -0.02 | **-0.06** | **-0.08** | -0.04 | -0.04 | **-0.09** |
| | nrc.disgust | **0.07** | **0.08** | **0.07** | **0.07** | 0.02 | **0.08** | 0.05 |
| | nrc.sadness | **0.13** | **0.12** | **0.11** | **0.16** | 0.06 | **0.10** | **0.10** |
| | nrc.anger | **0.08** | **0.14** | **0.12** | **0.15** | **0.07** | **0.10** | **0.08** |
| | emo_sad | **0.10** | **0.10** | **0.10** | **0.10** | **0.08** | **0.10** | 0.05 |
| | emo_anger | **0.08** | 0.06 | **0.07** | 0.05 | 0.06 | **0.06** | **0.06** |
| | emo_neg | **0.19** | **0.18** | **0.20** | **0.21** | **0.15** | **0.21** | **0.18** |
| | emotion | **0.10** | **0.14** | **0.12** | **0.11** | **0.10** | **0.12** | **0.06** |
| | pronoun | **0.12** | 0.01 | **0.11** | **0.09** | **0.10** | **0.09** | **0.10** |
| | i | **0.15** | -0.02 | **0.16** | **0.15** | **0.15** | **0.14** | **0.12** |

**Table F8**: Pearson correlations between relevant dictionary categories and the moral foundations of Fairness, Loyalty, Authority, Purity, and Care before and after LLM (Gemini, GPT3.5, Llama 3) rewrite. Significant correlations are **boldfaced**.

| Morality Dimension | Linguistic Cue | Original | Rephrase Gemini | Rephrase GPT3.5 | Rephrase Llama 3 | Syntax-Grammar Gemini | Syntax-Grammar GPT3.5 | Syntax-Grammar Llama 3 |
|---|---|---|---|---|---|---|---|---|
| Fairness | mfd.authority.vice | **-0.06** | -0.04 | -0.04 | -0.03 | -0.05 | -0.04 | -0.04 |
| | relig | **-0.06** | -0.03 | -0.05 | -0.01 | -0.05 | **-0.06** | -0.02 |
| Loyalty | mfd.care.virtue | **0.06** | 0.04 | 0.04 | 0.07 | 0.05 | 0.04 | 0.06 |
| | mfd.sanctity.virtue | **0.14** | **0.10** | **0.12** | 0.08 | **0.13** | **0.12** | **0.12** |
| | friend | **0.07** | **0.06** | 0.06 | 0.07 | **0.06** | 0.06 | 0.06 |
| | family | **0.11** | **0.12** | **0.12** | 0.09 | **0.11** | **0.11** | 0.10 |
| | socrefs | **0.09** | **0.13** | **0.11** | **0.16** | **0.11** | **0.10** | 0.09 |
| | Affect | **0.10** | **0.10** | **0.13** | **0.21** | **0.12** | **0.11** | **0.11** |
| | relig | **0.12** | **0.12** | **0.13** | 0.09 | **0.12** | **0.12** | **0.11** |
| | Social | **0.08** | **0.12** | **0.10** | **0.17** | **0.10** | **0.11** | **0.12** |
| | affiliation | **0.06** | **0.10** | **0.08** | **0.11** | **0.07** | 0.06 | 0.07 |
| | you | **0.08** | **0.10** | **0.09** | **0.15** | **0.09** | **0.10** | **0.12** |
| Authority | socrefs | **0.16** | **0.15** | **0.14** | **0.19** | **0.16** | **0.13** | **0.12** |
| | pronoun | **0.13** | **0.11** | **0.11** | **0.18** | **0.13** | **0.12** | **0.12** |
| | friend | **0.07** | **0.06** | 0.05 | **0.08** | 0.06 | 0.04 | 0.07 |
| | mfd.care.virtue | **0.09** | **0.06** | **0.06** | 0.10 | **0.08** | 0.06 | 0.08 |
| | Affect | **0.15** | **0.13** | **0.13** | **0.25** | **0.16** | **0.13** | **0.15** |
| | i | **0.07** | **0.07** | **0.08** | **0.12** | **0.08** | **0.08** | **0.11** |
| | affiliation | **0.12** | **0.11** | **0.11** | **0.14** | **0.11** | **0.09** | **0.10** |
| | prosocial | **0.06** | **0.07** | 0.05 | **0.13** | **0.07** | 0.06 | **0.12** |
| | family | **0.16** | **0.15** | **0.17** | **0.14** | **0.15** | **0.16** | **0.14** |
| | relig | **0.16** | **0.15** | **0.15** | 0.06 | **0.16** | **0.16** | 0.09 |
| | mfd.sanctity.virtue | **0.16** | **0.13** | **0.15** | 0.04 | **0.15** | **0.16** | 0.10 |
| | you | **0.11** | **0.11** | **0.09** | **0.19** | **0.11** | **0.09** | **0.14** |
| | Social | **0.16** | **0.14** | **0.12** | **0.17** | **0.16** | **0.12** | **0.14** |
| Purity | we | **0.07** | **0.09** | **0.08** | 0.06 | **0.08** | **0.06** | 0.02 |
| | socrefs | **0.19** | **0.20** | **0.20** | **0.19** | **0.20** | **0.19** | **0.13** |
| | pronoun | **0.13** | **0.11** | **0.13** | **0.17** | **0.13** | **0.14** | **0.11** |
| | friend | **0.07** | **0.07** | **0.06** | 0.08 | **0.07** | **0.06** | 0.10 |
| | mfd.care.virtue | **0.11** | **0.09** | **0.09** | 0.09 | **0.11** | **0.09** | 0.07 |
| | Affect | **0.12** | **0.12** | **0.13** | **0.20** | **0.15** | **0.11** | **0.12** |
| | i | **0.06** | 0.04 | **0.06** | **0.11** | 0.05 | **0.06** | **0.11** |
| | affiliation | **0.15** | **0.15** | **0.15** | **0.14** | **0.14** | **0.12** | **0.10** |
| | prosocial | **0.06** | **0.07** | 0.05 | **0.12** | **0.08** | **0.06** | **0.13** |
| | family | **0.15** | **0.15** | **0.17** | 0.10 | **0.15** | **0.15** | 0.10 |
| | relig | **0.21** | **0.21** | **0.21** | **0.11** | **0.21** | **0.22** | **0.16** |
| | mfd.sanctity.virtue | **0.21** | **0.19** | **0.22** | **0.11** | **0.21** | **0.22** | **0.18** |
| | mfd.sanctity.vice | **-0.06** | 0.01 | 0.02 | -0.02 | 0.00 | -0.02 | -0.05 |
| | you | **0.13** | **0.13** | **0.11** | **0.17** | **0.13** | **0.12** | **0.15** |
| | Social | **0.19** | **0.20** | **0.18** | **0.18** | **0.20** | **0.18** | **0.16** |
| Care | mfd.care.virtue | **0.09** | **0.09** | **0.10** | 0.02 | **0.07** | **0.09** | 0.08 |
| | mfd.authority.vice | **-0.06** | -0.03 | -0.03 | -0.04 | -0.04 | -0.03 | -0.06 |
| | affiliation | **0.07** | **0.09** | **0.10** | 0.06 | **0.08** | **0.07** | 0.07 |

**Table F9**: Pearson correlations between word frequencies of LIWC categories and personal distress (PD), empathetic concern (EC), Fantasy (FS), and perspective-taking (PT) before and after LLM (Gemini, GPT3.5, Llama 3) rewrites. Significant correlations are **boldfaced**.

| Dispositional Empathy | | | Rephrase | | | Syntax-Grammar | | |
|---|---|---|---|---|---|---|---|---|
| Dimension | Linguistic Cue | Original | Gemini | GPT3.5 | Llama 3 | Gemini | GPT3.5 | Llama 3 |
| PD | Affect | **0.13** | 0.06 | 0.07 | **0.09** | 0.08 | 0.06 | **0.12** |
|  | differ | **0.09** | **0.11** | **0.16** | **0.09** | **0.12** | **0.10** | **0.08** |
|  | we | **0.08** | 0.02 | 0.07 | 0.05 | **0.11** | 0.07 | **0.08** |
|  | tentat | **0.11** | 0.06 | **0.07** | 0.06 | 0.05 | 0.07 | **0.10** |
|  | empathy.low_distress | **0.10** | 0.04 | -0.02 | 0.03 | -0.00 | 0.02 | **0.10** |
| EC | differ | **-0.10** | -0.07 | **-0.14** | **-0.09** | -0.08 | **-0.13** | **-0.11** |
|  | we | **-0.22** | **-0.13** | **-0.17** | **-0.11** | **-0.20** | **-0.21** | **-0.15** |
|  | shehe | **0.14** | 0.08 | **0.14** | 0.07 | **0.11** | **0.13** | **0.13** |
|  | pronoun | **-0.15** | 0.00 | **-0.13** | 0.00 | -0.07 | **-0.11** | -0.05 |
|  | emo_neg | **0.13** | 0.07 | **0.14** | **0.11** | **0.13** | **0.12** | **0.11** |
|  | tentat | **-0.13** | **-0.12** | **-0.16** | -0.02 | **-0.13** | **-0.17** | **-0.09** |
|  | cogproc | **-0.13** | **-0.12** | **-0.14** | **-0.12** | **-0.13** | **-0.13** | -0.07 |
|  | empathy.low_distress | **-0.09** | **-0.09** | -0.04 | **-0.11** | -0.05 | -0.03 | **-0.16** |
| FS | Affect | **0.08** | 0.07 | **0.09** | 0.02 | 0.05 | **0.12** | 0.01 |
|  | emo_neg | **0.11** | 0.07 | **0.09** | 0.05 | **0.08** | **0.11** | **0.07** |
|  | empathy.low_empathy | **-0.08** | -0.05 | -0.03 | **-0.08** | -0.02 | -0.04 | -0.06 |
|  | empathy.low_distress | **-0.10** | -0.03 | -0.01 | -0.06 | -0.02 | -0.04 | -0.07 |
| PT | we | **-0.22** | **-0.17** | **-0.17** | **-0.13** | **-0.19** | **-0.22** | **-0.18** |
|  | shehe | **0.09** | 0.03 | **0.10** | **0.08** | 0.04 | **0.09** | **0.09** |
|  | pronoun | **-0.22** | -0.06 | **-0.20** | -0.03 | **-0.14** | **-0.17** | **-0.07** |
|  | tentat | **-0.09** | -0.05 | **-0.11** | 0.00 | **-0.12** | **-0.12** | -0.05 |
|  | cogproc | **-0.12** | **-0.14** | **-0.18** | **-0.10** | **-0.13** | **-0.14** | **-0.08** |

**Table F10**: Average word frequencies ($M$ean) for relevant LIWC categories in two political affiliations (Democrat or Republican) before and after LLM (Gemini, GPT3.5, Llama 3) rewrites. Significant t-tests are **boldfaced**.

| Political Aff. | | | Rephrase | | | | | | Syntax-Grammar | | | | | |
|---|---|---|---|---|---|---|---|---|---|---|---|---|---|---|
|  | Original | | Gemini | | GPT3.5 | | Llama 3 | | Gemini | | GPT3.5 | | Llama 3 | |
| Linguistic Cue | $M_D$ | $M_R$ | $M_D$ | $M_R$ | $M_D$ | $M_R$ | $M_D$ | $M_R$ | $M_D$ | $M_R$ | $M_D$ | $M_R$ | $M_D$ | $M_R$ |
| mfd.loyalty.virtue | 0.02 | 0.02 | **0.03** | **0.02** | **0.03** | **0.03** | 0.03 | 0.03 | 0.02 | 0.02 | **0.03** | **0.03** | **0.03** | **0.02** |
| emo_anx | **0.07** | **0.06** | **0.08** | **0.05** | 0.07 | 0.07 | 0.06 | 0.06 | **0.07** | **0.06** | 0.08 | 0.07 | **0.07** | **0.06** |
| adverb | **3.41** | **3.71** | 1.98 | 2.07 | **2.22** | **2.35** | **2.31** | **2.45** | **3.06** | **3.28** | **2.38** | **2.52** | **2.82** | **3.04** |
| emo_neg | **0.31** | **0.27** | 0.31 | 0.28 | 0.26 | 0.25 | 0.25 | 0.25 | **0.30** | **0.26** | 0.29 | 0.27 | 0.30 | 0.27 |
| i | **1.64** | **1.75** | 1.16 | 1.27 | 1.78 | 1.88 | 1.88 | 2.00 | **1.66** | **1.79** | 1.63 | 1.68 | 1.73 | 1.82 |
| cogproc | **9.28** | **9.58** | **8.32** | **8.77** | **8.08** | **8.54** | **8.87** | **9.43** | **9.14** | **9.48** | **8.22** | **8.64** | **9.01** | **9.41** |
| certitude | **0.49** | **0.56** | **0.22** | **0.24** | **0.21** | **0.24** | **0.23** | **0.27** | **0.43** | **0.48** | **0.27** | **0.30** | **0.37** | **0.41** |

**Table F11**: Average word frequencies (*M*ean) for relevant LIWC categories in two investigated categories of gender (Male and Female) before and after LLM (Gemini, GPT3.5, Llama 3) rewrites. Significant t-tests are **boldfaced**.

| Gender | | | Rephrase | | | | | | Syntax-Grammar | | | | | |
|---|---|---|---|---|---|---|---|---|---|---|---|---|---|---|
| | Original | | Gemini | | GPT3.5 | | Llama 3 | | Gemini | | GPT3.5 | | Llama 3 | |
| Linguistic Cue | $M_M$ | $M_F$ | $M_M$ | $M_F$ | $M_M$ | $M_F$ | $M_M$ | $M_F$ | $M_M$ | $M_F$ | $M_M$ | $M_F$ | $M_M$ | $M_F$ |
| article | **8.52** | **8.01** | **8.72** | **8.20** | **8.45** | **7.99** | **8.98** | **8.41** | **8.60** | **8.01** | **8.79** | **8.31** | **8.75** | **8.17** |
| social | **7.51** | **8.26** | **6.27** | **7.01** | **6.14** | **6.69** | **7.52** | **8.06** | **7.18** | **8.01** | **5.97** | **6.55** | **7.43** | **8.16** |
| emo_anx | **0.06** | **0.07** | **0.05** | **0.08** | **0.06** | **0.08** | **0.05** | **0.07** | **0.06** | **0.07** | **0.07** | **0.08** | **0.06** | **0.07** |
| i | **1.65** | **1.76** | 1.20 | 1.23 | 1.79 | 1.88 | 1.91 | 1.97 | **1.66** | **1.80** | **1.62** | **1.70** | 1.75 | 1.81 |
| emo_neg | **0.28** | **0.30** | 0.30 | 0.30 | 0.25 | 0.27 | 0.24 | 0.26 | 0.27 | 0.29 | **0.27** | **0.29** | 0.28 | 0.29 |
| affect | **4.62** | **4.78** | 5.64 | 5.80 | **5.70** | **5.96** | **5.92** | **6.08** | **4.70** | **4.87** | **5.14** | **5.37** | **5.09** | **5.25** |
| tentat | **1.41** | **1.30** | **0.89** | **0.83** | **0.91** | **0.80** | **0.95** | **0.88** | **1.35** | **1.21** | **1.10** | **0.99** | **1.17** | **1.08** |
| swear | **0.01** | **0.00** | 0.00 | 0.00 | 0.00 | 0.00 | 0.00 | 0.00 | **0.01** | **0.00** | 0.00 | 0.00 | 0.00 | 0.00 |
| cogproc | **9.50** | **9.33** | **8.69** | **8.34** | **8.52** | **8.02** | **9.27** | **8.97** | **9.48** | **9.07** | **8.60** | **8.19** | **9.37** | **8.99** |

**Table F12**: Pearson correlations between LIWC categories and age before and after LLM (Gemini, GPT3.5, Llama 3) rewrites. Significant correlations are **boldfaced**.

| Age | | Rephrase (R) | | | Syntax-Grammar (SG) | | |
|---|---|---|---|---|---|---|---|
| Linguistic Cue | Original | Gemini | GPT3.5 | Llama 3 | Gemini | GPT3.5 | Llama 3 |
| we | **-0.07** | **-0.16** | **-0.10** | **-0.12** | -0.06 | **-0.12** | **-0.07** |
| cogproc | **-0.06** | **-0.13** | **-0.11** | **-0.11** | **-0.08** | **-0.15** | **-0.10** |
| social | **-0.07** | **-0.11** | **-0.09** | **-0.09** | **-0.07** | **-0.09** | **-0.09** |
| focusfuture | **-0.09** | -0.05 | **-0.09** | **-0.10** | **-0.07** | **-0.08** | **-0.09** |
| emo_neg | **-0.06** | -0.05 | **-0.10** | -0.06 | **-0.07** | **-0.09** | **-0.08** |




\end{document}